\PassOptionsToPackage{headheight=58pt}{geometry}
\documentclass[]{bytedance_seed}

\usepackage{amsmath,amssymb,amsthm}
\usepackage{pifont}
\newcommand{\xmark}{\ding{55}}
\usepackage{mathtools}
\usepackage{array}
\usepackage{url}
\usepackage{tikz}
\usepackage{float}
\usepackage{placeins}
\usetikzlibrary{arrows.meta,calc,positioning,fit,decorations.pathreplacing}

\usepackage{flafter}

\newcommand{\fittotextwidth}[1]{\resizebox{\textwidth}{!}{#1}}

\newcommand{\archname}{SMELT}

\newcommand{\institutionlogobar}[1][\linewidth]{%
  \resizebox{#1}{!}{%
    \begin{tabular}{@{}c@{\hspace{2.5mm}}c@{\hspace{2.5mm}}c@{\hspace{2.5mm}}c@{}}
      \raisebox{-2.9mm}{\includegraphics[width=52mm,keepaspectratio]{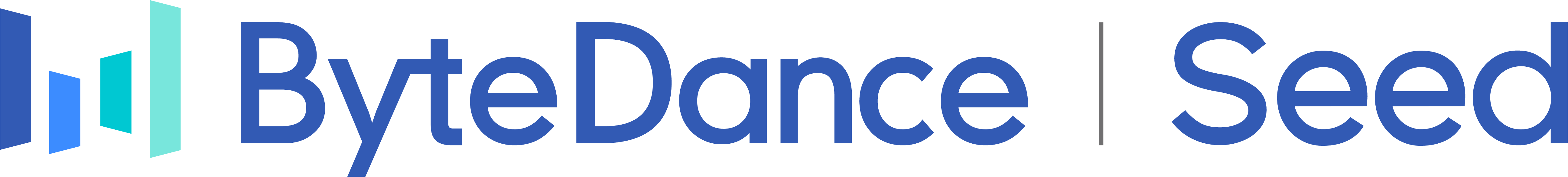}} &
      \raisebox{-7.425mm}{\makebox[16.5mm][c]{\includegraphics[height=14.85mm,keepaspectratio]{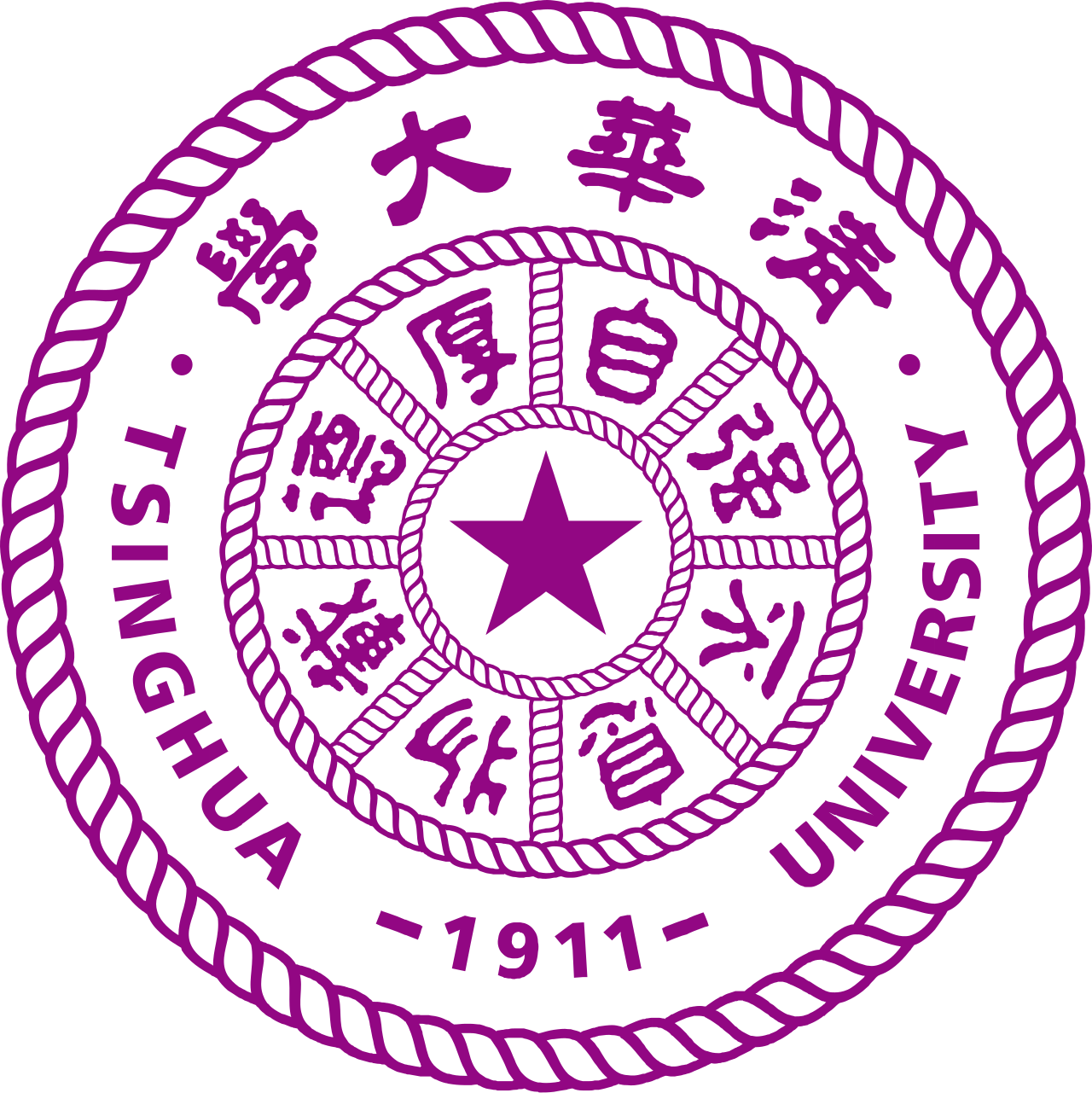}}} &
      \raisebox{-7.425mm}{\makebox[16.5mm][c]{\includegraphics[height=14.85mm,keepaspectratio]{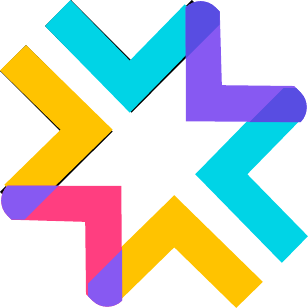}}} &
      \raisebox{-4mm}{\includegraphics[width=46mm,keepaspectratio]{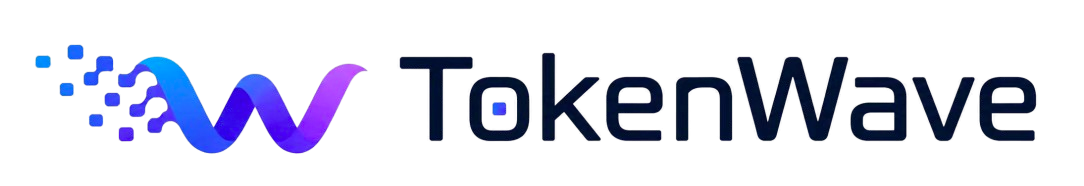}}
    \end{tabular}%
  }%
}
\fancypagestyle{firststyle}{%
  \fancyhf{}%
  \fancyhead[C]{\raisebox{-2mm}[0pt][0pt]{\institutionlogobar}}%
}

\theoremstyle{definition}

\theoremstyle{remark}

\makeatletter
\expandafter\let\expandafter\seed@origfigurestar\csname figure*\endcsname
\expandafter\let\expandafter\seed@endorigfigurestar\csname endfigure*\endcsname
\expandafter\let\expandafter\seed@origtablestar\csname table*\endcsname
\expandafter\let\expandafter\seed@endorigtablestar\csname endtable*\endcsname

\newenvironment{seedfigurewide}[1][t]{%
  \ifseedtwocolumnlayout
    \seed@origfigurestar[#1]%
  \else
    \figure[#1]%
  \fi
}{%
  \ifseedtwocolumnlayout
    \seed@endorigfigurestar
  \else
    \endfigure
  \fi
}

\newenvironment{seedtablewide}[1][t]{%
  \ifseedtwocolumnlayout
    \seed@origtablestar[#1]%
  \else
    \table[#1]%
  \fi
}{%
  \ifseedtwocolumnlayout
    \seed@endorigtablestar
  \else
    \endtable
  \fi
}

\expandafter\let\csname figure*\endcsname\seedfigurewide
\expandafter\let\csname endfigure*\endcsname\endseedfigurewide
\expandafter\let\csname table*\endcsname\seedtablewide
\expandafter\let\csname endtable*\endcsname\endseedtablewide

\newcommand{\seedinputappendix}{%
  \let\seedSavedAppendix\appendix
  \renewcommand{\appendix}{}%
\appendix

\section{Model Configurations}
\label{app:configs}

Tables~\ref{tab:app-configs} and~\ref{tab:app-smelt-configs} list the Baseline and \archname{} configurations in the $4 \times 4$ grid.
The Baseline configurations come from a tuned MoE family.
Scale identifies each matched compute tier by its Baseline active-parameter scale.
The Active params column gives the non-embedding parameter footprint of one physical traversal: attention, normalization, and router weights together with the selected top-8 expert weights.
A shared loop block contributes once to Active params.
The $S$ column gives the measured compute-equivalent sparsity from Eq.~\ref{eq:equivalent-sparsity}, whose FLOPs $F$ count every block execution; the paper refers to the three sparse levels by the integer labels $S \approx \{85\%,95\%,97\%\}$.

\paragraph{Implementation note.}
Our models are built on a proprietary Transformer family whose complete architecture and training stack cannot be released.
We report the quantities directly relevant to the matched comparisons: model scale, hidden dimension, physical depth, expert count, active parameter count, non-embedding parameter count, and compute-equivalent sparsity.
Both architectures are constructed using the same scaling rules and training stack; differences are restricted to the loop configuration and the budget-matching adjustments described in Section~\ref{sec:design-setup}.
For physical depth $L$ and loop-span length $m=L/2$, we place $\lfloor(L-m)/2\rfloor$ layers before the repeated span; with 1-based indexing, the looped layers are $3$--$7$, $4$--$9$, $6$--$15$, and $8$--$22$ for $L=10$, $12$, $20$, and $30$, respectively.
Note that we use different $H$ values at the 600M scale: $H=1408$ for $S=0$ and $S\approx85\%$, and $H=1512$ for $S\approx95\%$ and $97\%$.
This is because budget matching jointly adjusts $H$, the expert intermediate dimension, and attention-head geometry to obtain the closest match in per-token FLOPs, total parameters, and KV cache.

\begin{center}
\begin{minipage}{\linewidth}
\small
\captionof{table}{
Baseline configurations for all 16 grid cells.
$H$ is the hidden dimension and $L$ the physical depth.
Total experts and Active experts give the stored pool size and the number selected by each routing decision, respectively.
Parameter counts exclude embeddings.
}
\label{tab:app-configs}
\centering
\begin{tabular}{lrrrrrrr}
\toprule
Scale & $H$ & $L$ & \shortstack{Total\\experts} & \shortstack{Active\\experts} & \shortstack{Active\\params} & $S$ & Total params \\
\midrule
100M & 672 & 10 & 8   & 8 & 0.100B & 0.0\% & 0.10B \\
100M & 672 & 10 & 64  & 8 & 0.100B & 86.0\% & 0.71B \\
100M & 672 & 10 & 192 & 8 & 0.101B & 95.2\% & 2.12B \\
100M & 672 & 10 & 336 & 8 & 0.102B & 97.2\% & 3.70B \\
\midrule
200M & 1280 & 12 & 8   & 8 & 0.207B & 0.0\% & 0.21B \\
200M & 1280 & 12 & 64  & 8 & 0.207B & 84.3\% & 1.32B \\
200M & 1280 & 12 & 192 & 8 & 0.209B & 94.6\% & 3.87B \\
200M & 1280 & 12 & 336 & 8 & 0.212B & 96.9\% & 6.74B \\
\midrule
600M & 1792 & 20 & 8   & 8 & 0.663B & 0.0\% & 0.66B \\
600M & 1792 & 20 & 64  & 8 & 0.665B & 84.5\% & 4.28B \\
600M & 1792 & 20 & 192 & 8 & 0.670B & 94.7\% & 12.54B \\
600M & 1792 & 20 & 336 & 8 & 0.675B & 96.9\% & 21.83B \\
\midrule
1.6B & 2304 & 30 & 8   & 8 & 1.629B & 0.0\% & 1.63B \\
1.6B & 2304 & 30 & 64  & 8 & 1.632B & 84.5\% & 10.55B \\
1.6B & 2304 & 30 & 192 & 8 & 1.641B & 94.7\% & 30.94B \\
1.6B & 2304 & 30 & 336 & 8 & 1.651B & 96.9\% & 53.89B \\
\bottomrule
\end{tabular}
\end{minipage}
\end{center}

\begin{center}
\begin{minipage}{\linewidth}
\small
\captionof{table}{
\archname{} configurations for all 16 grid cells.
$H$ is the hidden dimension and $L$ the physical depth.
Total experts and Active experts give the stored pool size and the number selected by each routing decision, respectively.
Loop gives the repeated span and total number of passes.
Parameter counts exclude embeddings.
The final two columns give the per-token training-FLOPs and KV-cache ratios relative to the corresponding Baseline.
}
\label{tab:app-smelt-configs}
\centering
\scriptsize
\setlength{\tabcolsep}{2.5pt}
\begin{tabular}{lrrrrlrrrrr}
\toprule
Scale & $H$ & $L$ & \shortstack{Total\\experts} & \shortstack{Active\\experts} & \shortstack{Loop\\Configuration} & \shortstack{Active\\params} & $S$ & Total params & $F_{\mathrm{SMELT}}/F_{\mathrm{Base}}$ & $\mathrm{KV}_{\mathrm{SMELT}}/\mathrm{KV}_{\mathrm{Base}}$ \\
\midrule
100M & 576  & 10 & 16  & 8 & Mid. 50\% $\times 2$ & 0.071B & 0.0\% & 0.10B & 1.081 & 1.018 \\
100M & 576  & 10 & 96  & 8 & Mid. 50\% $\times 2$ & 0.067B & 86.8\% & 0.71B & 1.024 & 0.964 \\
100M & 576  & 10 & 288 & 8 & Mid. 50\% $\times 2$ & 0.069B & 95.5\% & 2.12B & 1.031 & 0.964 \\
100M & 576  & 10 & 504 & 8 & Mid. 50\% $\times 2$ & 0.070B & 97.4\% & 3.70B & 1.039 & 0.964 \\
\midrule
200M & 1056 & 12 & 16  & 8 & Mid. 50\% $\times 2$ & 0.142B & 0.0\% & 0.21B & 1.039 & 1.031 \\
200M & 1056 & 12 & 96  & 8 & Mid. 50\% $\times 2$ & 0.140B & 84.6\% & 1.32B & 1.021 & 1.031 \\
200M & 1056 & 12 & 288 & 8 & Mid. 50\% $\times 2$ & 0.142B & 94.7\% & 3.89B & 1.029 & 1.031 \\
200M & 1056 & 12 & 504 & 8 & Mid. 50\% $\times 2$ & 0.145B & 96.9\% & 6.78B & 1.037 & 1.031 \\
\midrule
600M & 1408 & 20 & 16  & 8 & Mid. 50\% $\times 2$ & 0.458B & 0.0\% & 0.67B & 1.046 & 1.031 \\
600M & 1408 & 20 & 96  & 8 & Mid. 50\% $\times 2$ & 0.443B & 85.1\% & 4.31B & 1.007 & 1.031 \\
600M & 1512 & 20 & 288 & 8 & Mid. 50\% $\times 2$ & 0.453B & 94.7\% & 12.44B & 1.024 & 0.984 \\
600M & 1512 & 20 & 504 & 8 & Mid. 50\% $\times 2$ & 0.459B & 96.9\% & 21.70B & 1.031 & 0.984 \\
\midrule
1.6B & 1848 & 30 & 16  & 8 & Mid. 50\% $\times 2$ & 1.101B & 0.0\% & 1.64B & 1.022 & 0.984 \\
1.6B & 1848 & 30 & 96  & 8 & Mid. 50\% $\times 2$ & 1.081B & 84.6\% & 10.45B & 1.000 & 0.984 \\
1.6B & 1848 & 30 & 288 & 8 & Mid. 50\% $\times 2$ & 1.091B & 94.8\% & 30.90B & 1.004 & 0.984 \\
1.6B & 1848 & 30 & 504 & 8 & Mid. 50\% $\times 2$ & 1.103B & 97.0\% & 53.90B & 1.008 & 0.984 \\
\bottomrule
\end{tabular}
\end{minipage}
\end{center}

\paragraph{Matching granularity.}
The residual mismatch is larger at the least sparse reference level, where the expert pool is small and each hardware-aligned change to model width, expert intermediate dimension, head geometry, or expert count moves a comparatively large fraction of the budget.
The feasible configurations therefore form a coarser grid than at the three sparse levels.
Across the 12 sparse cells used in that fit, the maximum absolute mismatches are $3.9\%$ in per-token FLOPs, $1.0\%$ in total parameters, and $3.6\%$ in KV cache.
Besides, these residual differences are not treated as exact budget equality in the scaling analysis: every run enters the fit at its measured per-token FLOPs $F$, and the reported compute-efficiency comparisons evaluate the two separately fitted surfaces at common $(C,S)$ coordinates.

\FloatBarrier

\section{Training Curves for Every Grid Cell}
\label{app:training-curves}

Section~\ref{sec:main-grid} shows training curves for two representative cells.
This appendix gives all 16 cells of the $4 \times 4$ grid: four scales crossed with the $S{=}0\%$ dense-reference control and three sparse levels $S \approx \{85\%,95\%,97\%\}$.
Each panel plots training loss against cumulative training tokens for one matched Baseline / \archname{} pair.
Solid lines are the stable constant-learning-rate phase; dashed lines are the six cosine-decay branches forked at steps 10{,}000, 20{,}000, 50{,}000, 100{,}000, 150{,}000, and 196{,}075.
\archname{} ends below the Baseline in all 16 cells.

\begin{figure}[H]
\centering
\begin{subfigure}[t]{0.48\linewidth}
\centering
\includegraphics[width=\linewidth]{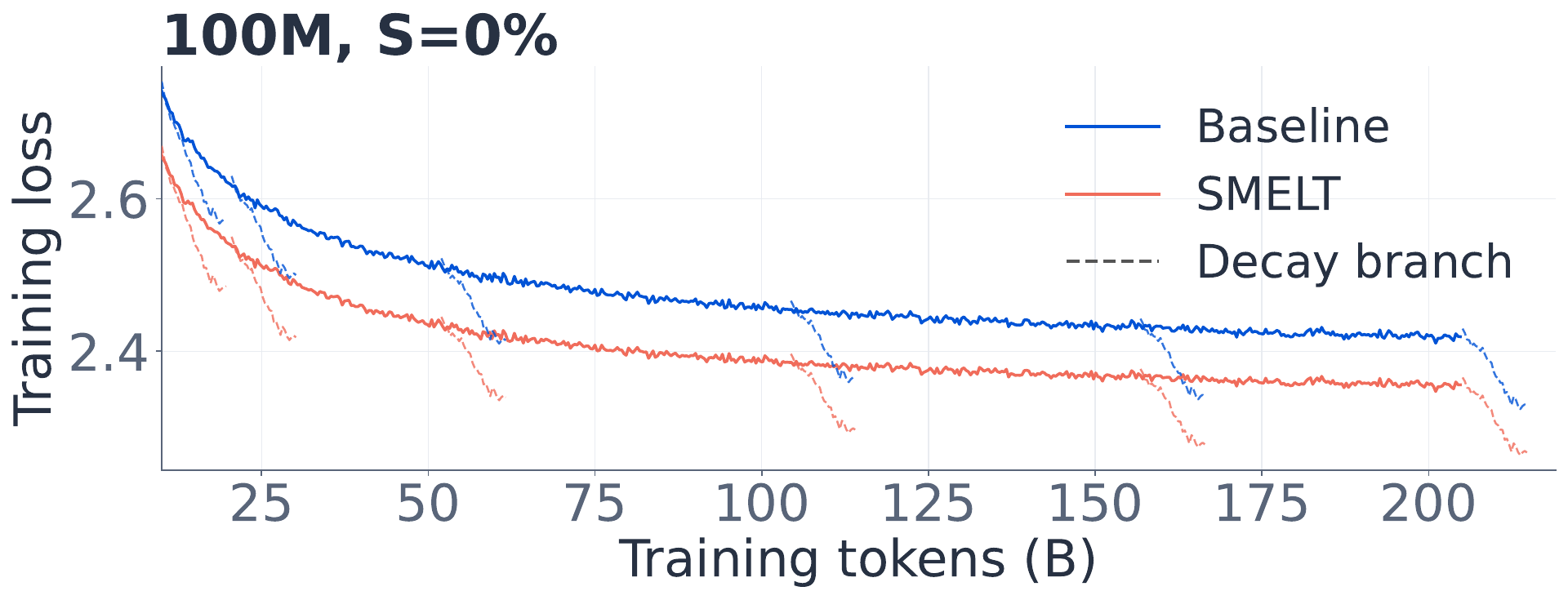}
\caption{$S = 0\%$ (dense-reference level).}
\end{subfigure}\hfill
\begin{subfigure}[t]{0.48\linewidth}
\centering
\includegraphics[width=\linewidth]{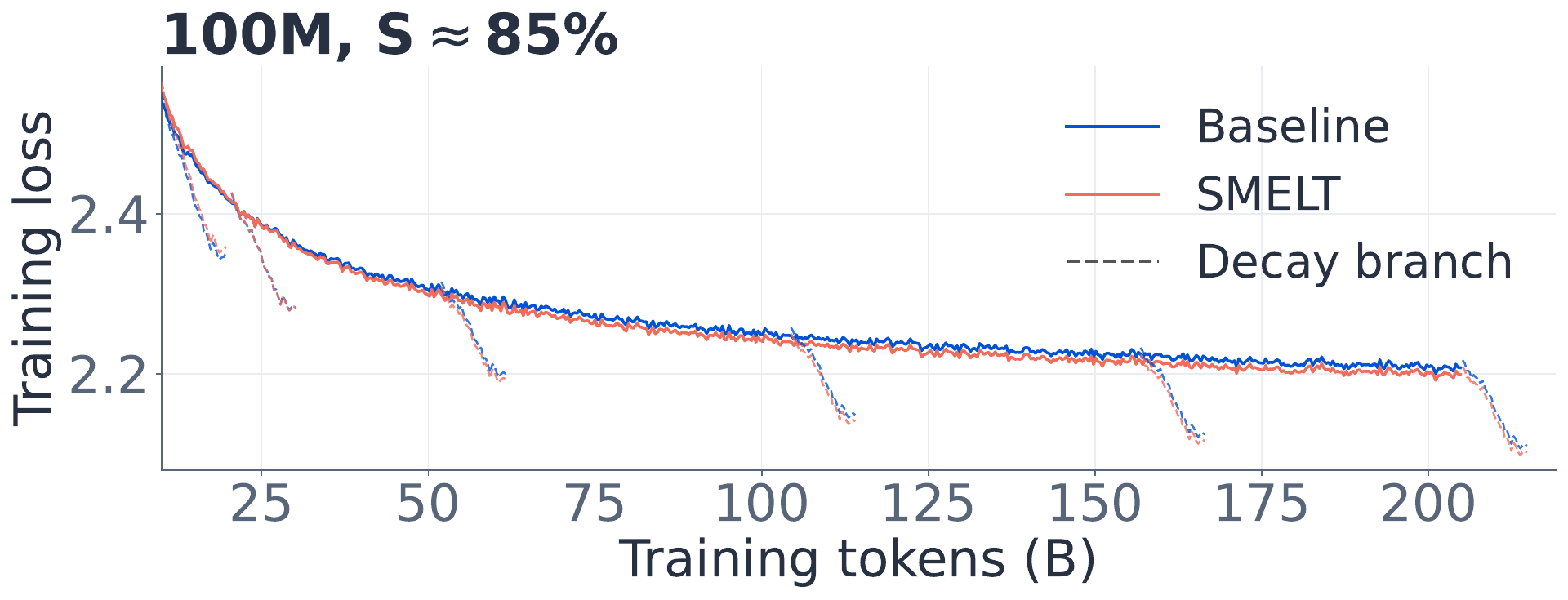}
\caption{$S \approx 85\%$.}
\end{subfigure}

\vspace{0.6em}

\begin{subfigure}[t]{0.48\linewidth}
\centering
\includegraphics[width=\linewidth]{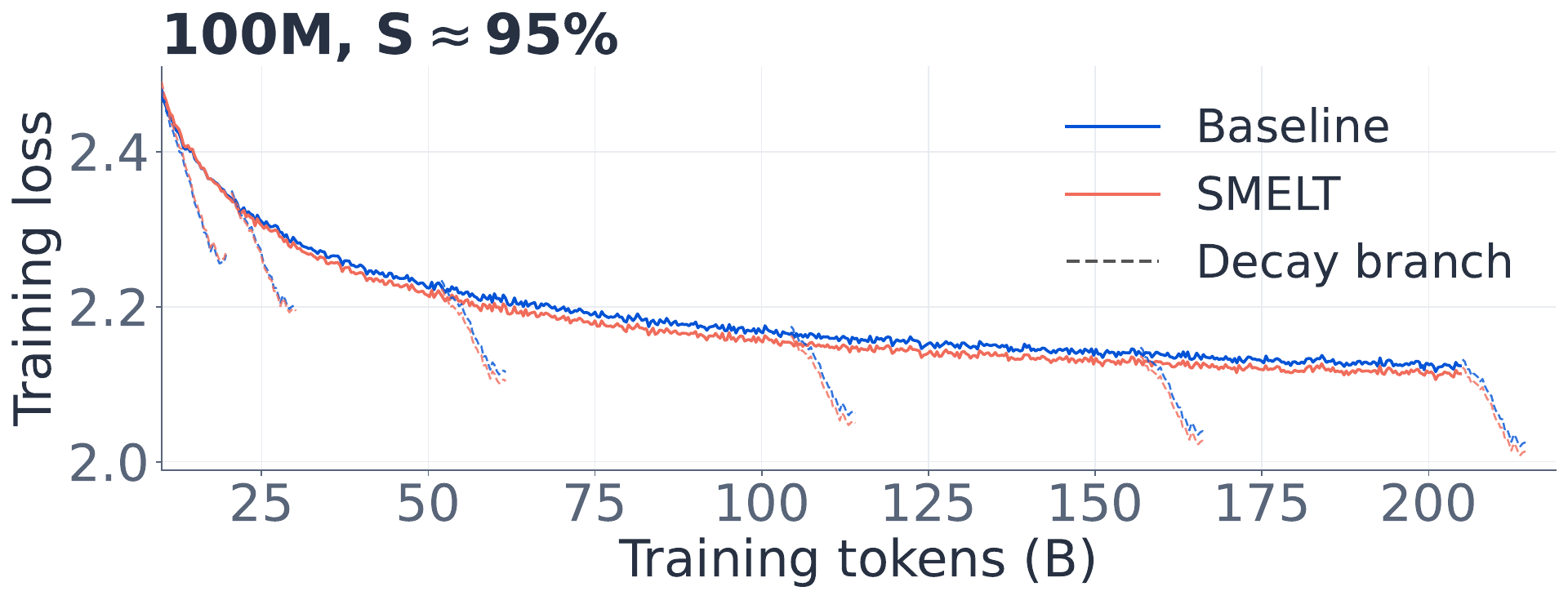}
\caption{$S \approx 95\%$.}
\end{subfigure}\hfill
\begin{subfigure}[t]{0.48\linewidth}
\centering
\includegraphics[width=\linewidth]{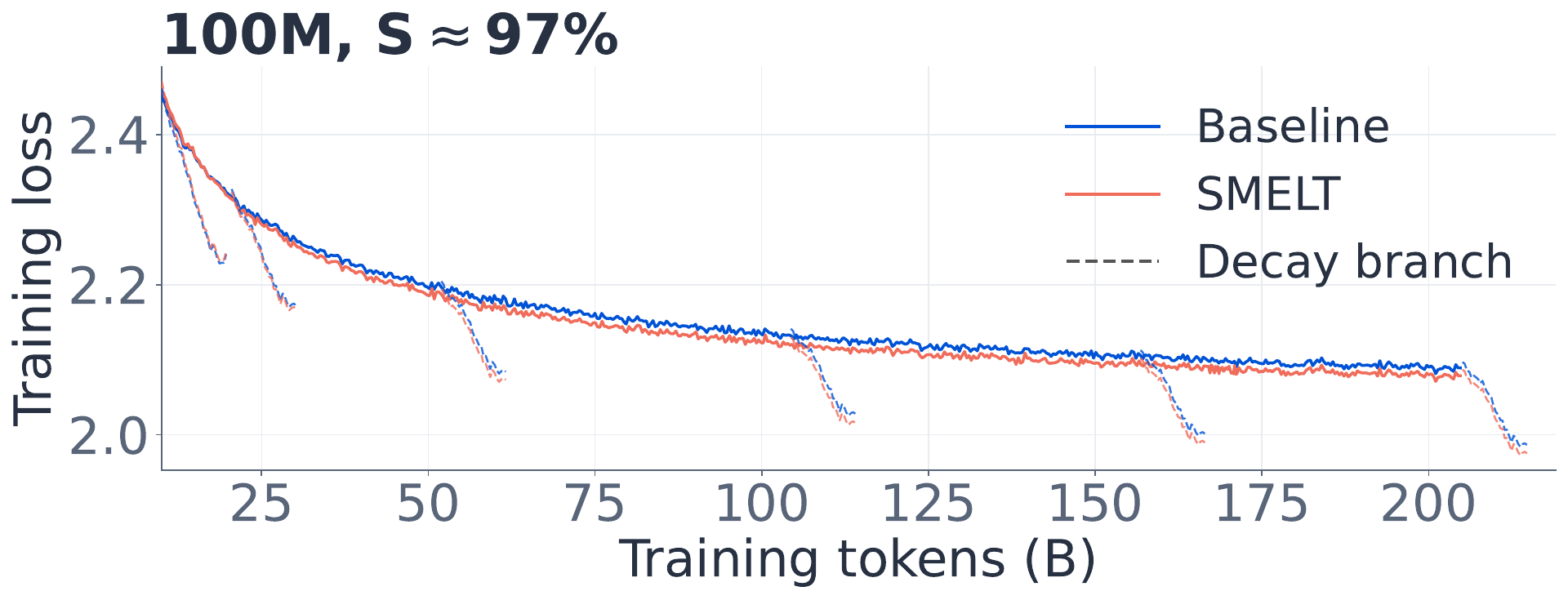}
\caption{$S \approx 97\%$.}
\end{subfigure}
\caption{
Training loss at the 100M scale across all four sparsity levels.
The Baseline is shown in blue and \archname{} in red.
}
\label{fig:app-curves-100m}
\end{figure}

\FloatBarrier

\begin{figure}[H]
\centering
\begin{subfigure}[t]{0.48\linewidth}
\centering
\includegraphics[width=\linewidth]{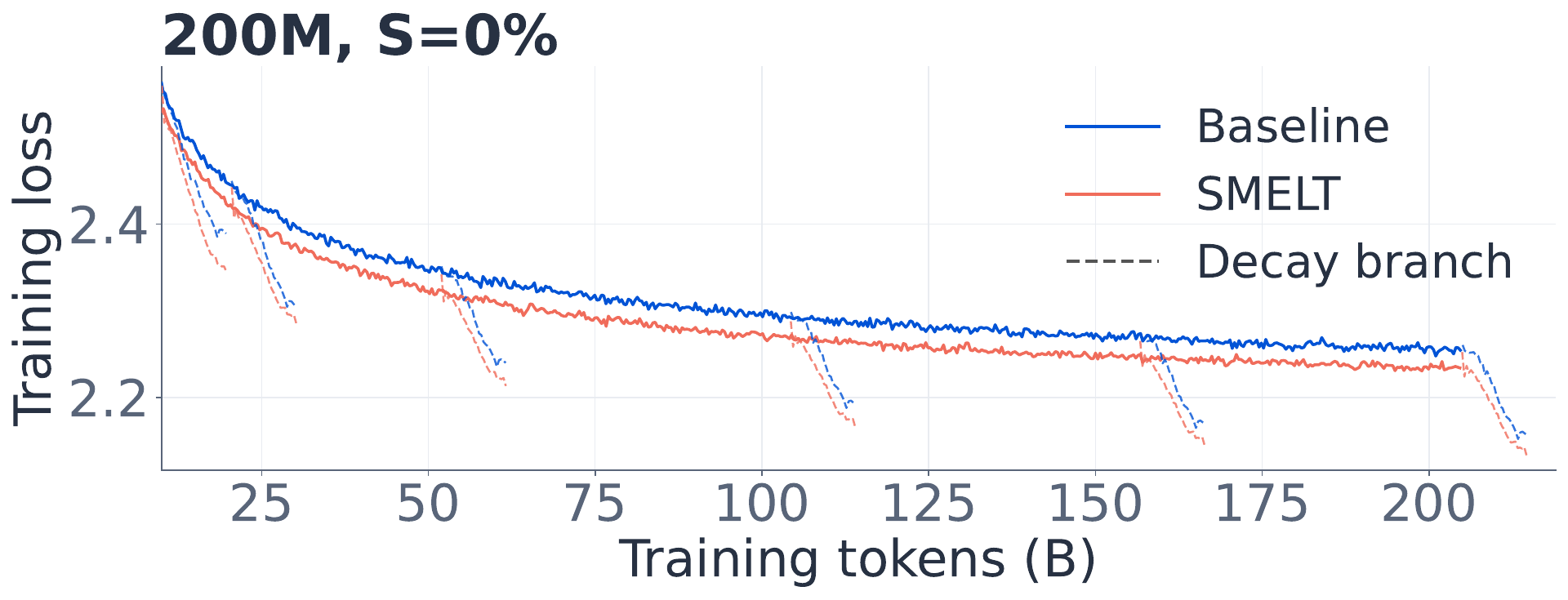}
\caption{$S = 0\%$ (dense-reference level).}
\end{subfigure}\hfill
\begin{subfigure}[t]{0.48\linewidth}
\centering
\includegraphics[width=\linewidth]{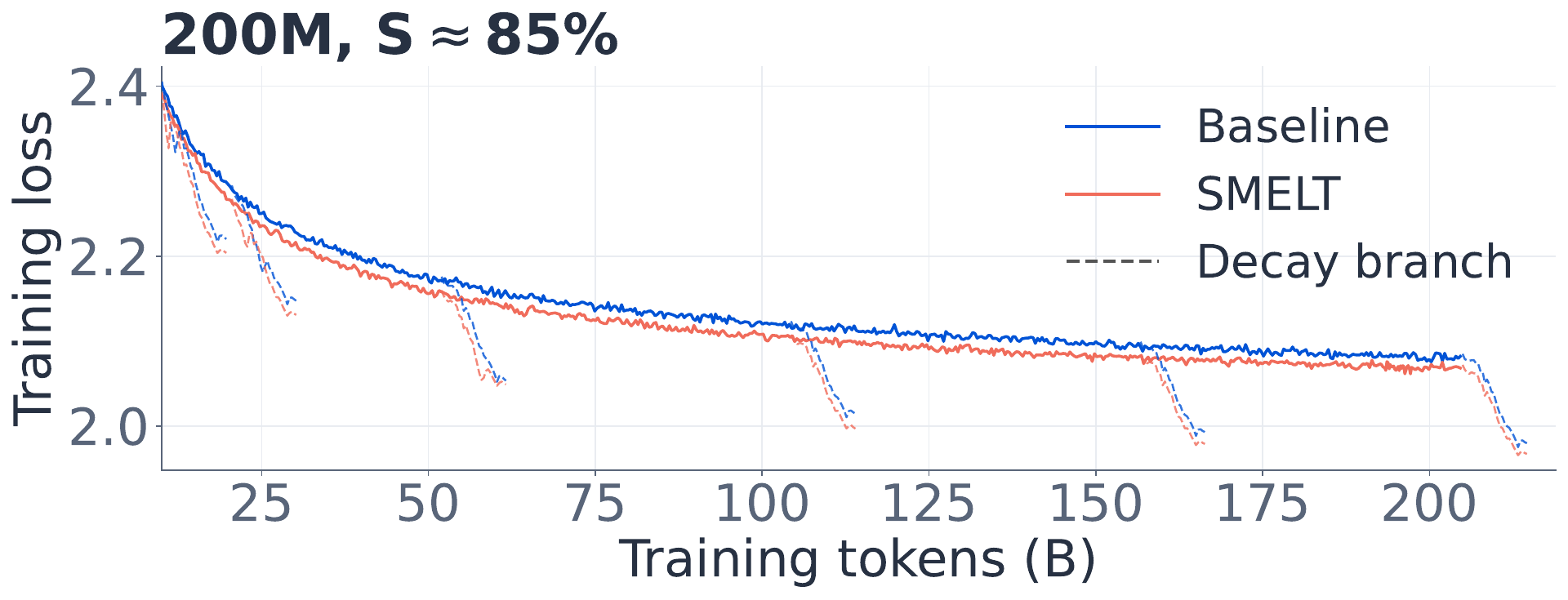}
\caption{$S \approx 85\%$.}
\end{subfigure}

\vspace{0.6em}

\begin{subfigure}[t]{0.48\linewidth}
\centering
\includegraphics[width=\linewidth]{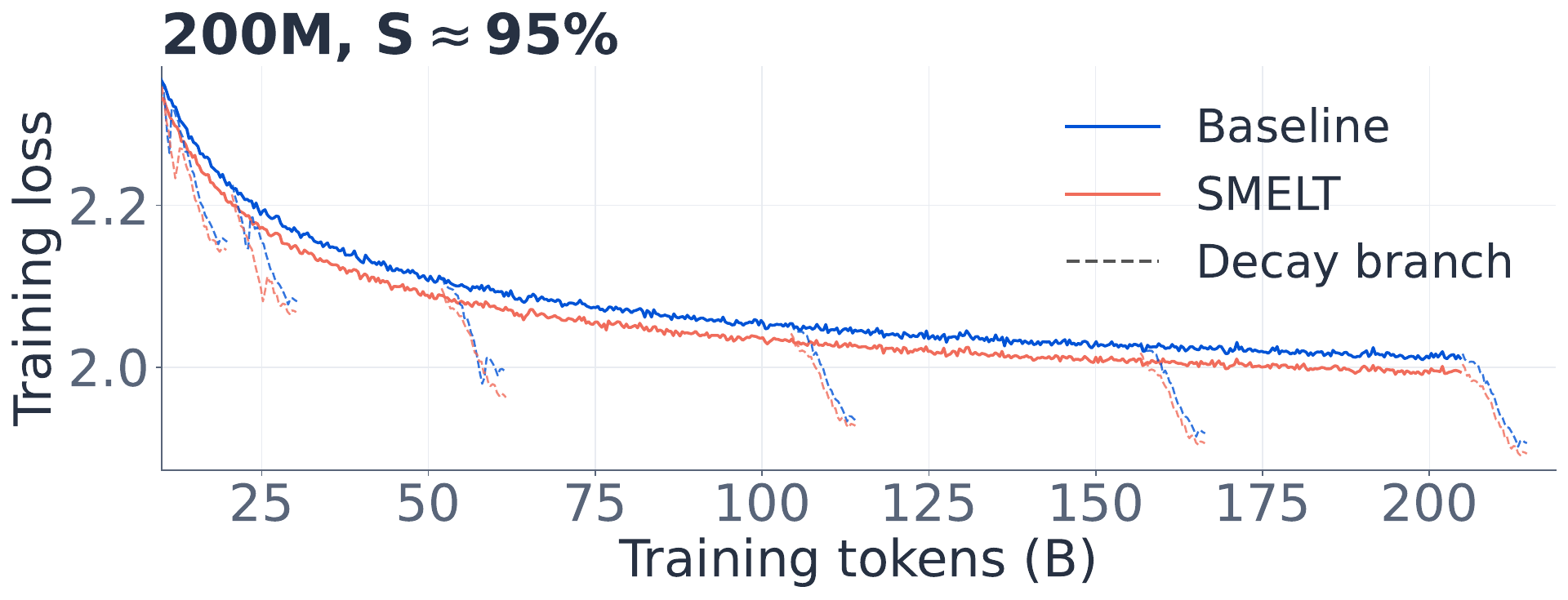}
\caption{$S \approx 95\%$.}
\end{subfigure}\hfill
\begin{subfigure}[t]{0.48\linewidth}
\centering
\includegraphics[width=\linewidth]{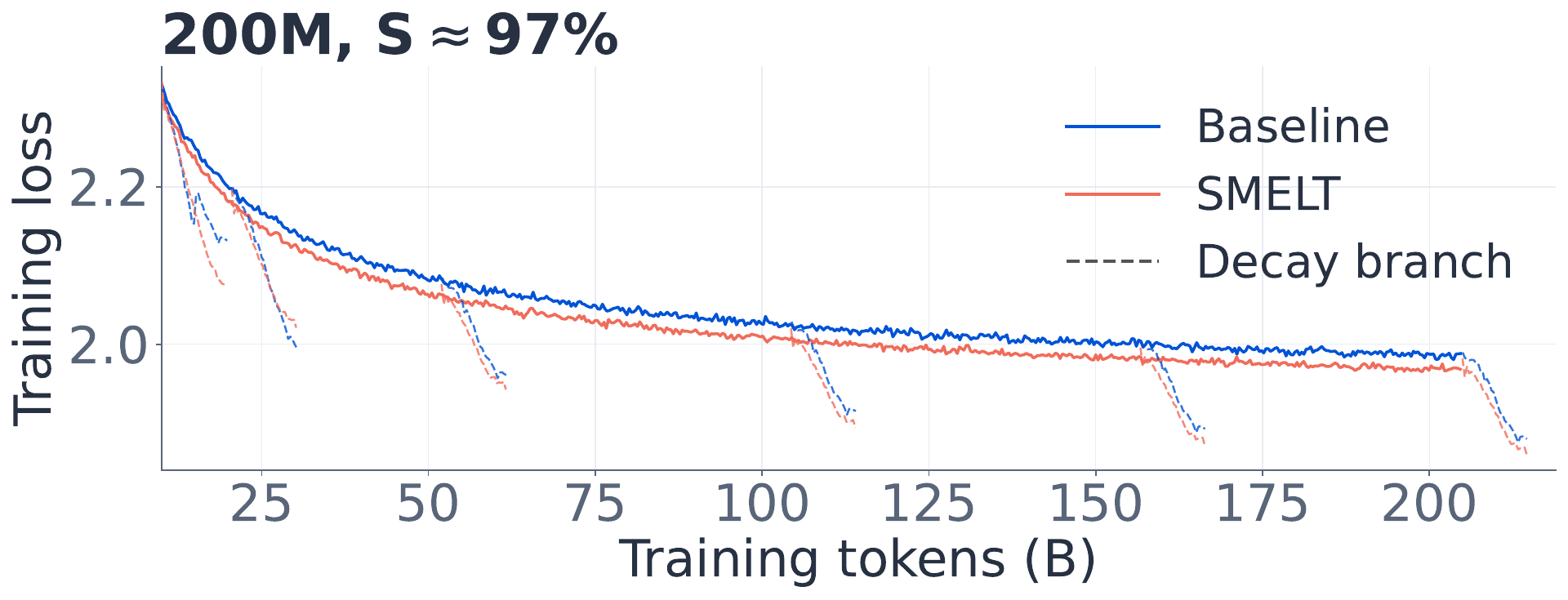}
\caption{$S \approx 97\%$.}
\end{subfigure}
\caption{
Training loss at the 200M scale across all four sparsity levels.
The Baseline is shown in blue and \archname{} in red.
}
\label{fig:app-curves-200m}
\end{figure}

\FloatBarrier

\begin{figure}[H]
\centering
\begin{subfigure}[t]{0.48\linewidth}
\centering
\includegraphics[width=\linewidth]{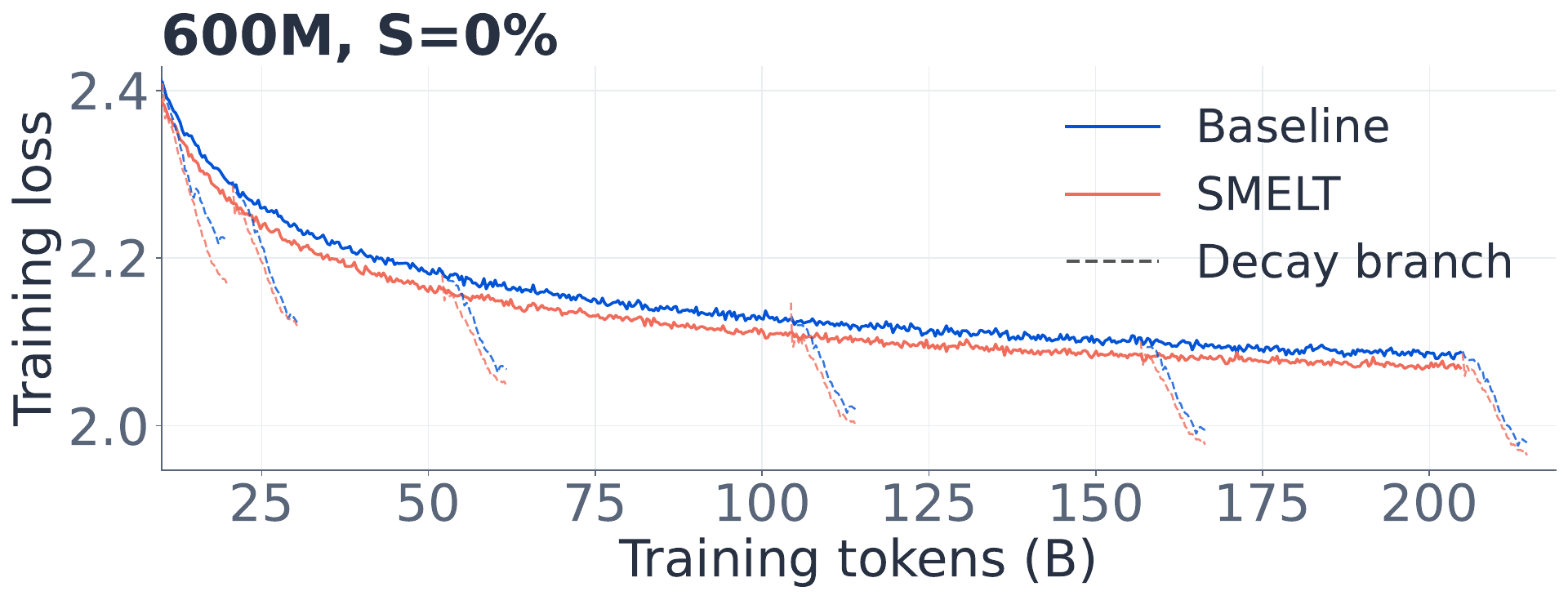}
\caption{$S = 0\%$ (dense-reference level).}
\end{subfigure}\hfill
\begin{subfigure}[t]{0.48\linewidth}
\centering
\includegraphics[width=\linewidth]{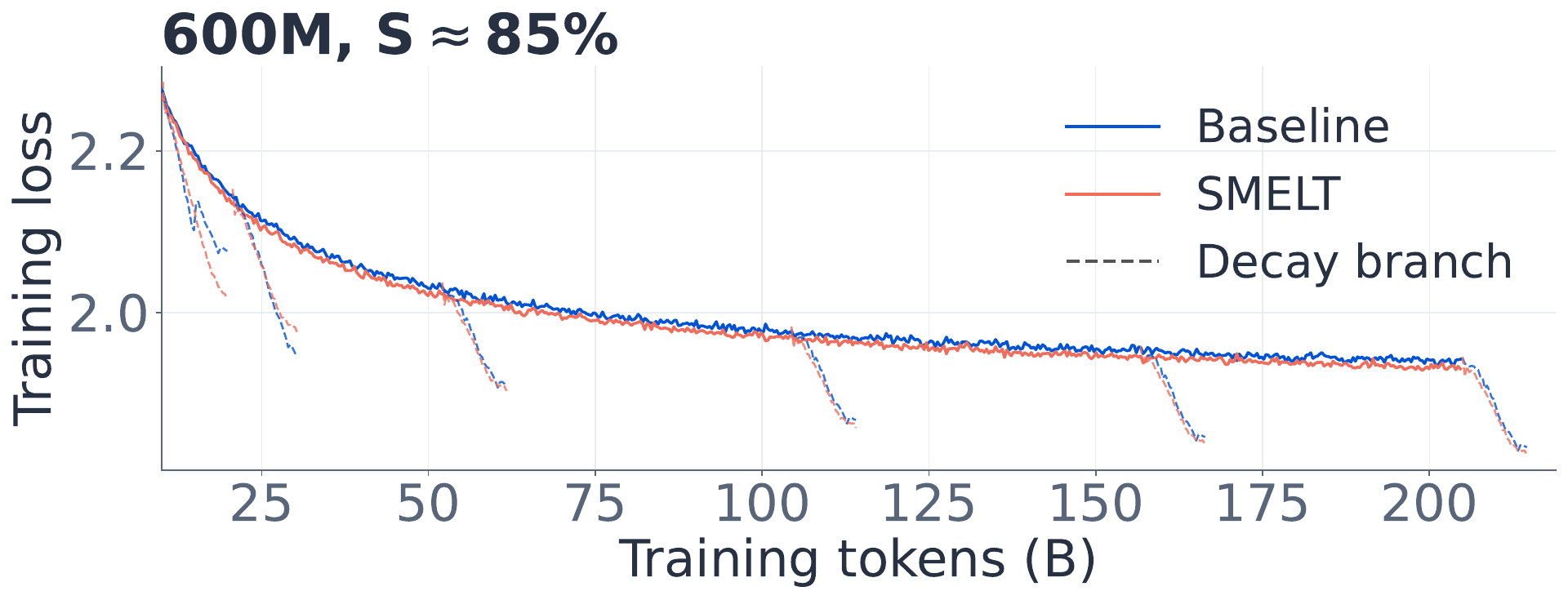}
\caption{$S \approx 85\%$.}
\end{subfigure}

\vspace{0.6em}

\begin{subfigure}[t]{0.48\linewidth}
\centering
\includegraphics[width=\linewidth]{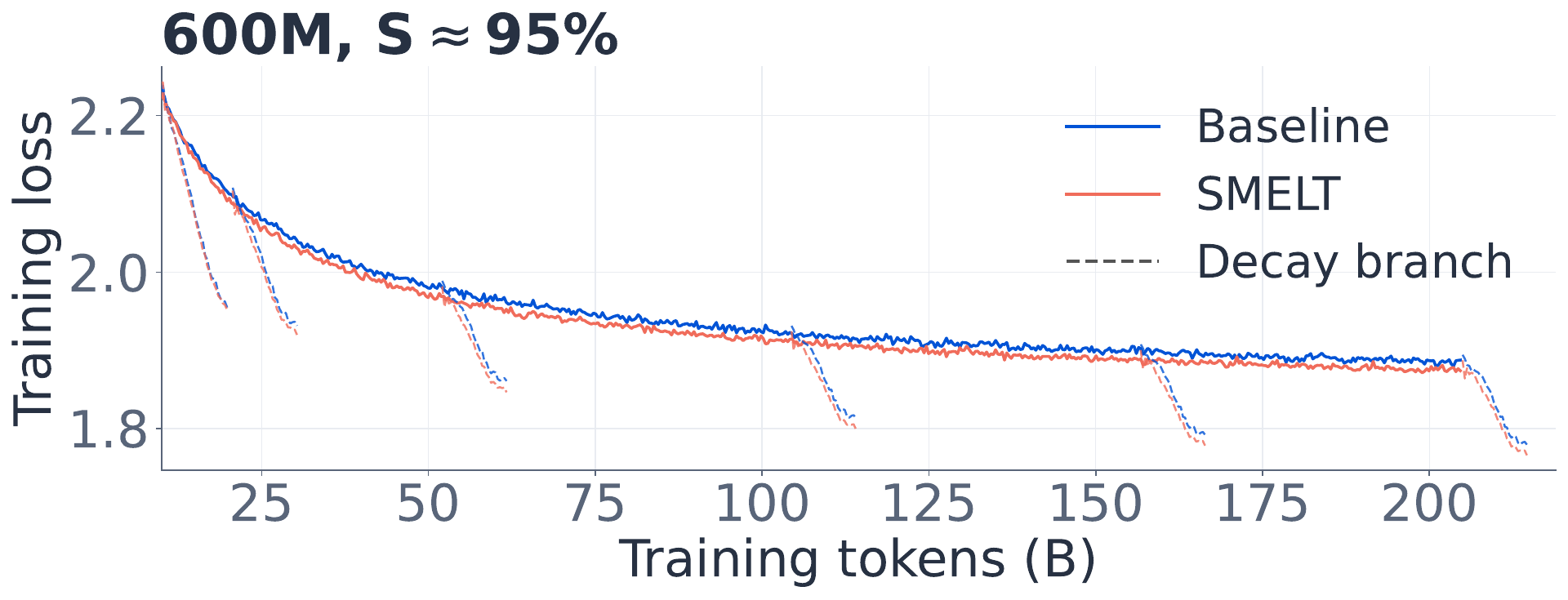}
\caption{$S \approx 95\%$.}
\end{subfigure}\hfill
\begin{subfigure}[t]{0.48\linewidth}
\centering
\includegraphics[width=\linewidth]{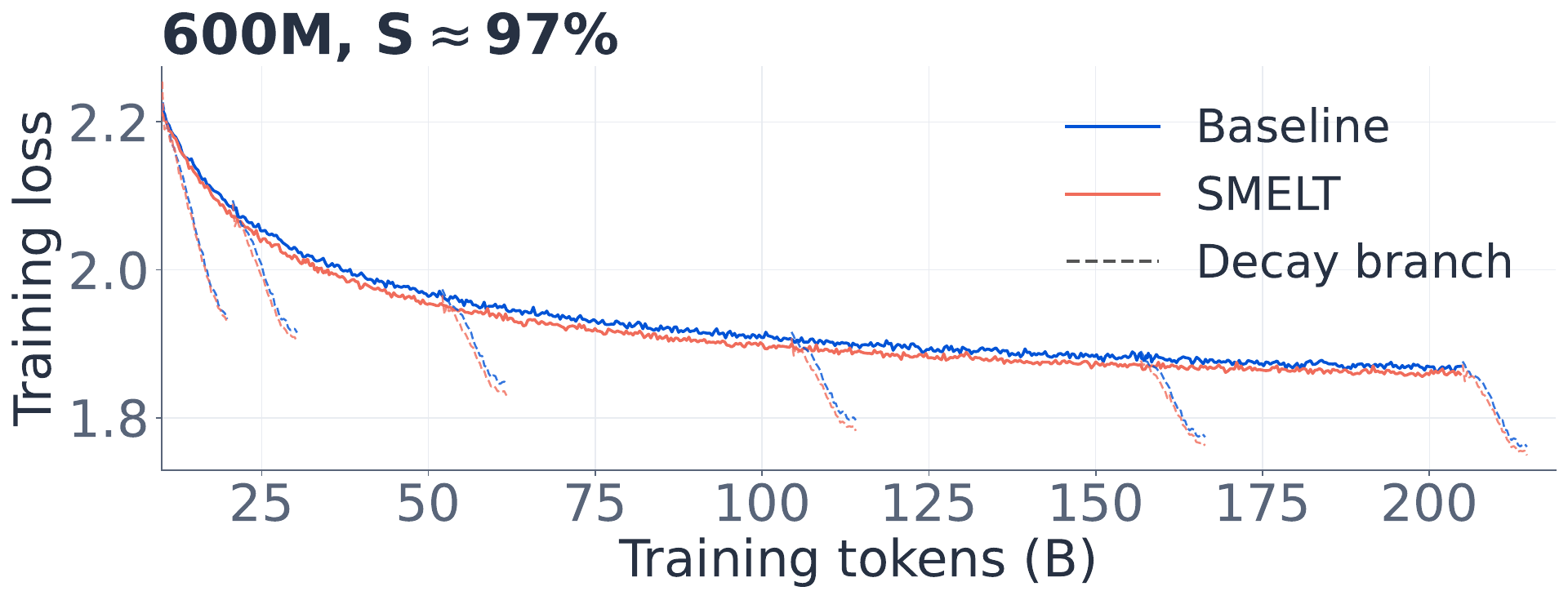}
\caption{$S \approx 97\%$ (also Figure~\ref{fig:training-curve}a).}
\end{subfigure}
\caption{
Training loss at the 600M scale across all four sparsity levels.
The Baseline is shown in blue and \archname{} in red.
}
\label{fig:app-curves-600m}
\end{figure}

\FloatBarrier

\begin{figure}[H]
\centering
\begin{subfigure}[t]{0.48\linewidth}
\centering
\includegraphics[width=\linewidth]{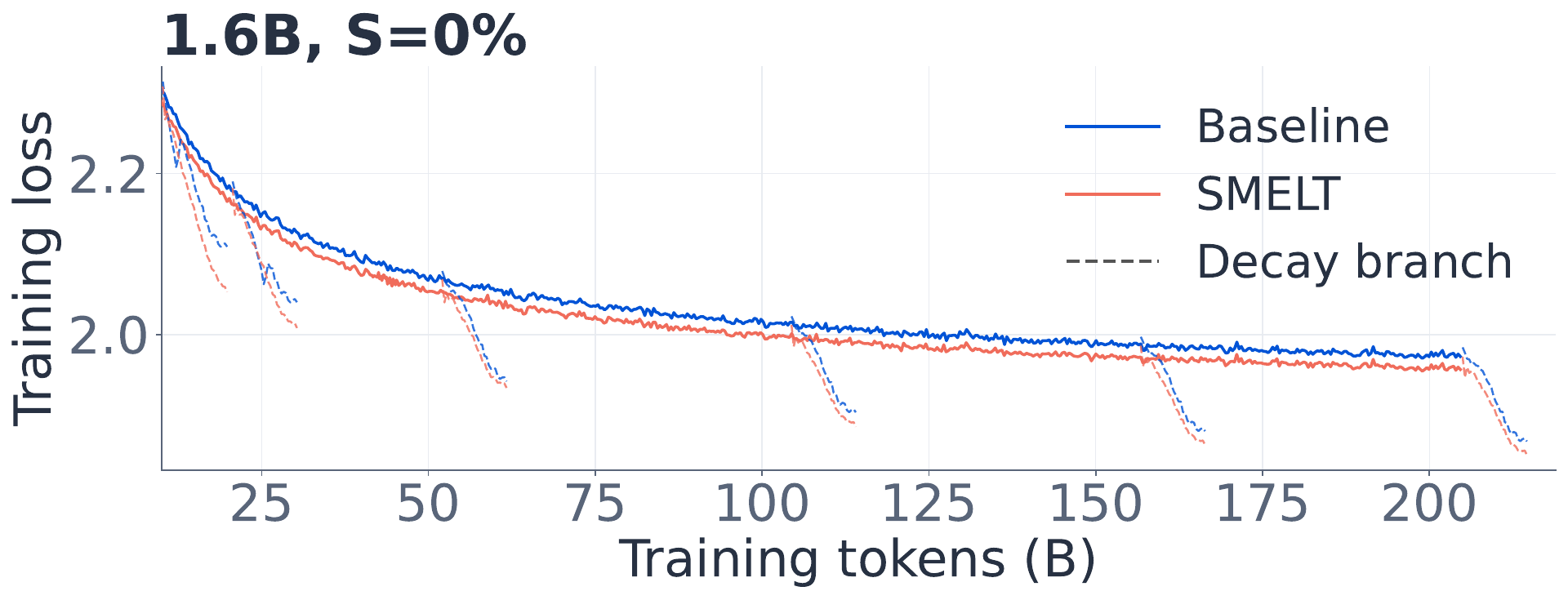}
\caption{$S = 0\%$ (dense-reference level).}
\end{subfigure}\hfill
\begin{subfigure}[t]{0.48\linewidth}
\centering
\includegraphics[width=\linewidth]{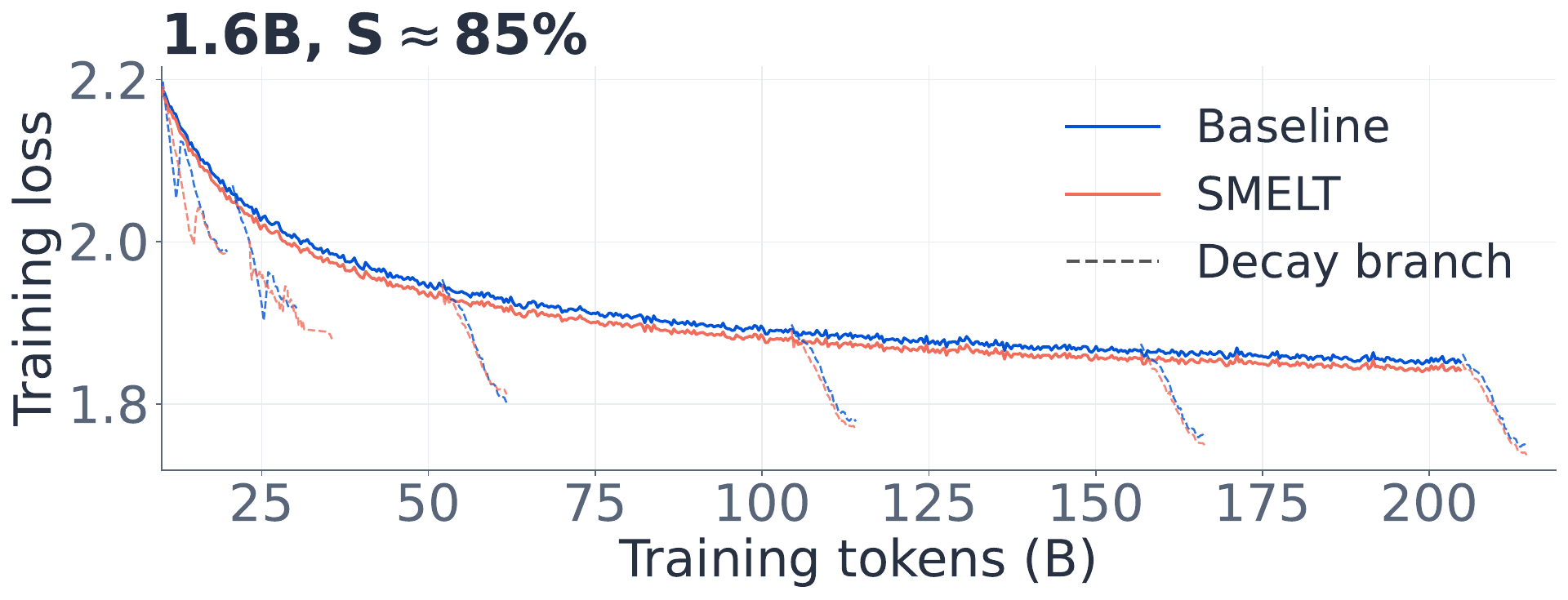}
\caption{$S \approx 85\%$.}
\end{subfigure}

\vspace{0.6em}

\begin{subfigure}[t]{0.48\linewidth}
\centering
\includegraphics[width=\linewidth]{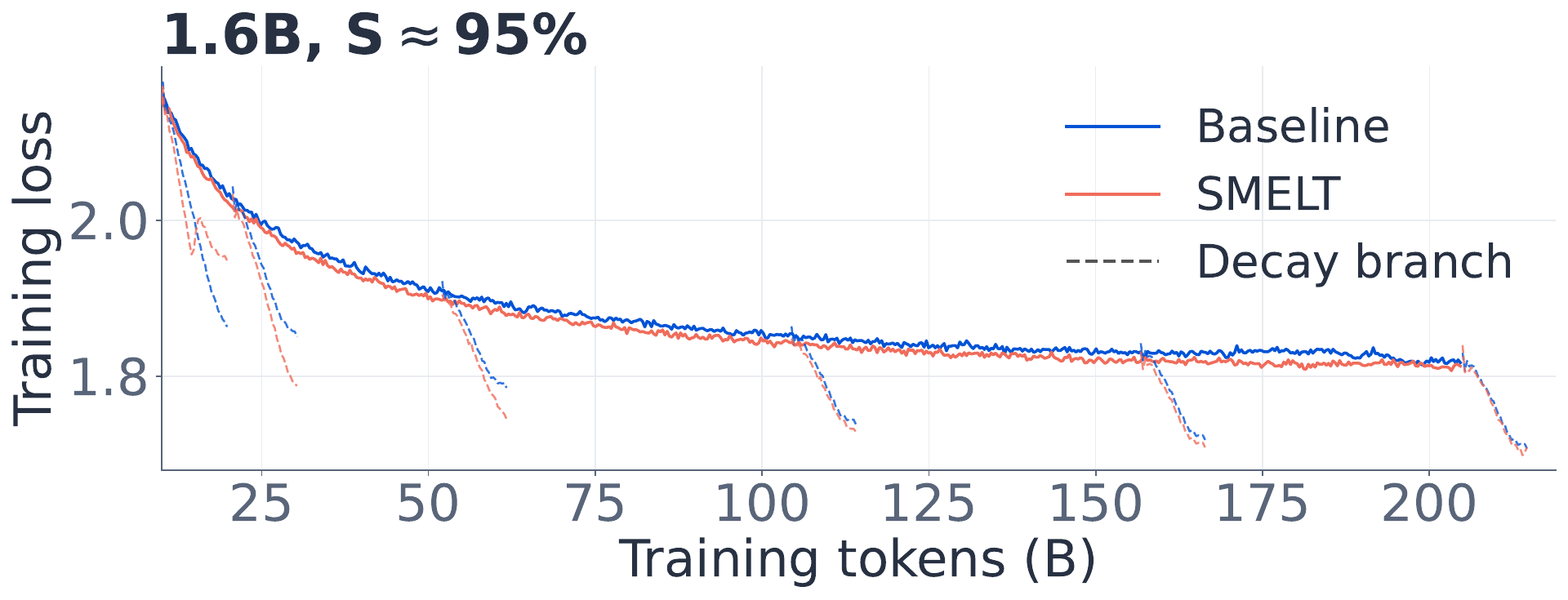}
\caption{$S \approx 95\%$.}
\end{subfigure}\hfill
\begin{subfigure}[t]{0.48\linewidth}
\centering
\includegraphics[width=\linewidth]{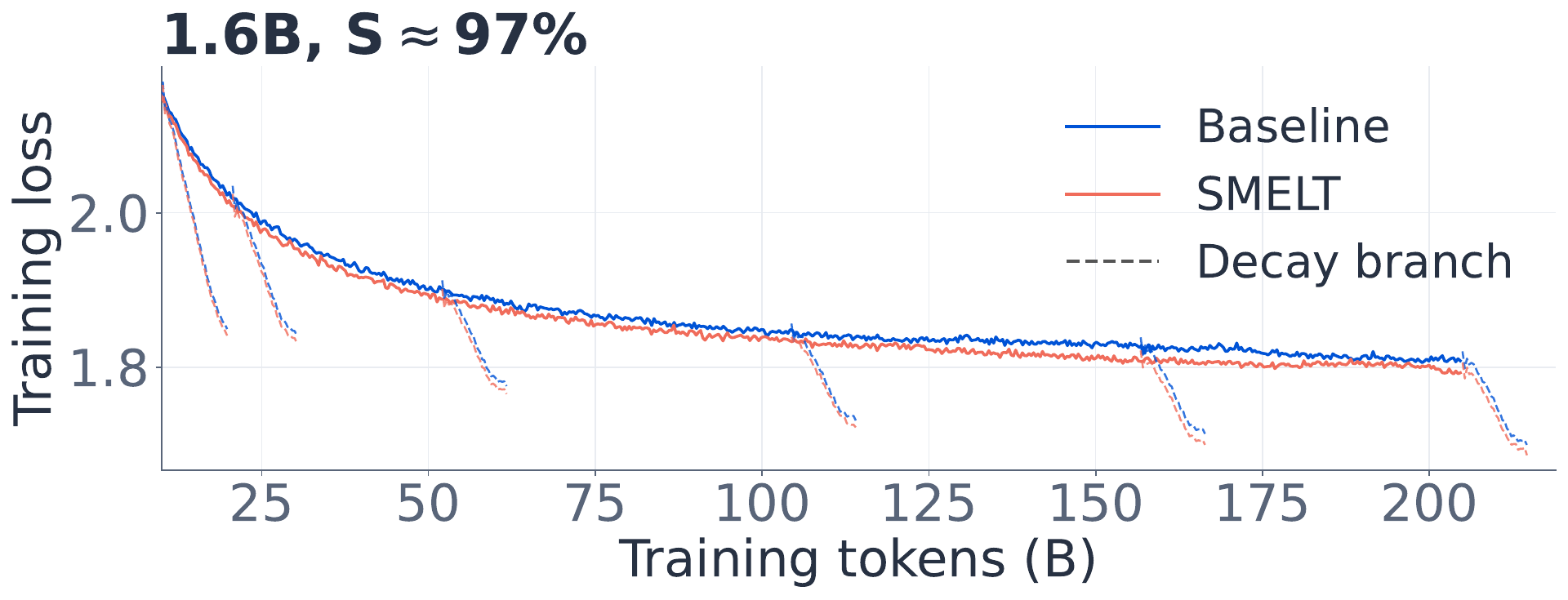}
\caption{$S \approx 97\%$ (also Figure~\ref{fig:training-curve}b).}
\end{subfigure}
\caption{
Training loss at the 1.6B scale across all four sparsity levels.
The Baseline is shown in blue and \archname{} in red.
}
\label{fig:app-curves-1p6b}
\end{figure}

\FloatBarrier

\section{Per-Task ICL Accuracy Curves}
\label{app:icl-tasks}

The shot sweep of Section~\ref{sec:length-icl} covers 16 few-shot tasks: the 15
DCLM Core tasks that take in-context demonstrations, plus MMLU.
Figure~\ref{fig:app-icl-per-task} shows accuracy as a function of the number
of in-context examples $k$ for the 14 of those 16 tasks where \archname{}'s
accuracy gain at the maximum shot count is positive with $p < 0.05$ under a
paired permutation test over the 10 random seeds.
Each curve averages first within each (scale, sparsity) grid cell where both
architectures score above chance, then across cells, so that no single scale
dominates.

\begin{figure}[H]
\centering
\includegraphics[width=\linewidth]{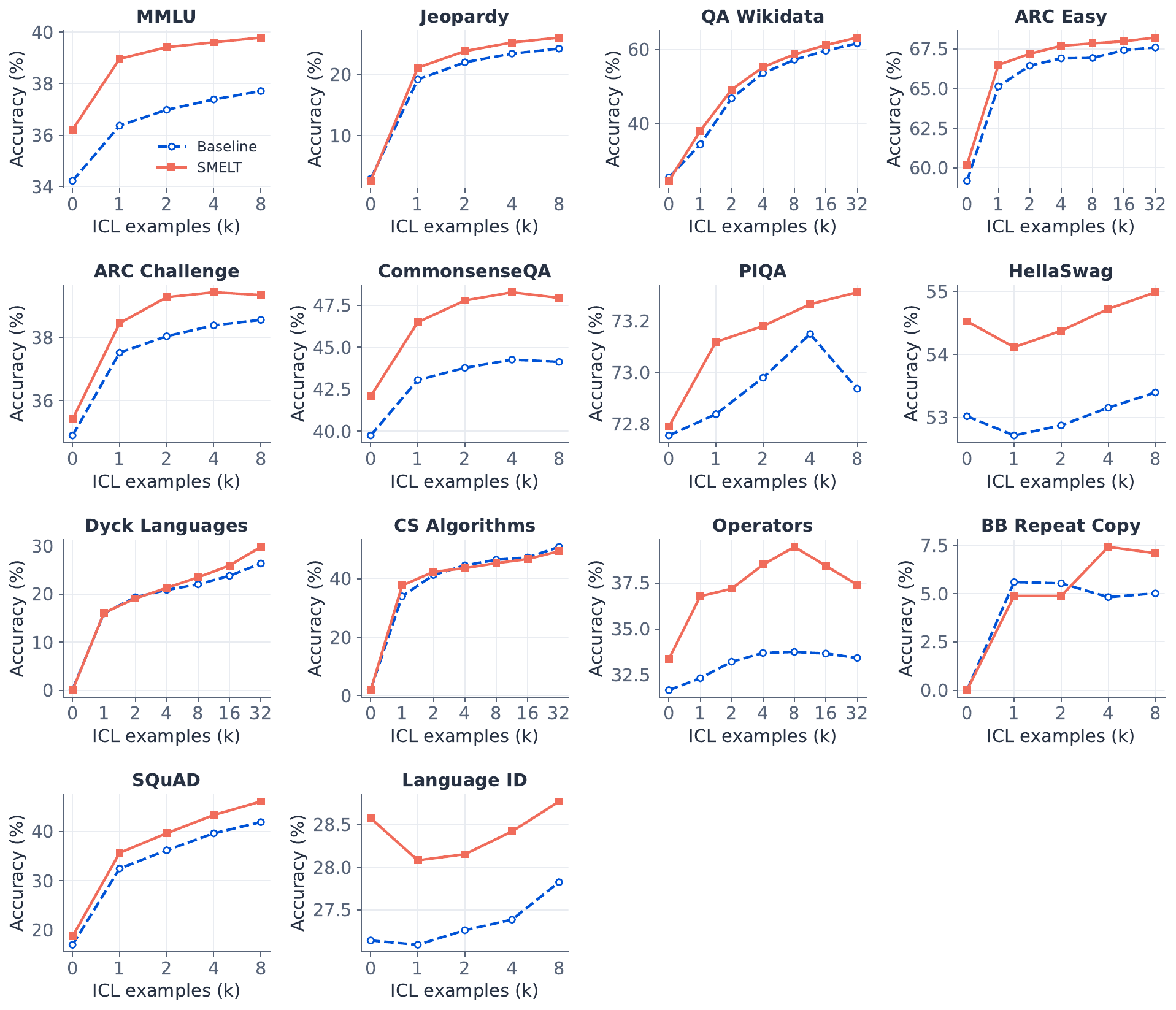}
\caption{Per-task accuracy vs.\ number of in-context examples $k$ for the
  14 of 16 tasks with a significant max-shot gain ($p < 0.05$, paired permutation test).  Baseline (dashed, blue) and
  \archname{} (solid, red).  Each curve averages first within each grid cell, then across cells.}
\label{fig:app-icl-per-task}
\end{figure}

\FloatBarrier

\section{Attention-Sink Profiles Across Scales}
\label{app:attention-sink-scales}

Figure~\ref{fig:app-attention-sink-scales} expands the 1.6B comparison in
Figure~\ref{fig:sink-depth} to all four model scales.  At each scale, the
Baseline and \archname{} have the same physical depth.  The Baseline profile
generally rises toward later layers, whereas \archname{}'s visit~2 remains
below visit~1 throughout the repeated block.

\begin{figure}[H]
\centering
\begin{subfigure}[t]{0.375\linewidth}
\centering
\includegraphics[width=\linewidth]{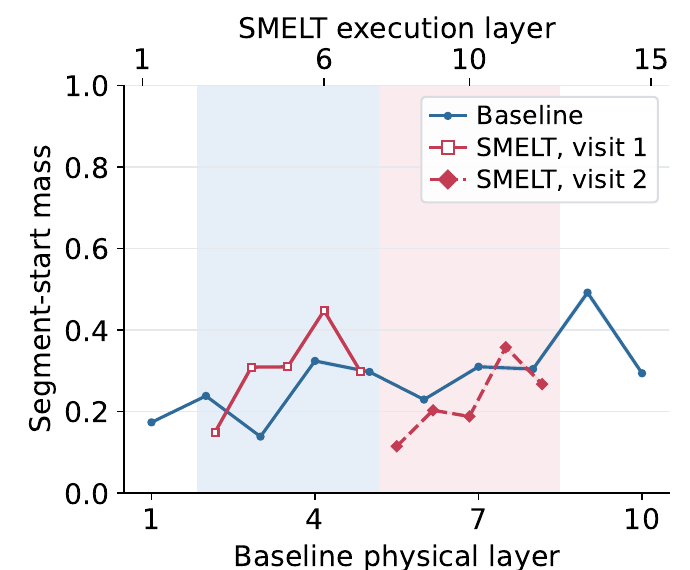}
\caption{100M.}
\end{subfigure}\hspace{0.04\linewidth}
\begin{subfigure}[t]{0.375\linewidth}
\centering
\includegraphics[width=\linewidth]{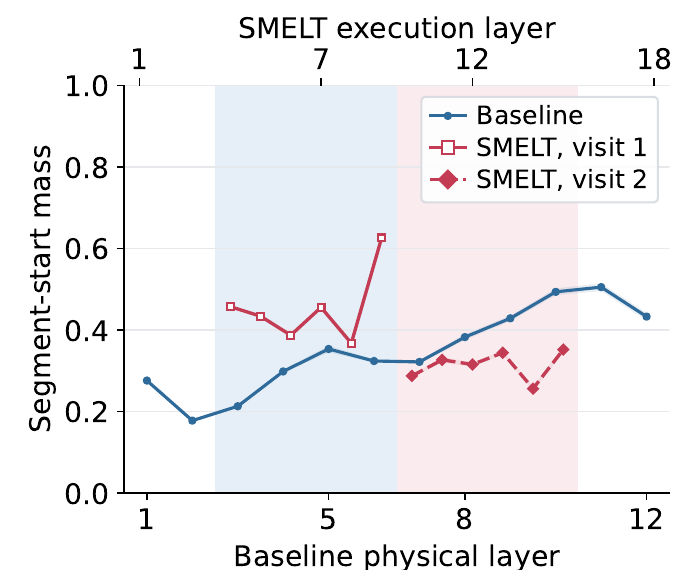}
\caption{200M.}
\end{subfigure}

\vspace{0.6em}

\begin{subfigure}[t]{0.375\linewidth}
\centering
\includegraphics[width=\linewidth]{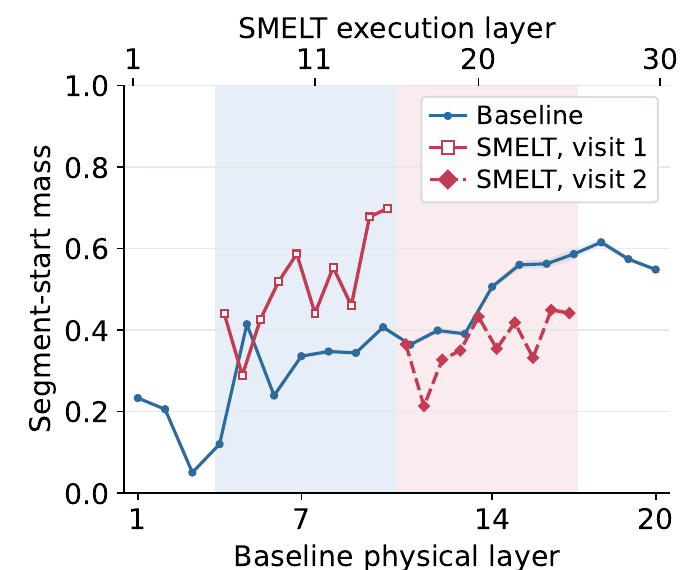}
\caption{600M.}
\end{subfigure}\hspace{0.04\linewidth}
\begin{subfigure}[t]{0.375\linewidth}
\centering
\includegraphics[width=\linewidth]{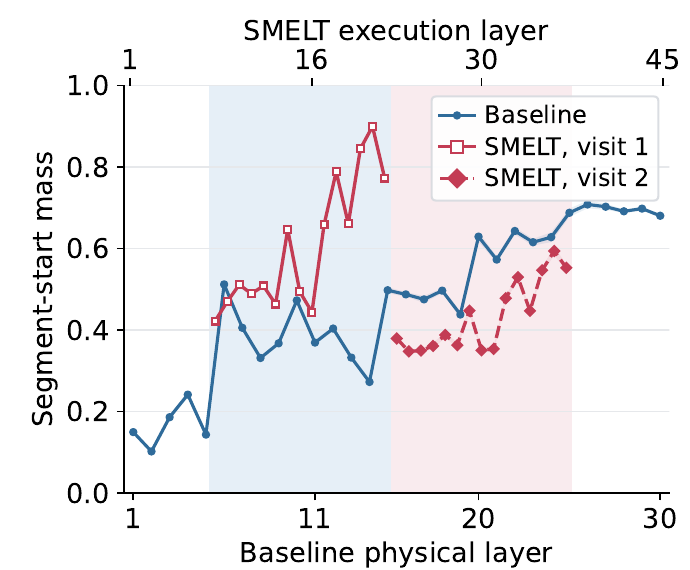}
\caption{1.6B.}
\end{subfigure}
\caption{Same-physical-depth attention-sink profiles on the 1M-token held-out
  sample ($S \approx 85\%$).  Blue circles show the Baseline; open red squares
  and filled red diamonds show \archname{} visits~1 and~2.  Bottom axes index
  Baseline physical layers and top axes index \archname{} execution layers.
  Across all four scales,
  the second visit reduces segment-start mass even though the unlooped
  Baseline tends to accumulate more sink at later depth.}
\label{fig:app-attention-sink-scales}
\end{figure}

\FloatBarrier

\section{Downstream Evaluation Protocol}
\label{app:dclm-categories}

This appendix defines the downstream metrics, multi-seed protocol, and task
categorization referenced by the main text.

\paragraph{Task suite.}
We evaluate on the 22-task DCLM Core suite~\cite{li2024datacomp} and, separately, on MMLU 5-shot~\cite{hendrycks2021mmlu}.
MMLU is not part of the official DCLM Core set but is widely used as a standalone benchmark; we report it individually rather than folding it into our aggregate metrics.
We use the v2 centered-accuracy baselines released by the DCLM project.
Table~\ref{tab:dclm-categories} lists the full task--category mapping.

\paragraph{DCLM Core (centered accuracy).}
For each task $t$ with raw accuracy $a_t$ and DCLM-v2 reference baseline
$\rho_t$ (the ``Random baseline'' field in DCLM's evaluation metadata),
the centered score is
\begin{equation}
  c_t \;=\; \frac{a_t - \rho_t}{1 - \rho_t}\,,
\end{equation}
so that 0 corresponds to the DCLM-v2 reference baseline and 1 to perfect
accuracy.
DCLM Core is the unweighted mean of $c_t$ over the 22 tasks.

\paragraph{DCLM Completion (gold-completion micro loss).}
Nine of the 22 tasks ask for a free-form answer rather than a choice among
options, so every item in them has a gold answer string: CoQA, SQuAD, Jeopardy,
LAMBADA, BB QA Wikidata, BB CS Algorithms, BB Dyck Languages, BB Operators, and
BB Repeat Copy.  For each item we compute the per-token cross-entropy loss on
the ground-truth answer tokens (the ``gold completion'').  DCLM Completion is
the token-weighted (micro) average of these losses over all items of those nine
tasks.  Because it is a loss rather than an accuracy, lower is better.  This
metric complements Core by providing a continuous, fine-grained signal that is
less affected by the discretization inherent in accuracy.

\paragraph{Multi-seed protocol.}
Each few-shot task is evaluated with 10 independent random seeds that
determine the selection and ordering of the few-shot demonstrations.  Reported
scores are the mean over these 10 runs.  Zero-shot tasks (e.g.\ LAMBADA,
Winograd) are deterministic and therefore evaluated with a single run.  All
evaluations preserve the beginning-of-sequence (BOS) token when truncating
context to the model's maximum sequence length.

\begin{table}[H]
\centering
\small
\caption{Evaluation tasks by domain category, each cited to its original
  source.  The 22 DCLM Core tasks are grouped into five categories following
  the official DCLM; MMLU 5-shot is listed
  separately from the Core set.  Parentheses give the number of tasks in each category, and the right column the number of in-context demonstrations used.}
\label{tab:dclm-categories}
\begin{tabular}{lll}
\toprule
Category & Tasks & Shots \\
\midrule
Reading Comprehension (3)
  & BoolQ~\cite{clark2019boolq} & 10 \\
  & CoQA~\cite{reddy2019coqa} & 0 \\
  & SQuAD~\cite{rajpurkar2016squad} & 10 \\
\midrule
World Knowledge (4)
  & ARC-Challenge~\cite{clark2018arc} & 10 \\
  & ARC-Easy~\cite{clark2018arc} & 10 \\
  & BB QA Wikidata~\cite{srivastava2023bigbench} & 10 \\
  & Jeopardy~\cite{kaggle2019jeopardy} & 10 \\
\midrule
Commonsense Reasoning (4)
  & CommonsenseQA~\cite{talmor2019commonsenseqa} & 10 \\
  & COPA~\cite{roemmele2011copa} & 0 \\
  & OpenBookQA~\cite{mihaylov2018openbookqa} & 0 \\
  & PIQA~\cite{bisk2020piqa} & 10 \\
\midrule
Language Understanding (6)
  & BB Language ID~\cite{srivastava2023bigbench} & 10 \\
  & HellaSwag 10-shot~\cite{zellers2019hellaswag} & 10 \\
  & HellaSwag 0-shot~\cite{zellers2019hellaswag} & 0 \\
  & LAMBADA~\cite{paperno2016lambada} & 0 \\
  & Winograd~\cite{levesque2012winograd} & 0 \\
  & Winogrande~\cite{sakaguchi2020winogrande} & 0 \\
\midrule
Symbolic Problem Solving (5)
  & LSAT AR~\cite{zhong2024agieval} & 3 \\
  & BB CS Algorithms~\cite{srivastava2023bigbench} & 10 \\
  & BB Dyck Languages~\cite{srivastava2023bigbench} & 10 \\
  & BB Operators~\cite{srivastava2023bigbench} & 10 \\
  & BB Repeat Copy~\cite{srivastava2023bigbench} & 10 \\
\midrule
\textit{Reported separately} (1)
  & MMLU 5-shot~\cite{hendrycks2021mmlu} & 5 \\
\bottomrule
\end{tabular}
\end{table}
  \let\appendix\seedSavedAppendix
}
\makeatother

\title{\archname: Scaling Laws for Compute-Matched MoE Looped Transformers}
\hypersetup{
  pdftitle={SMELT: Scaling Laws for Compute-Matched MoE Looped Transformers},
  pdfauthor={Shaowen Wang, Ge Zhang, Kairong Luo, Yuhao Wu, Shaofan Liu, Jiaheng Liu, Wenhao Huang, Shen Yan, Jian Li}
}

\affiliation[1]{Tsinghua University}
\affiliation[2]{ByteDance Seed}
\affiliation[3]{M-A-P}
\affiliation[4]{TokenWave.AI}
\contribution{Full author list in Contributions}

\abstract{%
Looped Transformers increase effective depth by iterating a shared block of layers, but most evaluations compare at fixed model size, conflating architectural advantage with extra FLOPs.
We study looping on Mixture-of-Experts Transformers while closely matching per-token FLOPs, total non-embedding parameters, and KV cache.
Through a series of ablations, we arrive at a recipe we call \archname{} (\textbf{S}parse \textbf{M}o\textbf{E} Transformer, middle layers \textbf{L}oop \textbf{T}wice), which loops the middle half of layers twice while matching the unlooped Baseline on all three budgets.
We scale \archname{} across four sizes up to 54B non-embedding parameters and fit a separate Chinchilla-style scaling law for each architecture.
\archname{}'s loss drops faster with compute, saving 6.8--18.0\% of training FLOPs on the compute-optimal frontier.
The advantage transfers to downstream benchmarks beyond what validation loss predicts, is largest on Code, and grows with sample length and the number of in-context examples.
Mechanistic analysis shows that the second visit reduces the attention sink and redirects mass toward content-relevant tokens, an inductive bias that may underlie the observed performance gains.
These results show that looping can improve Transformers even under budget matching, offering a practical recipe that turns depth reuse into measurable gains.
}

\date{August 10, 2026}

\begin{document}

\maketitle

\section{Introduction}
\label{sec:introduction}

Looped Transformers increase a model's effective depth by repeating a shared block of layers rather than stacking new ones~\cite{dehghani2019universal}.
The idea has drawn intense recent interest: looped models match or exceed unlooped models several times their size on arithmetic, multi-hop induction, and math~\cite{geiping2025scaling,saunshi2025reasoning,zhu2025scaling}, and learn algorithmic procedures in context at a fraction of the parameter count~\cite{yang2024looped}.
Since effective depth sets how many sequential computation steps each token receives~\cite{li2024serial,merrill2023parallelism}, looping promises reasoning capability without additional parameters.

\begin{figure*}[t]
\centering
\begin{subfigure}[t]{0.38\linewidth}
\centering
\includegraphics[width=\linewidth]{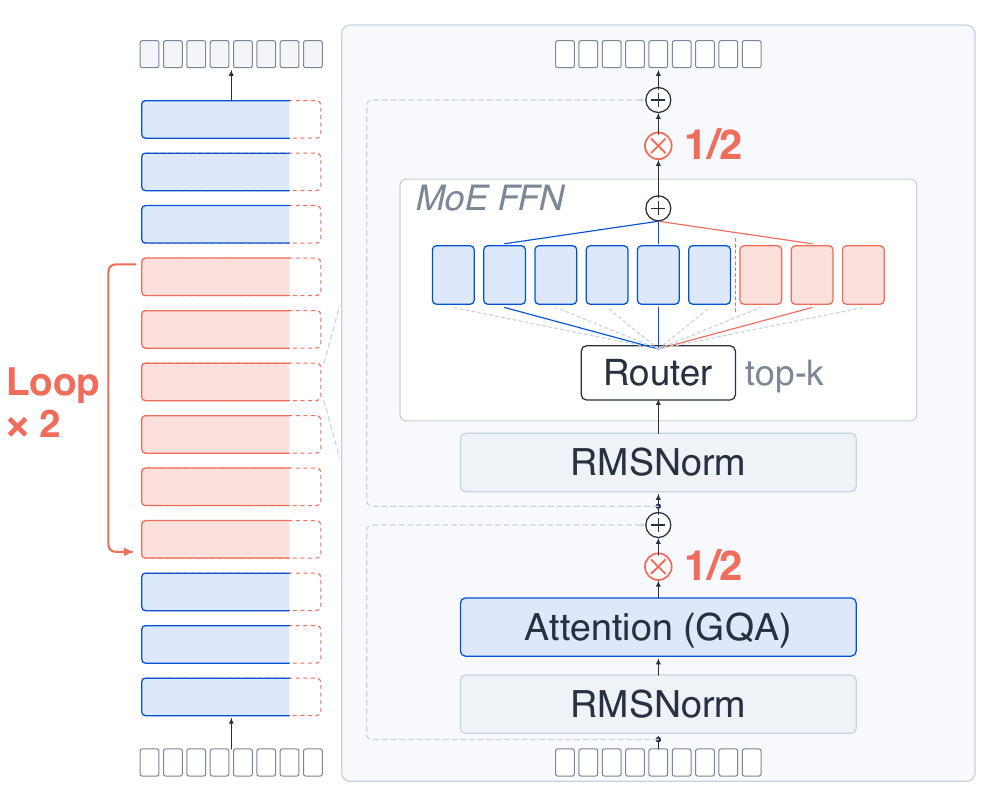}
\caption{The \archname{} recipe.}
\label{fig:hero-arch}
\end{subfigure}\hfill
\begin{subfigure}[t]{0.29\linewidth}
\centering
\includegraphics[width=\linewidth]{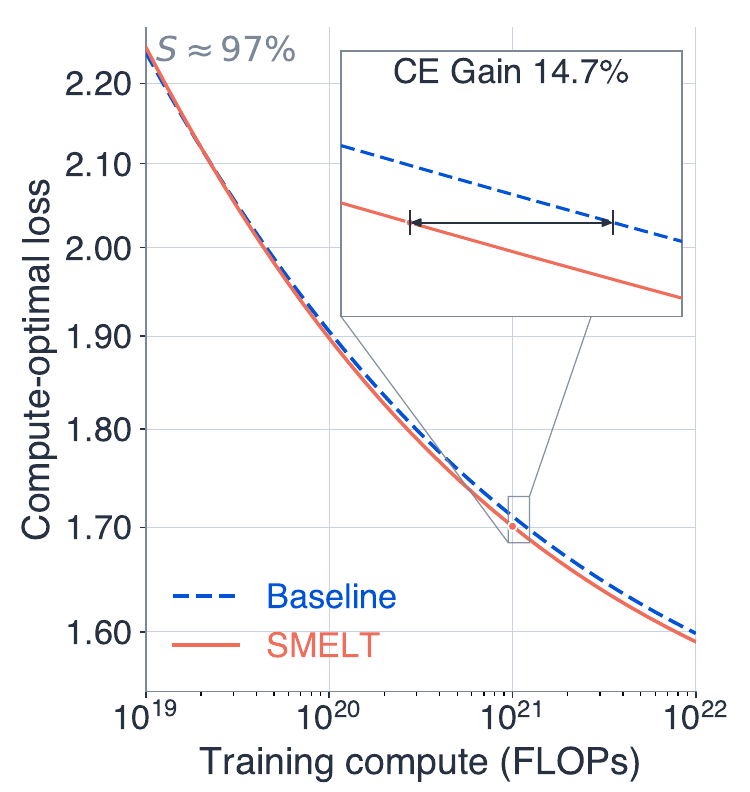}
\caption{Compute-optimal frontier.}
\label{fig:hero-frontier}
\end{subfigure}\hfill
\begin{subfigure}[t]{0.29\linewidth}
\centering
\includegraphics[width=\linewidth]{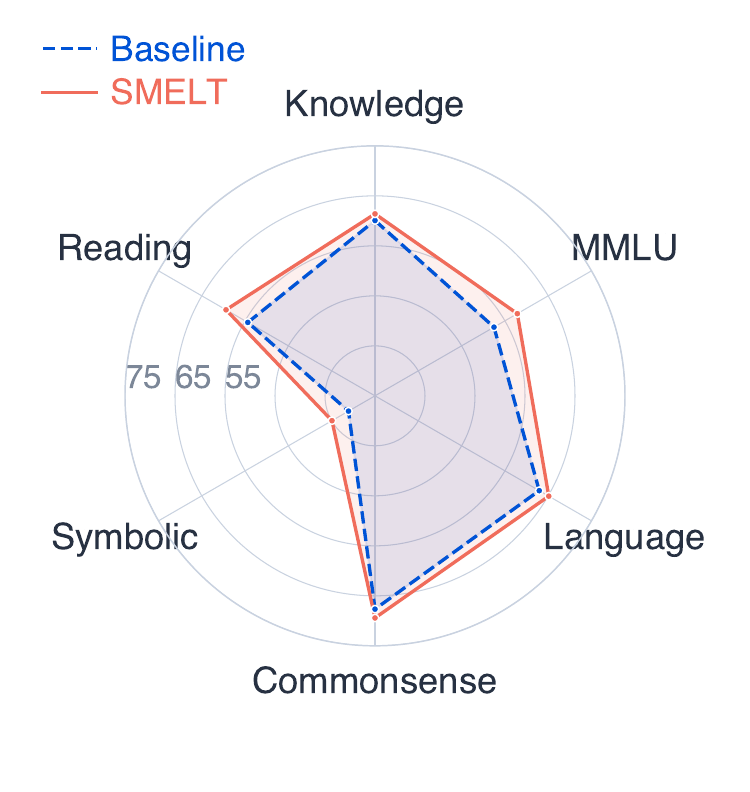}
\caption{Downstream accuracy.}
\label{fig:hero-radar}
\end{subfigure}
\caption{\textbf{\archname{} overview and main results.}
  (a)~The \archname{} recipe on a compute-matched MoE Baseline: loop the middle 50\% of layers twice, narrow the hidden dimension, raise the expert count to recover total parameters, and scale the looped residuals by $1/2$, alongside a smaller attention head size at a higher GQA ratio that holds KV cache nearly unchanged across the extra layer executions.
  Together they closely match per-token FLOPs, total non-embedding parameters, and KV cache (Section~\ref{sec:design}).
  (b)~Compute-optimal loss under separate Chinchilla-style scaling fits at compute-equivalent sparsity $S\!\approx\!97\%$ (defined in Section~\ref{sec:protocol}).
  \archname{}'s frontier drops faster; at $C\!=\!10^{21}$~FLOPs it reaches the same loss with 14.7\% less compute (Section~\ref{sec:scaling-law}).
  (c)~Downstream accuracy at the 1.6B scale (54B non-embedding parameters, $S\!\approx\!97\%$), over the five DCLM Core categories (Appendix~\ref{app:dclm-categories}), abbreviated on the axes as Reading (Comprehension), Knowledge (World Knowledge), Commonsense (Reasoning), Language (Understanding) and Symbolic (Problem Solving), with MMLU on its own axis.
  Each axis is the unweighted mean accuracy over its tasks.
  \archname{} leads the Baseline on all six axes.
}
\label{fig:hero}
\end{figure*}

But a fraction of the parameter count is not a fraction of the cost: looping a 12-layer model to 24 executed layers stores half the weights, yet spends roughly a 24-layer model's per-token FLOPs and needs its full KV cache.
Prior evaluations often keep the stored parameter count fixed while increasing recurrent depth~\cite{geiping2025scaling,prairie2026parcae}, or emphasize parameter efficiency relative to larger untied models~\cite{yang2024looped,zhu2025scaling}; in either case the reported gains conflate architectural advantage with uncontrolled extra computation.
Whether looping has an architectural advantage beyond this extra computation is unknown.
Schwethelm et al.~\cite{schwethelm2026recurrence} control for this: they hold per-token FLOPs fixed for a dense model and find that $r$ recurrences contribute like $r^{0.46}$ unique-block equivalents, but fixing FLOPs also shrinks the looped model's unique parameters, so the deficit may reflect parameter loss rather than a flaw in looping itself.
A clean answer requires holding three budgets all fixed at once: \textit{1)~per-token FLOPs}, which set training and inference cost; \textit{2)~total parameters}, which bound knowledge capacity; and \textit{3)~KV cache}, which limits servable context length.

Mixture-of-Experts Transformers make this budget matching feasible: a looped model can pay for its extra visit by narrowing the hidden dimension, and because MoE decouples total parameters from per-token FLOPs~\cite{fedus2022switch,shazeer2017outrageously}, it can recover the lost capacity by raising expert count rather than giving it up. KV cache parity is restored separately by adjusting the head size and GQA ratio.
Under this matching, we search the loop design space through three ablations and find three rules: \textit{1)~loop the middle half} of layers rather than the full stack, \textit{2)~give the looped model a larger effective depth-to-width ratio} than the Baseline, and \textit{3)~loop twice} rather than three or four times.
Together these lock in a recipe we call \textbf{\archname{}}, a \textbf{S}parse \textbf{M}o\textbf{E} Transformer whose middle layers \textbf{L}oop \textbf{T}wice (Figure~\ref{fig:hero}; Section~\ref{sec:design}).

We scale the recipe up to 54B non-embedding parameters across four model sizes and four sparsity levels, comparing \archname{} against the Baseline at each grid point.
\archname{} reaches lower loss than the Baseline at every scale and sparsity level.
To predict how the advantage evolves at scale, we fit separate Chinchilla-style scaling laws~\cite{hoffmann2022training} for the two architectures.
\archname{}'s loss drops faster with both compute and data, and the CE Gain increases across the fitted compute range (Section~\ref{sec:scaling-law}).

Beyond aggregate validation loss, we examine downstream benchmarks and per-domain performance.
On DCLM~\cite{li2024datacomp} and MMLU~\cite{hendrycks2021mmlu}, \archname{} outperforms the Baseline in nearly all matched pairs, with an improvement that exceeds what the validation loss gap alone would predict.
We also find that among training domains, Code benefits the most.
The advantage is further amplified on longer samples and when more in-context demonstrations are available, suggesting that looping becomes more valuable as the input grows richer in retrievable structure (Section~\ref{sec:data-analysis}).

To build intuition for the advantages observed above, we examine what the looped layers compute differently on their second visit.
We find that the experts selected and the tokens attended to overlap substantially across the two visits, while the residual updates grow substantially larger and stay aligned, indicating that the second visit amplifies the first rather than overwriting it.
Within the attention mechanism, values change more than queries and keys, and the attention sink is broadly reduced, redirecting mass toward content-relevant tokens, which may relate to the observed ICL advantage (Section~\ref{sec:model-analysis}).

\begin{table}[t]
  \centering
  \footnotesize
  \setlength{\tabcolsep}{4pt}
  \renewcommand{\arraystretch}{1.2}
  \caption{Comparison of looped Transformer studies in terms of architecture, loop span, matched budgets, scaling analysis, and key finding.
    \checkmark~= matched; \xmark~= unmatched or grows with loop count; (\checkmark)~= partially matched (see notes).
    Span denotes the fraction of the layer stack that is repeated, and $r$ denotes the number of times that span executes per token (both are defined in Section~\ref{sec:protocol}).}
  \label{tab:related-comparison}
  \fittotextwidth{%
  \begin{tabular}{@{}l c c c c c
    >{\raggedright\arraybackslash}m{3.0cm}
    >{\raggedright\arraybackslash}m{3.6cm}@{}}
  \toprule
   & & & \multicolumn{3}{c}{Matched budgets} & & \\
  \cmidrule(lr){4-6}
  Work & Arch. & Span & FLOPs/token & Stored params & KV cache
    & Scaling analysis
    & Key finding \\
  \midrule
  Huginn~\cite{geiping2025scaling}
    & Dense & Mid.\ 50\%
    & \xmark & \checkmark & \xmark
    & Fixed size (3.5B)
    & Prelude-recur-coda looping improves reasoning; extra FLOPs uncontrolled \\
  Ouro~\cite{zhu2025scaling}
    & Dense & Full (100\%)
    & \xmark & \checkmark & \xmark
    & Fixed size (1.4--2.6B)
    & Full-stack looping improves reasoning; extra FLOPs uncontrolled \\
  Saunshi et al.~\cite{saunshi2025reasoning}
    & Dense & Mid.\ 33\%
    & (\checkmark)$^{a}$ & \xmark & \xmark
    & Fixed size (170M--1B)
    & Middle-block looping lowers perplexity; inductive bias toward reasoning \\
  Schwethelm et al.~\cite{schwethelm2026recurrence}
    & Dense & Mid.$^{c}$
    & \checkmark & \xmark & \checkmark
    & Iso-depth law (25M--1.6B)
    & $r$ recurrences $\approx r^{0.46}$ unique equivalents; looping loses \\
  Prairie et al.~\cite{prairie2026parcae}
    & Dense & Mid.\ 33\%
    & \xmark & \checkmark & \xmark
    & Scaling law at fixed params (100M--1.3B)
    & Optimal $r$ grows with compute; FLOPs and KV grow with $r$ \\
  Lee et al.~\cite{lee2026sparse}
    & MoE & Full (100\%)
    & \checkmark & \xmark & \checkmark
    & IsoFLOP law (16M--305M active)
    & Sparse routing mitigates the loop penalty; Looped-MoE beats dense Base but trails MoE in loss \\
  LoopMoE~\cite{chen2026loopmoe}
    & MoE & Mid.\ 33\%
    & \checkmark & \checkmark & \xmark
    & Fixed size (3B, 9B)
    & Wins 8/9 benchmarks at 3B and 9/9 at 9B \\
  Gao et al.~\cite{gao2026loopie}
    & MoE & Each layer
    & (\checkmark)$^{b}$ & \xmark & \xmark
    & Fixed size (6B, 20B)
    & Wins; $r\!=\!2$ optimal; memory savings reinvested into width \\
  \midrule
  \textbf{\archname{} (ours)}
    & MoE & Mid.\ 50\%
    & \checkmark & \checkmark & \checkmark
    & Scaling ladder across 0.1B--54B non-embedding parameters
    & Wins; 6.8--18.0\% FLOPs saving on validation loss; benchmark gains exceed validation loss prediction \\
  \bottomrule
  \end{tabular}%
  }%
  
  \vspace{2pt}
  \raggedright\footnotesize
  $^{a}$~Matched-FLOPs perplexity comparison for the prefix-loop-suffix layout; the accuracy-vs-loop-count results are at fixed model size.\quad
  $^{b}$~Matches wall-clock training time rather than per-token FLOPs.\quad
  $^{c}$~Physical span varies with loop count (approx.\ 67\%, 50\%, 33\% for $r{=}2,3,4$).
  \end{table}

Our results show that looping can improve Transformers even under budget matching, with particularly strong gains on structured data and in-context learning tasks.
Together with the growing body of work exploring looped language models from complementary angles~\cite{geiping2025scaling,zhu2025scaling,saunshi2025reasoning,schwethelm2026recurrence,prairie2026parcae}, they suggest that looping is a robust and increasingly well-understood design axis.
\section{Related Work}
\label{sec:related-work}

\paragraph{Looped Transformers.}
The Universal Transformer~\cite{dehghani2019universal} introduced weight sharing across depth, applying a shared block iteratively to decouple effective depth from unique parameter count.
Huginn~\cite{geiping2025scaling} scales this idea to 3.5B parameters with a prelude--recur--coda layout (2--4--2 layers at its largest scale) that loops the middle 50\%, establishing the template that most subsequent work adopts: an unshared entry block, a shared middle block that is iterated, and an unshared exit block.
Ouro~\cite{zhu2025scaling} takes a different partition: it loops the entire $L$-layer stack at 1.4--2.6B.
Both show that looped models match or exceed much larger unlooped models on reasoning benchmarks.
A recurring finding across this body of work is that the middle layers benefit most from repetition.
Saunshi et al.~\cite{saunshi2025reasoning} compare looped models against both iso-parameter and iso-FLOP baselines: looped models have worse perplexity than the iso-FLOP baseline but disproportionately improve reasoning tasks, and iterating only a middle span, the prelude-recur-coda layout above, further improves perplexity over full-stack looping.
Kapl et al.~\cite{kapl2026growing} report that, among the block placements tested in their setting, adapting a localized middle block for looping yields the strongest overall results.
A related but untied construction appears in SOLAR~\cite{kim2024solar}, whose depth-up-scaling method also targets the middle layers, duplicating them to grow a pretrained model before continued pretraining.
ALBERT~\cite{lan2020albert} shares weights across all layers for parameter efficiency, an early instance of full cross-layer tying that predates the partial-loop designs above.
Beyond the choice of which layers to loop, other work explores how to loop: Relaxed Recursive Transformers~\cite{bae2025relaxed} add per-layer LoRA adapters to relax strict weight tying, Mixture-of-Recursions~\cite{bae2025mixture} routes each token to a learned recursion depth, and Think-at-Hard~\cite{fu2025think} learns when to iterate on hard tokens.
Among these studies, Saunshi et al.\ additionally report an iso-FLOP comparison in which the looped model uses fewer unique parameters than the untied baseline~\cite{saunshi2025reasoning}.
RINS~\cite{alabdulmohsin2025recursive} matches total training compute by training its early-block recursive model on fewer tokens than its baseline, whereas we additionally match the token budget.
Most other works above compare looped models against parameter-matched baselines, so the reported gains conflate architectural advantage with the extra FLOPs that repetition spends; whether looping helps or hurts under jointly matched compute and token budgets remains open.

\paragraph{Mixture-of-Experts and loop recurrence.}
Answering the question above requires holding three budgets fixed at once: per-token FLOPs, total parameters, and KV cache (Table~\ref{tab:related-comparison}).
MoE makes this feasible by decoupling total parameters from per-token FLOPs~\cite{shazeer2017outrageously,fedus2022switch}: a looped model can pay for its second visit by narrowing the hidden dimension, then recover the lost capacity by adding experts.
MoEUT~\cite{csordas2024moeut} proposes an MoE-based Universal Transformer that, for the first time, matches standard Transformers on language modeling.
Lee et al.~\cite{lee2026sparse} find that sparse layers are critical to scaling looped models, because routing divergence across visits recovers expressivity lost to weight tying.
Jaggi~\cite{jaggi2026tying} similarly reuses experts across middle layers and reinvests the saved parameters in larger expert pools.
Several concurrent works train MoE Looped Transformers at scale: LoopMoE~\cite{chen2026loopmoe} matches total parameters and per-token FLOPs at 3B and 9B; Gao et al.~\cite{gao2026loopie} train 6B and 20B models under wall-clock matching and independently find two loop steps optimal, corroborating our loop-count ablation (Section~\ref{sec:loop-count}).
Knupp et al.~\cite{knupp2026dreamer} (DREAMER) recurrently apply a single MoE layer 16--32 times with attention over previous depths under matched FLOPs, parameters, and memory; we view this single-layer recurrence as closer to implicit-depth models than to the block-level looping studied here.
Chen et al.~\cite{chen2026moue} share a universal expert pool across layers, converting depth into virtual MoE width rather than looping the full layer block.
These concurrent studies show that looping improves MoE models at individual scales.
We ask a complementary question: does the advantage hold across a full scaling ladder when per-token FLOPs, total parameters, and KV cache are all matched?
For block-level looping, our comparison is the first multi-scale study that closes all three budgets simultaneously and fits a separate scaling surface for each architecture, converting point observations into an attributable and extrapolable claim (Section~\ref{sec:scaling-law}).

\paragraph{Scaling laws.}
Kaplan et al.~\cite{kaplan2020scaling} established power-law relations between loss, model size, and training tokens.
Hoffmann et al.~\cite{hoffmann2022training} (Chinchilla) refined the compute-optimal allocation and showed that parameters and tokens should scale at roughly equal rates with compute.
For MoE models, Clark et al.~\cite{clark2022unified} and Abnar et al.~\cite{abnar2025parameters} extend these laws with routing variables such as expert count and sparsity; Li et al.~\cite{li2025moe} study the conditions under which MoE surpasses dense models at equal resources.
Recent work extends these analyses to looped architectures.
Schwethelm et al.~\cite{schwethelm2026recurrence} fit a joint scaling law under an iso-depth protocol that holds effective depth (and therefore FLOPs and KV cache) fixed while reducing unique parameters.
In their law, the effective parameter count of the recurrent block is $N_{\mathrm{once}} + r^{0.46}\,N_{\mathrm{rec}}$: $r$ recurrent executions contribute like $r^{0.46}$ unique-block equivalents.
The sublinear exponent means recurrence yields diminishing capacity per iteration, so the looped model loses at matched compute.
Prairie et al.~\cite{prairie2026parcae} fit compute-optimal recurrence at fixed unique parameter count, but depth, inference FLOPs, and KV cache all grow with $r$, so the fit cannot separate the per-parameter sharing cost from the growing budgets.
These two results bracket the question from opposite sides: Schwethelm et al.\ price looping's cost at fixed depth, Prairie et al.\ find looping's benefit at fixed parameters.
We fit a separate Chinchilla-style surface for each architecture on MoE models while closely matching per-token FLOPs, total non-embedding parameters, and KV cache, which lets us isolate the architectural effect and measure the compute saving on the frontier (Section~\ref{sec:scaling-law}).

\paragraph{Attention sinks.}
Autoregressive Transformers concentrate attention mass on initial tokens regardless of semantic content.
Xiao et al.~\cite{xiao2024streamingllm} identify this attention-sink phenomenon and exploit it for streaming inference; Gu et al.~\cite{gu2025sink} characterize when sinks emerge during pretraining; Sun et al.~\cite{sun2024massive} link them to massive activations in the residual stream, with an analogous register effect in vision models~\cite{darcet2024registers}; Barbero et al.~\cite{barbero2025firsttoken} argue that sinks counteract over-mixing and therefore strengthen as depth grows, a trend also observed geometrically by Ruscio et al.~\cite{ruscio2025sinking}; and Qiu et al.~\cite{qiu2025gated} show that a query-dependent sigmoid gate after attention strongly mitigates the sink in the evaluated models.
These works treat the sink as a static property of a trained model.
Our mechanistic analysis (Section~\ref{sec:model-analysis}) adds a dynamic observation: within a single forward pass, a second weight-tied visit reduces sink mass and redirects it toward content-relevant tokens.
\section{Design Recipe under Matched Compute}
\label{sec:design}

\subsection{Training protocol}
\label{sec:protocol}

We write \textbf{Baseline} for the standard unlooped MoE Transformer and \textbf{Looped Transformer} for the variant that repeats a contiguous span of $m$ layers $r$ times.
The ablations in Sections~\ref{sec:loop-span}--\ref{sec:loop-count} sweep both $m$ and $r$; the values they select define \archname{} in Section~\ref{sec:loop-count}.
All runs share the same architecture template, optimizer, data, and evaluation.
Only the loop configuration (span, count, and effective depth) and the matched width/expert/head hyperparameters (Section~\ref{sec:design-setup}) differ between the two.
In the Looped Transformer, each sublayer's residual update within the looped span is scaled by $1/r$ where $r$ is the loop count~\cite{wang2026residual} (Figure~\ref{fig:hero-arch}); this prevents correlated weight-tied updates from inflating the residual stream across visits.

\paragraph{Architecture.}
Each model is a decoder-only Transformer with sparse MoE feed-forward layers.
Every MoE layer routes each token to its top-8 experts.
Attention uses grouped-query attention (GQA)~\cite{ainslie2023gqa}.
We adopt an internal Baseline family and use its active non-embedding parameter count to label four matched scales: 100M, 200M, 600M, and 1.6B (physical depth $L = 10$, $12$, $20$, $30$).
Appendix~\ref{app:configs} lists the configuration of every Baseline in the grid.

\paragraph{Compute-equivalent sparsity.}
An MoE model stores $N$ parameters but activates only a fraction per token.
We measure this fraction against a fully active control: an unlooped Baseline that activates all experts per token at the same scale, so its per-token FLOPs reflect 100\% utilization of total parameters.
Let $F_0$ and $N_0$ be this control's per-token training FLOPs and total non-embedding parameters.
For any configuration with per-token FLOPs $F$ and $N$ total non-embedding parameters, we define
\begin{equation}
N_{\mathrm{act}}^{\mathrm{eq}} = \frac{F}{F_0}\,N_0,
\qquad
S = 1 - \frac{N_{\mathrm{act}}^{\mathrm{eq}}}{N}.
\label{eq:equivalent-sparsity}
\end{equation}
$N_{\mathrm{act}}^{\mathrm{eq}}$ scales the control's parameter count by the FLOPs ratio: a model that spends half the control's FLOPs is treated as activating half the parameters.
$S$ is the inactive share, analogous to the parameter-ratio sparsity of Abnar et al.~\cite{abnar2025parameters} but computed from FLOPs ratios to account for context-dependent attention cost; the control maps to $S = 0$ by construction.
A Looped Transformer matched to this control at $S = 0$ is not dense.
Matching (Section~\ref{sec:design-setup}) narrows $H$ and raises expert count to recover total parameters, so the Looped Transformer routes to a top-8 subset of a larger pool even at $S = 0$ (e.g., 16 experts per layer at 200M).
$S = 0$ means the per-parameter compute intensity matches the fully active Baseline, not that every expert is activated.
We report matched Baseline / Looped Transformer pairs at $S \approx 85\%$, $95\%$, and $97\%$; the same label applies to both architectures in all figures.

\paragraph{Optimizer and schedule.}
We use AdamW~\cite{loshchilov2019decoupled} with a warmup-stable-decay (WSD) schedule~\cite{hu2024minicpm}.
The stable phase trains at a constant learning rate over 196{,}075 steps with a global batch size of 256 sequences (${\approx}\text{1M}$ tokens per step).
From each stable run we branch six cosine-decay schedules at steps 10{,}000, 20{,}000, 50{,}000, 100{,}000, 150{,}000, and 196{,}075, each decaying over 10B additional tokens.
The six branch points give six token horizons per configuration.

\paragraph{Data.}
We pretrain on an internal corpus.
The stable phase of the WSD schedule consumes 205B tokens; each cosine-decay branch then trains on 10B additional tokens drawn from fresh data not seen during the stable phase, bringing the longest branch to approximately 215B tokens with no repeated data.
All matched Baseline / \archname{} pairs at the same scale and sparsity level are trained on identical token sequences.
The held-out validation set covers 39 individual sources grouped into five categories: Code, Math/STEM, Knowledge, Finance, and Web.
Training sequences are packed into 4096-token contexts with segment-level attention masks, where each segment is one document within the packed context.

\paragraph{Evaluation.}
We score three metrics throughout.
Validation loss is the token-weighted cross-entropy on the held-out validation set, averaged across all 39 sources.
The other two come from the 22-task DCLM Core suite~\cite{li2024datacomp}, evaluated with 10 random few-shot seeds per few-shot task and BOS-preserving truncation (details in Appendix~\ref{app:dclm-categories}).
The seven zero-shot tasks in the suite are deterministic and run once.
DCLM Core is the mean centered accuracy over the 22 tasks (rescaled so that the DCLM-v2 reference baseline maps to zero).
We define \textit{DCLM Completion} as the token-weighted cross-entropy on the gold answer tokens of the nine tasks in the suite that ask for a free-form answer rather than a choice among options (lower is better).
We also run MMLU 5-shot~\cite{hendrycks2021mmlu} and report it on its own, as it is widely used as a standalone benchmark.
To prevent benchmark contamination, we filtered known evaluation items from the pretraining corpus before training.

\subsection{Matching as a compute-allocation problem}
\label{sec:design-setup}

We care about three quantities when we compare the Looped Transformer with the Baseline.
Total parameters control how much knowledge a model can memorize; per-token FLOPs set inference and training cost; KV cache determines the longest context a deployment can serve.
To attribute performance gaps primarily to architecture, we closely match all three between the Looped Transformer and the Baseline.

Consider the cost of looping.
A standard MoE Transformer with $L$ physical layers spends per-token training FLOPs proportional to $L \cdot H^2$, where $H$ is the hidden dimension.
When we loop a contiguous span of $m$ layers $r$ times, the effective depth grows to $L_{\mathrm{eff}} = L + (r{-}1)\,m$, and FLOPs grow in proportion, while total parameters stay the same.
To keep per-token FLOPs fixed, we must shrink the model elsewhere: we narrow $H$ or reduce $L$.
That shrinkage also cuts total parameters and KV cache, so we must compensate on three axes at once.

Table~\ref{tab:matching} lists the three budgets, the adjustment that holds each, and the residual mismatch after matching.
Suppose we want a Looped Transformer at 200M, $S \approx 95\%$ that loops the middle 6 of its 12 layers twice, executing 18 layers in total.
The matched Baseline has $L = 12$, hidden dimension $H = 1280$, and $192$ experts per layer ($1.33 \times 10^{9}$ training FLOPs per token, $3.87 \times 10^{9}$ total non-embedding parameters).
The 6 extra layer executions cost FLOPs, so we narrow $H$ from 1280 to 1056 to keep per-token FLOPs close.
Narrowing $H$ also shrinks every expert's FFN, which reduces total parameters.
To recover them, we raise the per-layer expert count from 192 to 288.
Expert count is uniform across all layers in both architectures.
After both adjustments, per-token FLOPs land at $1.37 \times 10^{9}$ ($+2.9\%$ vs.\ the Baseline's $1.33 \times 10^{9}$), total parameters at $3.89 \times 10^{9}$ ($+0.4\%$), and KV cache within $4\%$.
Both FLOPs counts are measured at the mean document length within the packed 4096-token contexts (since segment-level masks restrict attention to each document) and include the attention cost that scales with that length.

\begin{center}
\begin{minipage}{\linewidth}
\small
\captionof{table}{
Budget-matching adjustments and typical residual mismatch.
Exact per-configuration ratios are reported in Appendix~\ref{app:configs}.
}
\label{tab:matching}
\centering
\begin{tabular}{lll}
\toprule
Budget & Adjustment & Typical mismatch \\
\midrule
Per-token FLOPs & Shrink $H$, or change $L$ & $<4\%$ \\
Total parameters & Add experts & $<1\%$ \\
KV cache & Head size or GQA ratio & $<4\%$ \\
\bottomrule
\end{tabular}
\end{minipage}
\end{center}

We apply this matching at each compute-equivalent sparsity level $S \approx 85\%$, $95\%$, and $97\%$ (Section~\ref{sec:protocol}).
With matching in place, we search the Looped Transformer design space through three ablations at 200M: which layers to loop, how deep the model should be, and how many passes to run.

\subsection{Looping the middle half beats full looping}
\label{sec:loop-span}

We first ask which layers should loop, starting with two passes.
This ablation fixes physical depth at $L = 12$ and varies which contiguous middle segment loops twice.
The segment length ranges from 0 (Baseline) through 12 (full looping).
We run the sweep at $S \approx 85\%$ and $S \approx 95\%$ to test whether the peak moves with sparsity.

\begin{center}
\begin{minipage}{\linewidth}
\small
\captionof{table}{Loop-span sweep at 200M and $L=12$.
  Validation loss is minimized near 50\% span for both sparsity levels; we adopt this as the default loop span.
  Bold marks the best value within each sparsity level.}
\label{tab:loop-span}
\centering
\begin{tabular}{rrrccc}
\toprule
Looped layers & Span (\%) & Effective layers &
Val.\ loss $\downarrow$ & DCLM Core $\uparrow$ & DCLM Completion $\downarrow$ \\
\midrule
\multicolumn{6}{c}{$S \approx 85\%$} \\
0  & 0   & 12 & 1.9445 & $24.92 \pm 0.13$ & 2.2887 \\
2  & 17  & 14 & 1.9384 & $26.20 \pm 0.14$ & 2.2824 \\
4  & 33  & 16 & 1.9275 & $25.13 \pm 0.16$ & 2.2733 \\
6  & 50  & 18 & \textbf{1.9257} & $\mathbf{27.57 \pm 0.13}$ & \textbf{2.2635} \\
8  & 67  & 20 & 1.9374 & $25.84 \pm 0.10$ & 2.2843 \\
10 & 83  & 22 & 1.9413 & $24.45 \pm 0.17$ & 2.2835 \\
12 & 100 & 24 & 1.9322 & $25.31 \pm 0.17$ & 2.2759 \\
\midrule
\multicolumn{6}{c}{$S \approx 95\%$} \\
0  & 0   & 12 & 1.8735 & $29.34 \pm 0.10$ & 2.2189 \\
2  & 17  & 14 & 1.8562 & $30.84 \pm 0.20$ & 2.2022 \\
4  & 33  & 16 & 1.8524 & $31.22 \pm 0.18$ & 2.1970 \\
6  & 50  & 18 & \textbf{1.8517} & $29.68 \pm 0.13$ & 2.1917 \\
8  & 67  & 20 & 1.8544 & $\mathbf{31.78 \pm 0.18}$ & \textbf{2.1891} \\
10 & 83  & 22 & 1.8572 & $31.49 \pm 0.15$ & 2.1964 \\
12 & 100 & 24 & 1.8601 & $30.82 \pm 0.14$ & 2.2038 \\
\bottomrule
\end{tabular}
\end{minipage}
\end{center}

Validation loss reaches its minimum near 50\% span for both sparsity levels (Table~\ref{tab:loop-span}).
We select the span on validation loss because the DCLM metrics do not track it in this sweep.
At $S \approx 95\%$, for instance, DCLM Core peaks at 67\% span and dips at 50\% span---the opposite of the validation-loss ranking---while validation loss, which averages over billions of tokens, varies smoothly.
The reported standard errors ($\pm0.1$--$0.2$ points) reflect variability across the 10 evaluation seeds; they do not capture training-run variability, since each configuration is trained once.
We therefore fix the span at 50\% for all subsequent experiments and treat the DCLM columns as a consistency check rather than a selection criterion.
The optimum agrees with depth-dependent specialization in Transformer representations~\cite{lad2024remarkable,tenney2019bert} and with partial-looping studies~\cite{saunshi2025reasoning,kapl2026growing,cai2026t2mlr}: looping a middle block consistently outperforms full-stack looping, as the first and last layers serve specialized roles that benefit from independent parameters.

\subsection{The Looped Transformer prefers a larger effective depth-to-width ratio}
\label{sec:depth-vs-width}

With the span fixed at 50\% and still under two passes (Section~\ref{sec:loop-span}), we sweep the depth-to-width ratio for both the Baseline and the Looped Transformer at 200M, $S \approx 85\%$ (Table~\ref{tab:depth-width}).
At each depth, we re-match the remaining architectural hyperparameters under the FLOPs, parameter, and KV-cache constraints of Section~\ref{sec:design-setup}.

\begin{center}
\begin{minipage}{\linewidth}
\small
\captionof{table}{Effective-depth sweep at 200M, $S \approx 85\%$, 50\% loop span.
  Physical depth is unique layers; effective depth is layers run in a forward pass.
  The $12/18$ Looped Transformer wins on all three metrics; its physical depth matches the Baseline's validation loss optimum (12).
  Bold marks the best value within each architecture.}
\label{tab:depth-width}
\centering
\begin{tabular}{rrccc}
\toprule
Physical depth & Effective depth &
Val.\ loss $\downarrow$ & DCLM Core $\uparrow$ & DCLM Completion $\downarrow$ \\
\midrule
\multicolumn{5}{c}{Baseline} \\
9  & 9  & 1.9638 & $24.01 \pm 0.14$ & 2.2983 \\
12 & 12 & \textbf{1.9445} & $24.92 \pm 0.13$ & 2.2887 \\
15 & 15 & 1.9457 & $\mathbf{25.84 \pm 0.15}$ & \textbf{2.2824} \\
18 & 18 & 1.9542 & $24.26 \pm 0.14$ & 2.2967 \\
\midrule
\multicolumn{5}{c}{Looped Transformer} \\
8  & 12 & 1.9591 & $24.39 \pm 0.17$ & 2.2932 \\
10 & 15 & 1.9389 & $25.93 \pm 0.14$ & 2.2871 \\
12 & 18 & \textbf{1.9257} & $\mathbf{27.57 \pm 0.13}$ & \textbf{2.2635} \\
14 & 21 & 1.9433 & $26.55 \pm 0.23$ & 2.2766 \\
\bottomrule
\end{tabular}
\end{minipage}
\end{center}

The Looped Transformer's optimal effective depth-to-width ratio is larger than the Baseline's (Table~\ref{tab:depth-width}): the Baseline peaks at physical depth 12, while the Looped Transformer peaks at $12/18$ (physical 12, executed 18).
On validation loss, the winning Looped Transformer has the same physical depth as the Baseline (12), so we set the Looped Transformer's physical depth equal to the Baseline's in all subsequent experiments.

We hypothesize that the Looped Transformer can sustain a larger effective depth-to-width ratio because its additional execution depth does not increase its physical depth: the shared layers receive gradient contributions from multiple visits, including later occurrences with shorter paths to the output.
This multi-depth gradient signal may make additional serial computation easier to optimize than an equally deep stack of independently parameterized layers.

\subsection{Two loops beat three or four}
\label{sec:loop-count}

Sections~\ref{sec:loop-span}--\ref{sec:depth-vs-width} fix the span at 50\% and the physical depth at the Baseline's optimum.
The remaining question is how many times to loop those middle layers.
Table~\ref{tab:loop-count} reports the 200M, $S \approx 85\%$ sweep from one to four visits.
Extra visits cost FLOPs, so the budget-matching procedure of Section~\ref{sec:design-setup} narrows width to stay on the same budget line; more loops therefore mean a thinner model.

\begin{center}
\begin{minipage}{\linewidth}
\small
\captionof{table}{Loop-count sweep at 200M, $S \approx 85\%$, 50\% loop span, physical depth matched to the Baseline.
  Two visits give the lowest validation loss and highest DCLM Core; more loops force a thinner model under matched FLOPs.
  Bold marks the best value.}
\label{tab:loop-count}
\centering
\begin{tabular}{lrccc}
\toprule
Model & Effective depth & Val.\ loss $\downarrow$ & DCLM Core $\uparrow$ & DCLM Completion $\downarrow$ \\
\midrule
Baseline ($1\times$) & 12 & 1.9445 & $24.92 \pm 0.13$ & 2.2887 \\
Looped Transformer $2\times$ & 18 & \textbf{1.9257} & $\mathbf{27.57 \pm 0.13}$ & \textbf{2.2635} \\
Looped Transformer $3\times$ & 24 & 1.9385 & $27.15 \pm 0.20$ & 2.2779 \\
Looped Transformer $4\times$ & 30 & 1.9360 & $27.30 \pm 0.19$ & 2.2820 \\
\bottomrule
\end{tabular}
\end{minipage}
\end{center}

Two visits win on all three metrics (Table~\ref{tab:loop-count}); both a third and a fourth visit regress because the matched FLOPs cap forces a thinner model.
The regression is not ordered in the loop count, so only the shared gap to $2\times$ is meaningful.

We therefore fix loop count at 2 for all subsequent scale-up experiments.
The three ablations together lock the recipe we call \textbf{\archname{}}, a \textbf{S}parse \textbf{M}o\textbf{E} Transformer whose middle layers \textbf{L}oop \textbf{T}wice, through three rules: \textit{1)~loop the middle half} of layers rather than the full stack (Section~\ref{sec:loop-span}), \textit{2)~give the looped model a larger effective depth-to-width ratio} than the Baseline (Section~\ref{sec:depth-vs-width}), and \textit{3)~loop twice} rather than three or four times.
\section{Scaling Laws and Compute Savings}
\label{sec:scaling-law}

The three ablations in Section~\ref{sec:design} locked the \archname{} recipe, but they all ran at the 200M scale (up to 3.9B non-embedding parameters).
Does the advantage survive when models grow to tens of billions of parameters?
This section scales the recipe across four sizes up to 54B non-embedding parameters, fits a separate scaling surface for each architecture, and quantifies how much compute \archname{} saves.

\subsection{The \archname{} recipe scales across the grid}
\label{sec:main-grid}

We apply the budget matching protocol and the locked recipe from Sections~\ref{sec:protocol}--\ref{sec:loop-count} to the full $4 \times 4$ grid: four scales (100M / 200M / 600M / 1.6B) crossed with an $S{=}0\%$ dense-reference control and three sparse levels $S \approx \{85\%,95\%,97\%\}$.
Each cell trains a matched Baseline / \archname{} pair under the shared WSD schedule from Section~\ref{sec:protocol}.
The six cosine-decay branches per run (Section~\ref{sec:protocol}) yield six token horizons, which provide the variation along the data axis needed to fit the data term of the scaling law (Section~\ref{sec:scaling-form}).
Forking decay branches from one stable run is cheaper than training a separate run per horizon, and annealed checkpoints are the ones that reflect converged loss~\cite{hu2024minicpm,tissue2025scaling}.
That gives 32 runs, $16 \times 6 = 96$ matched Baseline / \archname{} pairs, and 192 evaluation endpoints.

\begin{figure}[H]
\centering
\begin{subfigure}[t]{0.48\linewidth}
\centering
\includegraphics[width=\linewidth]{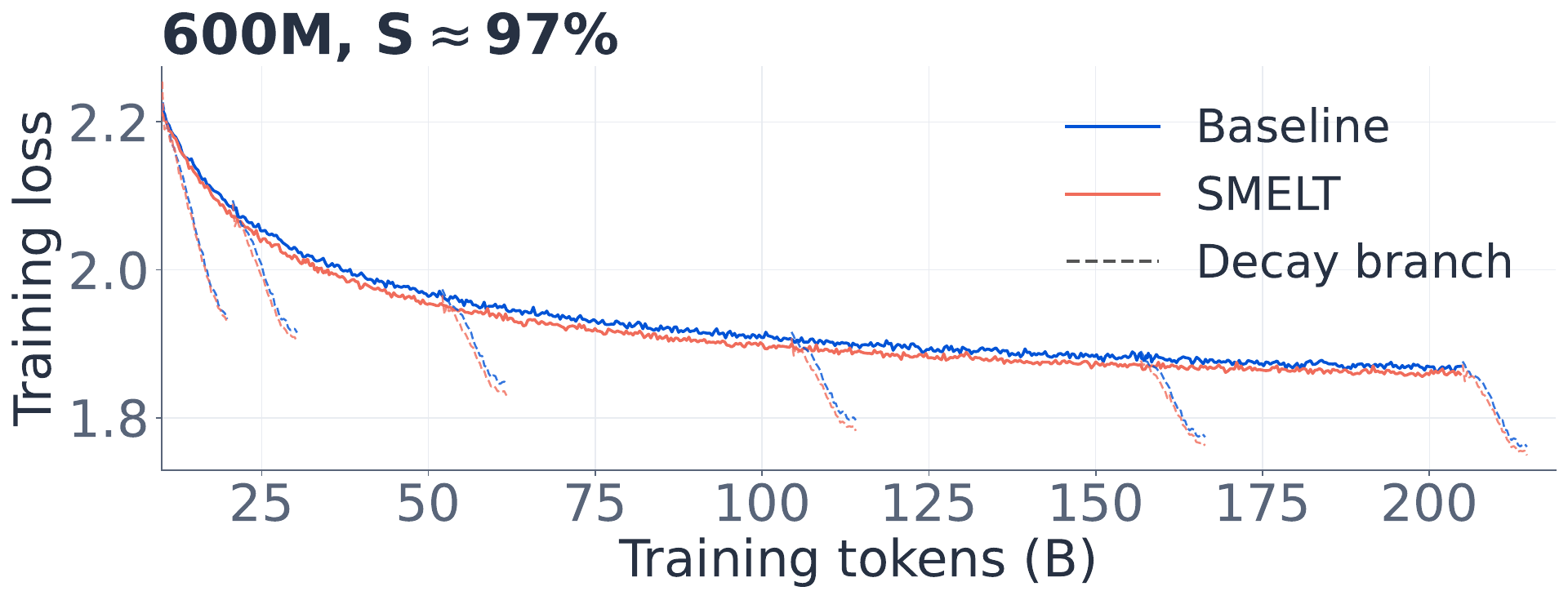}
\caption{600M, $S \approx 97\%$.}
\end{subfigure}\hfill
\begin{subfigure}[t]{0.48\linewidth}
\centering
\includegraphics[width=\linewidth]{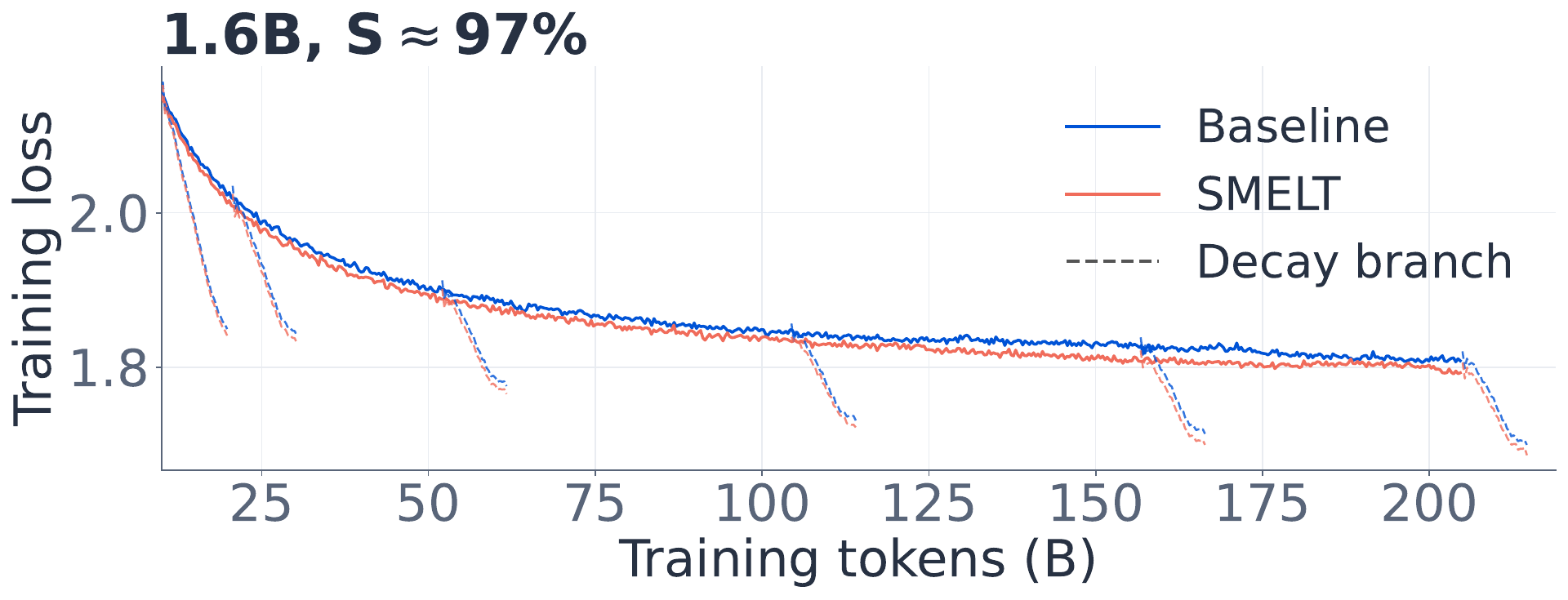}
\caption{1.6B, $S \approx 97\%$.}
\end{subfigure}
\caption{Training loss at $S \approx 97\%$ for two scales.  Solid lines are the
  stable (constant-LR) phase; dashed lines are cosine-decay branches forked
  at six checkpoints.  \archname{} (red) runs below the Baseline
  (blue) throughout.}
\label{fig:training-curve}
\end{figure}

Figure~\ref{fig:training-curve} shows two cells at $S \approx 97\%$: the 600M scale (22B non-embedding parameters) and the 1.6B scale (54B non-embedding parameters).
In both cells, the \archname{} curve tracks below the Baseline through most of training, and the gap persists into every cosine-decay branch.
Appendix~\ref{app:training-curves} gives the curves for all 16 grid cells.

\FloatBarrier

\begin{figure}[H]
\centering
\includegraphics[width=\linewidth]{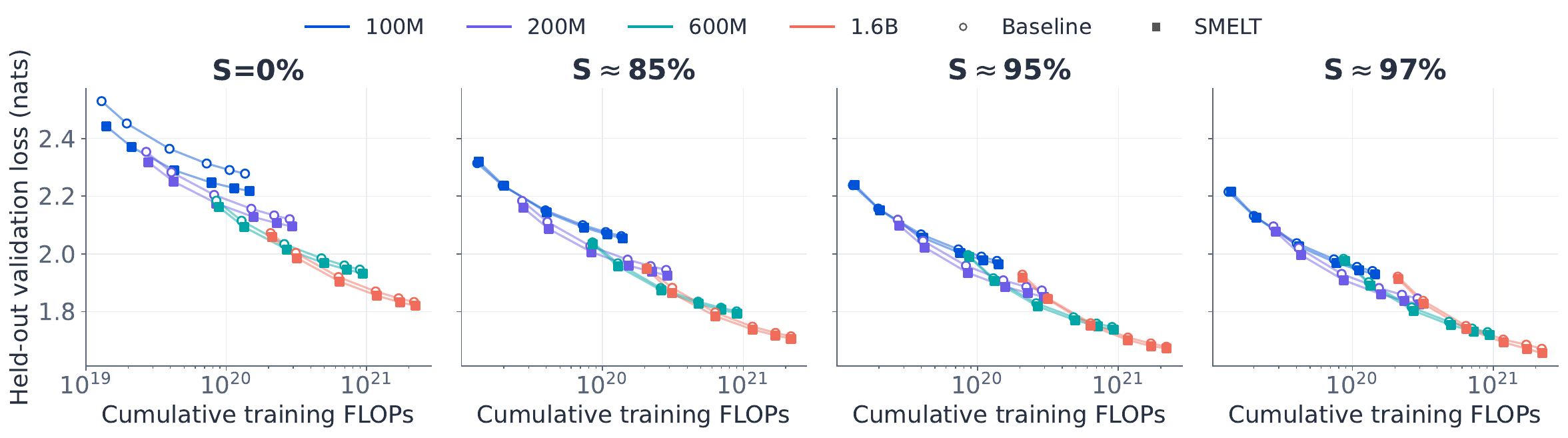}
\caption{Validation loss vs.\ cumulative training FLOPs, faceted by
  compute-equivalent sparsity $S$ and colored by scale.  Open circles = Baseline; filled squares =
  \archname{}.  At comparable measured compute, \archname{} reaches
  lower loss in every cell.}
\label{fig:loss-vs-compute}
\end{figure}

Figure~\ref{fig:loss-vs-compute} replaces the token axis with cumulative training FLOPs, a hardware-independent proxy for arithmetic training cost.
Each panel is one sparsity level; colors mark the four scales.
Open circles mark Baseline endpoints; filled squares mark \archname{}.

\FloatBarrier

\subsection{Scaling form and compute efficiency gain}
\label{sec:scaling-form}

\paragraph{Scaling form.}
Kaplan et al.~\cite{kaplan2020scaling} established power-law relations between loss, model size, and training tokens.
The Chinchilla law~\cite{hoffmann2022training} recasts the joint dependence as an additive form
$\mathcal{L}(N, D) = E + A N^{-a} + K D^{-c}$: an irreducible term, a capacity term, and a data term.
For MoE Transformers, total parameters $N$ do not determine how many parameters a token effectively activates, so one $N$ axis cannot describe a dense and a sparse model at once.
Prior MoE scaling laws therefore extend the capacity term with a variable that describes routing: expert count~\cite{clark2022unified}, expert granularity~\cite{ludziejewski2024scaling}, or sparsity~\cite{abnar2025parameters}.
We follow that line: we replace $N$ with per-token FLOPs $F$ and scale the capacity term by compute-equivalent sparsity $S$ from Eq.~\ref{eq:equivalent-sparsity}:
\begin{equation}
\label{eq:separate-surface}
\mathcal{L}(F, S, D)
\;=\;
E
\;+\;
\frac{A\,(1-S)^{b}}
{F^{a}}
\;+\;
\frac{K}{D^{c}}.
\end{equation}
$F$ is the measured training FLOPs per token at the mean document length within the packed 4096-token contexts (since segment-level masks restrict attention to each document); it includes the attention cost that scales with that length and therefore exceeds the parameter-only $6N$ approximation.
$E$ is the irreducible loss.
The middle term is capacity-limited: more per-token FLOPs $F$ shrink it, and so does higher sparsity (lower $1{-}S$), because a sparser model stores more total parameters for the same per-token cost.
The last term is data-limited: more training tokens $D$ reduce it.
We fit this form separately for the Baseline and for \archname{}, giving each its own six coefficients $(E, A, K, a, b, c)$.
There is no shared loop indicator: all six coefficients may differ.

\paragraph{Compute-optimal frontier.}
Given a total training compute budget $C$ and a sparsity level $S$, we want to split $C$ between per-token FLOPs $F$ and training tokens $D = C / F$ to minimize loss.
Following Chinchilla Approach~3~\cite{hoffmann2022training}, we minimize $\mathcal{L}(F, S, D)$ subject to $C = F \cdot D$.
The capacity term and the data term pull in opposite directions: a larger $F$ shrinks the first but leaves fewer tokens $D = C/F$ for the second.
Balancing the two gives the compute-optimal per-token FLOPs
\begin{equation}
\label{eq:f-star}
F^{*}
\;=\;
\Biggl(
\frac{a\, A\, (1-S)^{b}\,
      C^{c}}
{c\, K}
\Biggr)^{1/(a+c)},
\end{equation}
and the corresponding optimal tokens $D^{*} = C / F^{*}$.
By construction, both $F^{*}$ and $D^{*}$ are power laws in $C$, as in Chinchilla.
Substituting back, the lowest achievable loss at budget $C$ is
\begin{equation}
\label{eq:frontier}
\mathcal{L}^{*}(C, S)
\;=\;
E \;+\; \underbrace{f(A, K, a, b, c, S)}_{\text{prefactor}}\; C^{-\gamma},
\qquad
\gamma \;=\; \frac{a\,c}{a + c}.
\end{equation}
The prefactor depends on $A$, $K$, $a$, $b$, $c$, and $S$, while $\gamma$ controls how quickly the reducible term decreases with compute.

\paragraph{Compute efficiency gain.}
When two architectures are fitted with separate surfaces, they trace different frontiers.
Holding performance fixed and asking how much compute an improvement saves is the standard lens on algorithmic progress~\cite{hernandez2020measuring}.
We define the \textit{compute efficiency gain} (CE Gain) of a target architecture over a reference architecture as the fraction of compute the target saves at a common loss:
\begin{enumerate}
\item Evaluate the target's frontier loss at budget $C_{\mathrm{tgt}}$.
\item Invert the reference's frontier to find the budget $C_{\mathrm{ref}}$ at which it reaches the same loss.
\item Compute the fraction saved:
\end{enumerate}
\begin{equation}
\label{eq:ce-gain}
\mathrm{CE\ Gain}
\;=\;
1 - \frac{C_{\mathrm{tgt}}}{C_{\mathrm{ref}}}.
\end{equation}
A positive value means the target reaches the same loss with less compute than the reference.
Work on algorithmic progress measures the same quantity as a multiplier $C_{\mathrm{ref}}/C_{\mathrm{tgt}}$~\cite{davidson2023ceg,ho2024algorithmic}.
In our case the target is \archname{} and the reference is the Baseline.

\FloatBarrier

\subsection{\archname{} saves training compute on the frontier}
\label{sec:scaling-results}

\paragraph{Fit protocol.}
We fit Eq.~\ref{eq:separate-surface} separately for the Baseline and for \archname{} on the sparse grid $S \approx \{85\%,95\%,97\%\}$, using the same Huber-loss ($\delta = 10^{-3}$) minimization on log-loss with L-BFGS-B for both architectures~\cite{hoffmann2022training}.
Keeping the procedure identical for both is important, because compute-optimal conclusions are sensitive to fitting choices~\cite{porian2024resolving}.
Not all endpoints enter the same fit.
The $S = 0$ dense-reference control sits far from the three sparse levels; including it in a joint fit worsens the surface RMSE on the sparse grid, because the optimizer compromises between two regimes.
We therefore exclude $S = 0$ from the scaling-law fit.
The sparse fit uses 72 Baseline endpoints and 72 \archname{} endpoints from the three levels $S \approx \{85\%,95\%,97\%\}$.
These 144 endpoints span $1.3 \times 10^{19}$ to $2.2 \times 10^{21}$ cumulative training FLOPs, which is the window the fitted surfaces are supported on.

\begin{figure}[H]
\centering
\includegraphics[width=\linewidth]{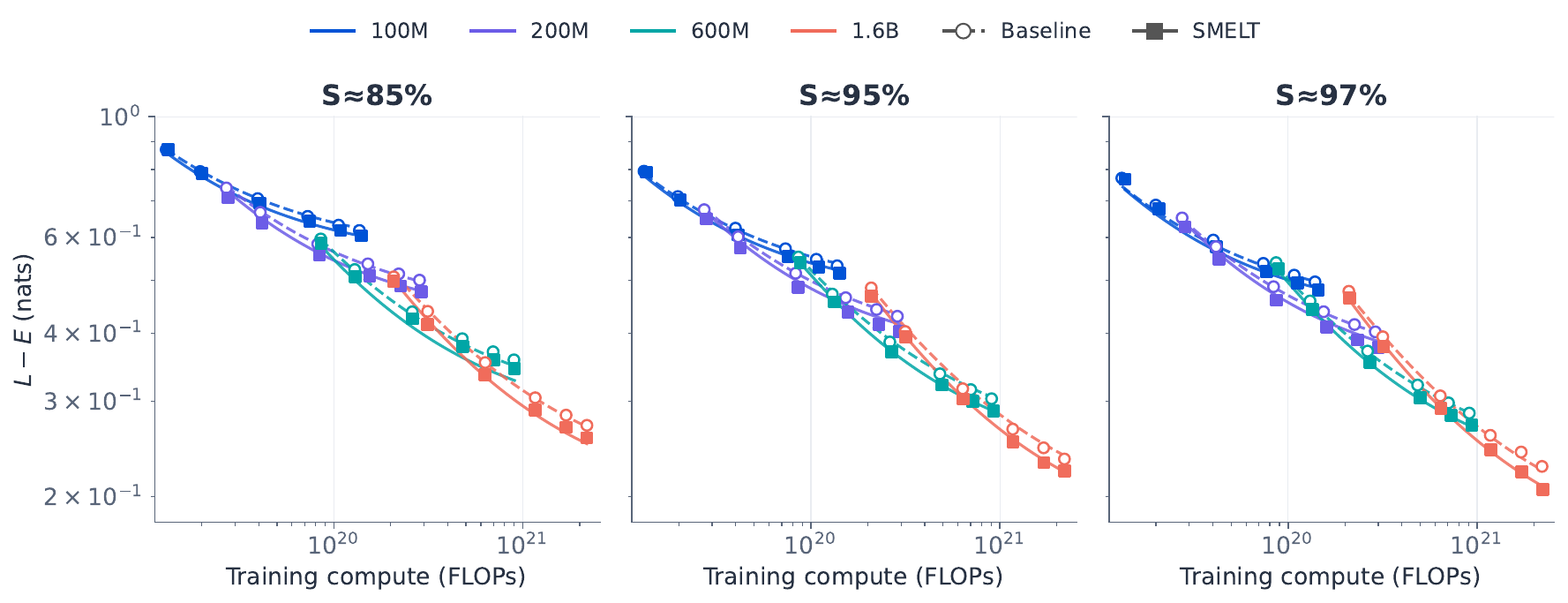}
\caption{Reducible loss $\mathcal{L} - E$, where $E$ is each architecture's own irreducible loss, vs.\ training compute on the sparse scaling-law grid $S \approx \{85\%,95\%,97\%\}$.
  The $S=0\%$ dense-reference control is omitted because its distinct regime is excluded from the sparse-grid fit.
  At every scale and sparsity, \archname{}'s reducible loss decreases faster with compute.}
\label{fig:reducible}
\end{figure}

\paragraph{Fitted coefficients.}
Table~\ref{tab:separate-fit} reports the six coefficients for each architecture.
\begin{center}
\begin{minipage}{\linewidth}
\small
\captionof{table}{Separate-surface coefficients on $S \approx \{85\%,95\%,97\%\}$,
  rounded to four decimal places ($A$ and $K$ to four significant figures in scientific notation).
  RMSE is in nats on validation loss itself, not on log-loss; the fit objective is
  Huber loss on log-loss.}
\label{tab:separate-fit}
\centering
\begin{tabular}{lccccccc}
\toprule
Arch & $E$ & $A$ & $K$ & $a$ & $b$ & $c$ & RMSE \\
\midrule
Baseline & 1.4439 & $1.366\times 10^{3}$ & $1.975\times 10^{6}$ & 0.3703 & 0.1530 & 0.6594 & 0.00554 \\
\archname{} & 1.4493 & $1.963\times 10^{3}$ & $5.264\times 10^{6}$ & 0.3892 & 0.1460 & 0.7011 & 0.00952 \\
\bottomrule
\end{tabular}
\end{minipage}
\end{center}
The exponents $a$ and $c$ decide how fast loss drops with scale and data.
\archname{}'s capacity exponent $a$ is 0.3892 vs.\ the Baseline's 0.3703, and its data exponent $c$ is 0.7011 vs.\ 0.6594.
Both are larger, so \archname{}'s reducible loss $\mathcal{L} - E$ drops faster along the compute axis.
Combining $a$ and $c$ into the frontier exponent $\gamma = ac/(a+c)$ from Eq.~\ref{eq:frontier} gives $\gamma_{\mathrm{base}} = 0.237$ and $\gamma_{\mathrm{SMELT}} = 0.250$.
The 5.5\% higher $\gamma$ means \archname{}'s frontier loss drops faster per unit compute.
Figure~\ref{fig:reducible} confirms this visually: after subtracting each architecture's own $E$, \archname{}'s reducible-loss curves are steeper at every scale and sparsity.

Note that $E_{\mathrm{SMELT}}$ is slightly higher than $E_{\mathrm{base}}$ in Table~\ref{tab:separate-fit}, but the gap (0.005 nats) is smaller than the fit RMSE of either architecture, so we attribute it to fitting noise rather than a genuine difference in asymptotic floors.
Predicting frontier behavior at much larger budgets (e.g.\ $C = 10^{25}$) would require training runs at those scales to anchor the fit.

\paragraph{CE Gain on the frontier.}
The steeper frontier translates into concrete compute savings.
Table~\ref{tab:ce-gain} applies the CE Gain procedure from Section~\ref{sec:scaling-form} at three compute budgets.
Its brackets come from a \textit{cell bootstrap}: we resample the twelve sparse grid cells with replacement, refit both surfaces on each resample, recompute the quantity of interest, and report the 2.5th and 97.5th percentiles over 2{,}000 draws.

\begin{center}
\begin{minipage}{\linewidth}
\small
\captionof{table}{CE Gain on the compute-optimal frontier: the fraction of
  training compute \archname{} saves at the same loss.
  Brackets are cell-bootstrap 95\% intervals (2{,}000 draws).}
\label{tab:ce-gain}
\centering
\begin{tabular}{lccc}
\toprule
\archname{} budget $C_{\mathrm{tgt}}$ & $S\approx85\%$ & $S\approx95\%$ & $S\approx97\%$ \\
\midrule
$10^{20}$ FLOPs & 10.0\% $[1, 22]$ & 7.8\% $[3, 15]$ & 6.8\% $[4, 14]$ \\
$10^{21}$ FLOPs & 18.0\% $[8, 28]$ & 15.8\% $[10, 25]$ & 14.7\% $[8, 25]$ \\
$10^{22}$ FLOPs$^\dagger$ & 23.5\% $[8, 42]$ & 20.9\% $[0, 48]$ & 19.6\% $[0, 51]$ \\
\bottomrule
\end{tabular}

\vspace{2pt}
\raggedright\footnotesize
$^\dagger$~Extrapolated beyond the fitted compute window.
\end{minipage}
\end{center}

At $C = 10^{20}$, \archname{} saves 6.8--10.0\% of training compute across sparsity levels.
At $C = 10^{21}$, the range widens to 14.7--18.0\%.
Because $\gamma_{\mathrm{SMELT}} > \gamma_{\mathrm{base}}$, the gap between the two frontiers compounds over the fitted compute range at all three sparsities (the additive increment from $10^{20}$ to $10^{21}$ is ${\sim}8$\,pp in each column).
$S$ affects the level of the gain rather than its growth rate: $S \approx 85\%$ has the largest CE Gain because the sparsity term $(1{-}S)^b$ amplifies the gap, not because the gap widens faster.

\paragraph{Compute-optimal allocation.}
The separate surfaces also let us ask whether \archname{} should allocate compute differently from the Baseline.
From Eq.~\ref{eq:f-star}, the optimal per-token FLOPs scale as $F^{*} \propto C^{c/(a+c)}$.
A larger $c/(a+c)$ means a larger share of budget goes to model size rather than to tokens.
We measure this through tokens per dense-equivalent parameter (TPP $= D^{*} / (F^{*}/6)$), where $F/6$ converts per-token FLOPs to an approximate active parameter count via the standard $6$~FLOPs-per-parameter-per-token ratio~\cite{kaplan2020scaling}.
Table~\ref{tab:tpp} reports TPP at $C = 10^{21}$.
Uncertainty comes from the same cell bootstrap as Table~\ref{tab:ce-gain}, here at 200 draws because each draw refits both surfaces and then re-solves the allocation.

\begin{center}
\begin{minipage}{\linewidth}
\small
\captionof{table}{Compute-optimal tokens per dense-equivalent parameter
  (TPP) at $C = 10^{21}$.  Brackets are cell-bootstrap 95\% intervals.}
\label{tab:tpp}
\centering
\begin{tabular}{lcc}
\toprule
$S$ & Baseline TPP & \archname{} TPP \\
\midrule
$\approx85\%$ & 56 $[51, 65]$ & 56 $[44, 77]$ \\
$\approx95\%$ & 78 $[71, 90]$ & 75 $[60, 101]$ \\
$\approx97\%$ & 91 $[83, 104]$ & 86 $[70, 114]$ \\
\bottomrule
\end{tabular}
\end{minipage}
\end{center}

The two architectures want nearly the same split: \archname{}'s TPP point estimate is within 6\% of the Baseline's at every sparsity, and the bootstrap intervals overlap almost completely (Table~\ref{tab:tpp}).
The data term in Table~\ref{tab:separate-fit} explains why the two stay close.
\archname{} has both a larger data coefficient $K$ ($5.264 \times 10^{6}$ vs.\ $1.975 \times 10^{6}$) and a larger data exponent $c$ (0.7011 vs.\ 0.6594).
The larger $K$ makes \archname{}'s data-limited loss $K/D^c$ higher at small
$D$, below the crossover at $D \approx 1.6 \times 10^{10}$ tokens.
The larger $c$ makes the term fall faster as $D$ grows.
The two effects pull the optimal split in opposite directions and roughly cancel, so \archname{}'s compute saving comes from reaching a lower loss at the same allocation rather than from reallocating the budget.
Both architectures sit well above the ${\sim}20$ TPP that Chinchilla reports for dense models~\cite{hoffmann2022training}; sparse MoE models generally prefer more tokens per active parameter, consistent with prior work~\cite{abnar2025parameters}.
\section{Downstream Performance and Domain Analysis}
\label{sec:data-analysis}

The CE Gain in Section~\ref{sec:scaling-law} comes from validation loss, a single number that averages over all domains and sample lengths.
A lower validation loss does not guarantee better downstream performance, and an aggregate improvement may hide biases across domains or sample types.
This section tests whether the advantage transfers to independent benchmarks and examines how it distributes across training domains, sample lengths, and in-context learning.

\subsection{Downstream performance amplifies the validation loss gap}
\label{sec:dclm-mmlu}

\begin{figure}[!ht]
\centering
\includegraphics[width=\linewidth]{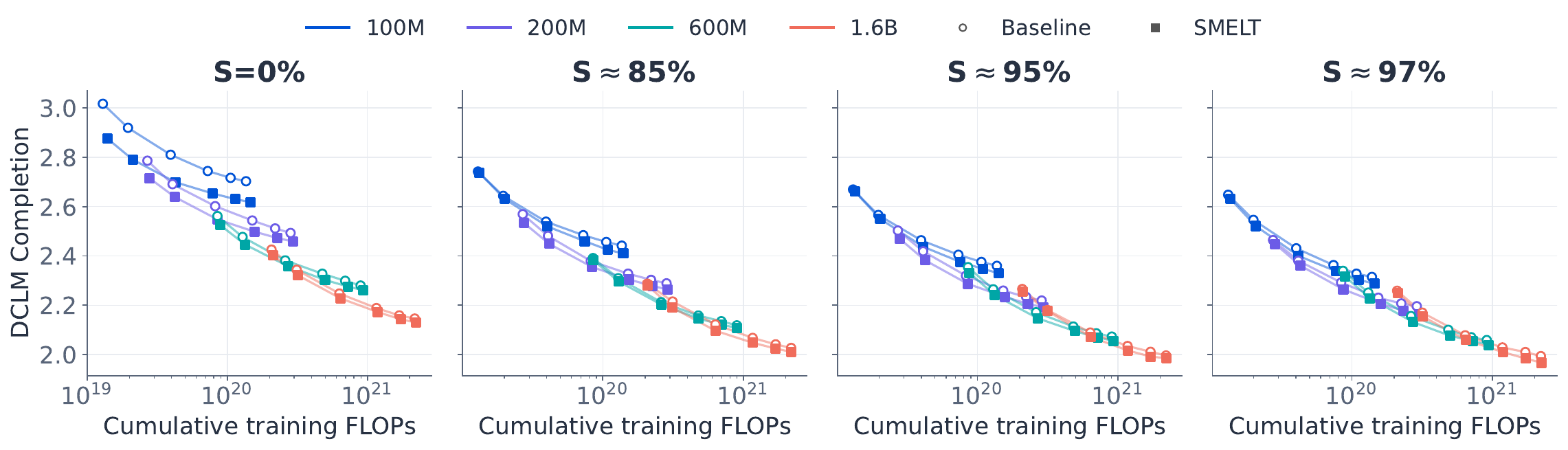}
\caption{DCLM Completion (gold-completion micro loss) vs.\ cumulative training FLOPs across compute-equivalent sparsity levels.
  Lower is better; \archname{} wins all 96 matched pairs.}
\label{fig:downstream-completion}
\end{figure}

\begin{figure}[!ht]
\centering
\includegraphics[width=\linewidth]{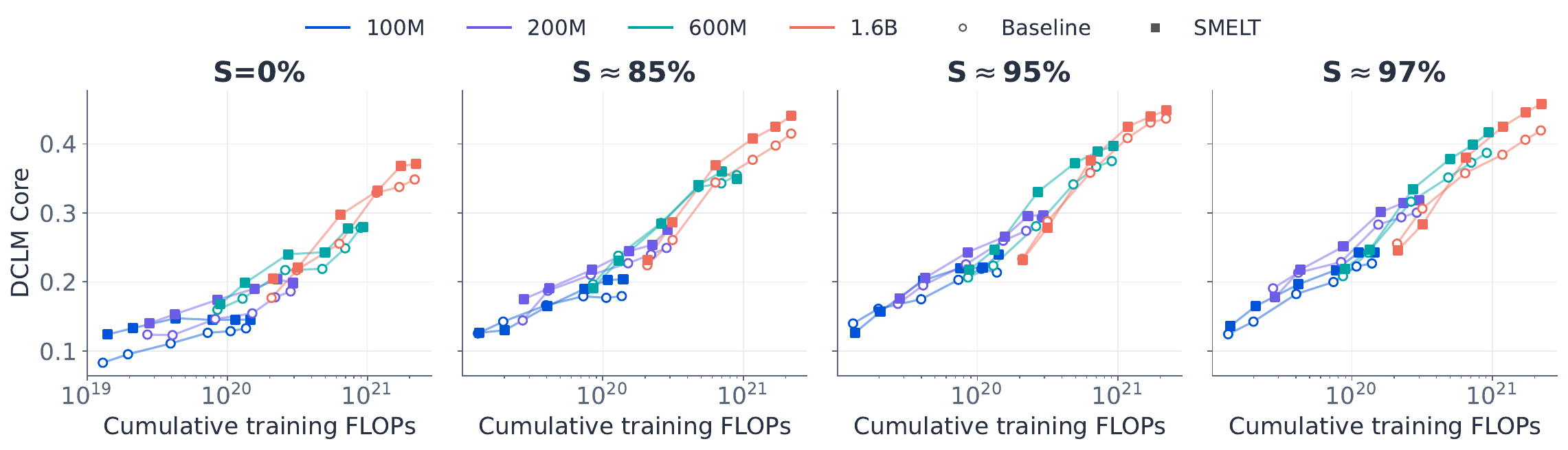}
\caption{DCLM Core vs.\ cumulative training FLOPs across compute-equivalent sparsity levels.
  Higher is better; \archname{} wins 83 of 96 matched pairs.}
\label{fig:downstream-core}
\end{figure}

\begin{figure}[!ht]
\centering
\includegraphics[width=\linewidth]{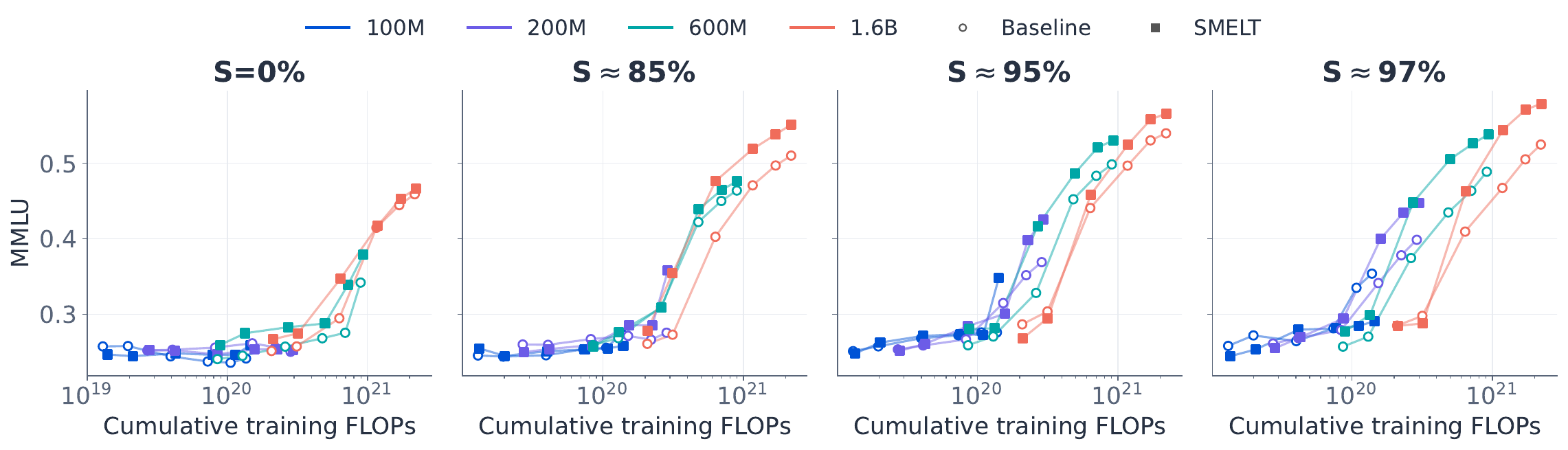}
\caption{MMLU accuracy vs.\ cumulative training FLOPs across compute-equivalent sparsity levels; higher is better.
  The dotted line marks chance accuracy (25\%).
  All 96 pairs are shown; because pairwise comparisons are unstable near the chance floor, the win-rate summary is restricted to the 30 pairs whose Baseline is at least 10 percentage points above chance, of which \archname{} wins 29.}
\label{fig:downstream-mmlu}
\end{figure}

\archname{} wins on DCLM Completion in \textbf{96 of 96} matched pairs, on DCLM Core in 83 of 96, and on MMLU in 29 of the 30 pairs whose Baseline scores at least 10 percentage points above chance (Figures~\ref{fig:downstream-completion}--\ref{fig:downstream-mmlu}).

\begin{figure}[!ht]
\centering
\begin{subfigure}[t]{0.44\linewidth}
\centering
\includegraphics[width=\linewidth]{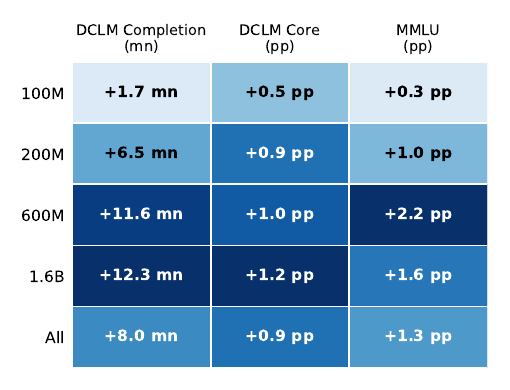}
\caption{By aggregate benchmark.}
\label{fig:residual-heatmap}
\end{subfigure}\hfill
\begin{subfigure}[t]{0.54\linewidth}
\centering
\includegraphics[width=\linewidth]{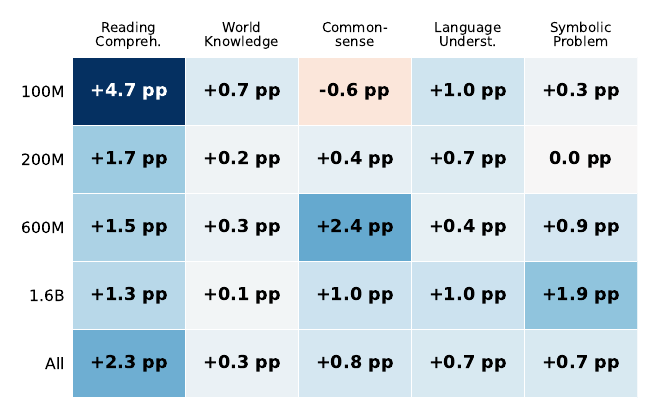}
\caption{By domain category.}
\label{fig:domain-loop-excess}
\end{subfigure}
\caption{Downstream residual of \archname{} beyond what the validation loss improvement predicts (Eq.~\ref{eq:residual}), broken down by model scale.
  In panel~(a), the mean residual is positive on every benchmark at every scale; it increases monotonically with scale on DCLM Completion and DCLM Core, while the two larger scales exceed the two smaller ones on MMLU.
  mn = millinats ($10^{-3}$ nats); pp = percentage points; DCLM Core and the category breakdown use centered accuracy (the DCLM-v2 reference baseline maps to zero), while MMLU uses raw accuracy.
  Panel~(b) fits the same sigmoid separately to each DCLM task before category aggregation.
  The ``All'' row pools residuals over endpoints from all four scales rather than fitting a separate row.
  Positive residuals mean \archname{} outperforms the Baseline-fitted curve; negative residuals mean it underperforms that curve.}
\label{fig:excess-heatmaps}
\end{figure}

\paragraph{Residual analysis against the Baseline curve.}
A model with lower validation loss should score better on any benchmark that tracks validation loss, so these win rates restate Section~\ref{sec:scaling-law} rather than add to it.
The key question is whether the downstream improvement is proportional to the validation loss reduction or exceeds it.
To test this, we fit a calibration from validation loss $\ell$ to each benchmark score $y$, using all 96 Baseline endpoints as training data.
Following prior task-scaling work that models loss-to-accuracy mappings with a sigmoid~\cite{bhagia2025taskscaling}, we use the same four-parameter monotone sigmoid family for all three metrics:
\begin{equation}
  \hat{y}_m(\ell)
  \;=\;
  b_m + \frac{a_m}{1+\exp[-d_m\alpha_m(\ell-\tau_m)]},
  \qquad a_m,\alpha_m>0,
  \label{eq:sigmoid-calibration}
\end{equation}
where $d_m=+1$ for DCLM Completion, whose loss increases with validation loss, and $d_m=-1$ for the higher-is-better DCLM Core and MMLU scores; $a_m,b_m,\alpha_m,\tau_m$ are fitted separately for each metric.
For the category breakdown in Figure~\ref{fig:domain-loop-excess}, we likewise fit the decreasing form ($d_m=-1$) independently to each of the 22 DCLM tasks before averaging the chance-centered residuals within each category.
The fits are tight: $R^2 = 0.997$ for DCLM Completion, $0.974$ for DCLM Core, and $0.911$ for MMLU.
For each \archname{} endpoint $i$ with validation loss $\ell_i$ and measured benchmark score $y_i$, we define the residual
\begin{equation}
  \delta_i \;=\; s\!\cdot\!\bigl(y_i \;-\; \hat{y}(\ell_i)\bigr),
  \label{eq:residual}
\end{equation}
where $s=+1$ for higher-is-better metrics and $s=-1$ for DCLM Completion (a loss), so that a positive $\delta_i$ always means \archname{} outperforms a Baseline that reaches the same validation loss.

\begin{figure}[!ht]
\centering
\includegraphics[width=\linewidth]{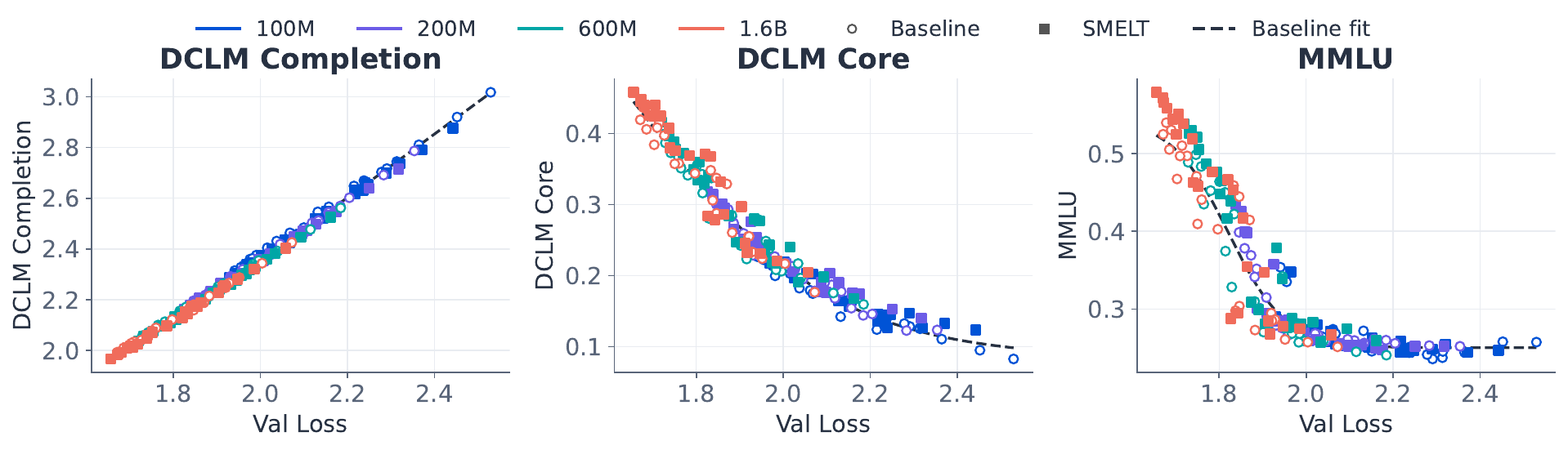}
\caption{Benchmark scores against validation loss.
  DCLM Completion is a loss (lower is better); DCLM Core and MMLU are scores (higher is better).
  The curve fitted to the 96 Baseline endpoints uses the monotone sigmoid family in Eq.~\ref{eq:sigmoid-calibration} for all three metrics ($R^2 = 0.997$, $0.974$, $0.911$ from left to right).
  At every scale on average, \archname{} lies below the Baseline curve on DCLM Completion and above it on DCLM Core and MMLU: its downstream advantage exceeds what its validation-loss improvement alone predicts.}
\label{fig:loss-to-benchmark}
\end{figure}

\paragraph{The residual is positive on all three benchmarks and the excess grows with scale.}
Figure~\ref{fig:loss-to-benchmark} shows that \archname{} points separate from the Baseline curve favorably (below for DCLM Completion, above for the other two); Figure~\ref{fig:residual-heatmap} breaks down the mean residual by scale.
The mean residual $\bar{\delta}$ is positive on every benchmark at every scale.
More importantly, $\bar{\delta}$ increases monotonically with model scale on DCLM Completion and DCLM Core; on MMLU the two larger scales exceed the two smaller ones.
\archname{}'s downstream advantage therefore exceeds what its validation loss improvement accounts for, and this excess grows as models scale up.
Figure~\ref{fig:domain-loop-excess} further breaks down the residual by DCLM domain category (Appendix~\ref{app:dclm-categories}).
Reading Comprehension shows the largest excess when all scales are pooled, but the dominant category shifts as models grow, and Symbolic Problem Solving overtakes it at 1.6B.
This suggests that the second visit increasingly benefits structured reasoning as model capacity grows.

\FloatBarrier

\subsection{\archname{} gains most on structured data}
\label{sec:per-source}

The CE Gain in Section~\ref{sec:scaling-results} averages over all validation domains.
We now ask whether the gain is uniform across data types or concentrated on specific domains.

\paragraph{Method.}
To compute CE Gain on a subset of the data, we need a scaling surface restricted to that subset.
A single subset does not have enough data to independently fit all six coefficients in Eq.~\ref{eq:separate-surface}.
We therefore share the exponents $(a, b, c)$ from the aggregate fit (Table~\ref{tab:separate-fit}) and fit only the intercepts $(E_s, A_s, K_s)$ per subset, separately for the Baseline and for \archname{}.
This lets each subset converge to its own entropy floor while borrowing curvature from the data-rich aggregate fit.
We then apply the CE Gain inversion from Section~\ref{sec:scaling-form} at $C = 10^{21}$ and $S \approx 95\%$.
We apply this procedure under two different groupings.

\paragraph{By domain category.}
The held-out validation set spans five categories (Section~\ref{sec:protocol}): Code, Math/STEM, Finance, Knowledge, and Web.
We fit per-category intercepts and compute CE Gain for each.

\paragraph{By Baseline validation loss.}
Instead of using human-defined categories, we can also group sources by how well the Baseline already models them.
Here we fit intercepts per individual source rather than per group, which gives a CE Gain for each of the 39 sources.
Within each matched pair we then rank the 39 sources by the Baseline's per-source validation loss and split them into four near-equal groups (10, 10, 10, 9): Q1 holds the sources with the lowest Baseline loss (data the Baseline already predicts well) and Q4 the highest.
The CE Gain of each source is fixed (determined by the per-source intercept fit), but its quartile assignment can change from pair to pair because the ranking is based on each pair's Baseline loss; the average is taken within each quartile and then across all 96 pairs.

\begin{figure}[!ht]
\centering
\begin{subfigure}[t]{0.48\linewidth}
\centering
\includegraphics[width=\linewidth]{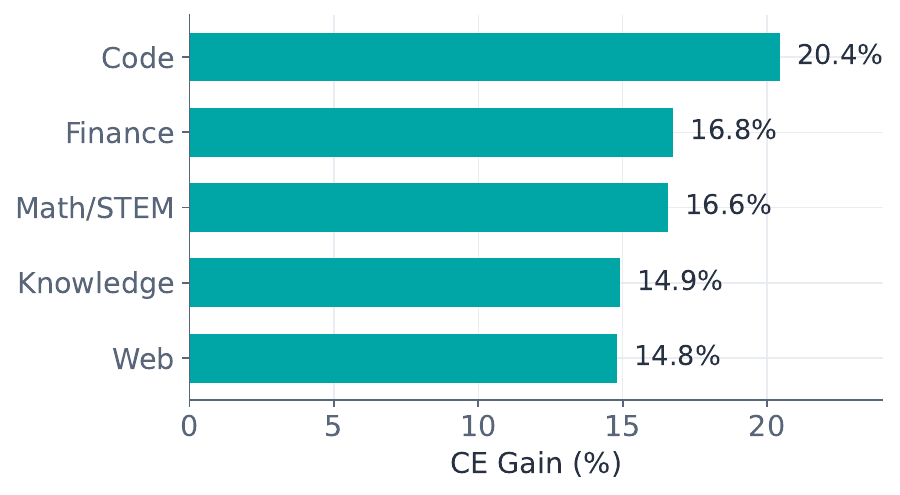}
\caption{By domain category.}
\label{fig:per-source}
\end{subfigure}\hfill
\begin{subfigure}[t]{0.48\linewidth}
\centering
\includegraphics[width=\linewidth]{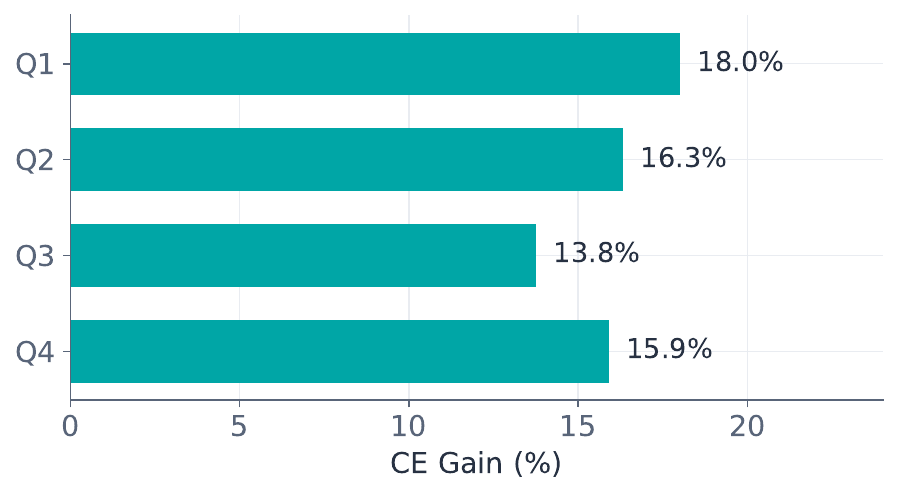}
\caption{By Baseline-loss quartile.}
\label{fig:refinement}
\end{subfigure}
\caption{CE Gain on the compute-optimal frontier at $C = 10^{21}$ and $S \approx 95\%$.
  (a)~From per-category intercepts: all five categories are positive and Code leads.
  (b)~Within each matched pair, the 39 validation sources are ranked by Baseline per-source loss and split into four near-equal groups (10/10/10/9), from Q1 (lowest loss) to Q4 (highest).
  Averaged per-source CE Gain is highest in Q1, and the profile is U-shaped with its minimum at Q3.}
\label{fig:per-source-combined}
\end{figure}

\paragraph{Results.}
Figure~\ref{fig:per-source-combined} shows CE Gain under both groupings.

CE Gain is positive on all five categories.
Code leads at 20.4\%; Finance and Math/STEM follow at 16.8\% and 16.6\%; Knowledge and Web sit at 14.9\% and 14.8\%.
The ranking tracks the degree of internal structure: Code has strict syntax and long-range dependencies; Web text is less constrained.

On the Baseline-loss axis, CE Gain is not monotone: it falls from Q1 (18.0\%) through Q2 (16.3\%) to a minimum at Q3 (13.8\%), then rebounds at Q4 (15.9\%).
The sources the Baseline already models well gain the most, but the hardest quartile is not the one that gains least.
We read the Q1 end as the informative one: it lines up with the category breakdown, since Code has both the strongest structure and the lowest Baseline loss, and it has the highest CE Gain.
Why the high-loss tail rebounds is not something our data settles; one possibility is that Q4 mixes genuinely noisy sources, where neither architecture can improve much, with hard-but-structured ones.

\FloatBarrier

\subsection{Gain grows with sample length and in-context examples}
\label{sec:length-icl}

Section~\ref{sec:per-source} groups the gain by data domain.
We now examine two orthogonal dimensions: sample length and number of in-context examples.

\begin{figure}[!ht]
\centering
\includegraphics[width=\linewidth]{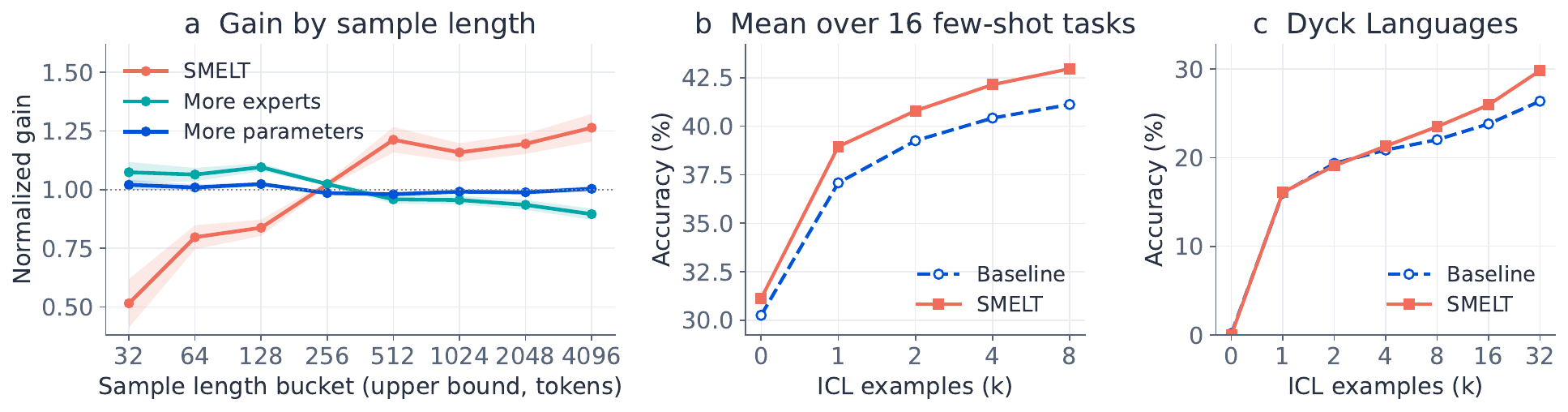}
\caption{\archname{}'s extra gain concentrates on long samples and on in-context learning.
  (a)~Loss improvement by document length, normalized to mean~1 within each contrast so that the curves compare shape rather than magnitude.
  The three families are \archname{} against its matched Baseline, and two Baseline-only controls that add experts or add active parameters.
  \archname{}'s gain on the four longest buckets is 1.52$\times$ its gain on the four shortest; the two controls stay flat (0.98 and 0.88).
  (b)~Mean accuracy over 16 few-shot tasks (15 DCLM tasks plus MMLU) against the number of in-context examples $k$.
  The gap is 0.9~pp at $k = 0$ and 1.9~pp once demonstrations are available.
  (c)~Dyck Languages task is demonstration-sensitive: both architectures score near 0\% without examples, and the only way to improve is to read the demonstration answers.
  Here the gap keeps widening with $k$ rather than plateauing, reaching 29.8\% against 26.4\% at $k = 32$.}
\label{fig:length-icl}
\end{figure}

\paragraph{\archname{}'s gain concentrates on long samples.}
We evaluate per-token loss on the validation set, grouping samples into eight length buckets by each document's own token count (upper bounds 32, 64, \ldots, 4096 tokens).
We construct three families of pairwise contrasts and, for each pair, normalize the per-bucket improvement by its mean so that only the shape of the gain profile is compared:
\begin{itemize}\setlength{\itemsep}{0pt}
  \item \archname{} vs.\ Baseline: each \archname{} run paired with its matched Baseline at the same scale and sparsity level;
  \item More experts: consecutive sparsity levels ($S = 0\% \to 85\%$, $85\% \to 95\%$, $95\% \to 97\%$) within the Baseline at the same model size;
  \item More parameters: consecutive active-parameter scales (e.g.\ 200M$\to$600M) within the Baseline at the same sparsity level.
\end{itemize}
Figure~\ref{fig:length-icl}(a) plots the averaged normalized shape for each family.
\archname{}'s gain concentrates on long samples: the mean normalized gain over the four longest buckets (512--4096 tokens) is 1.52$\times$ the mean over the four shortest (32--256 tokens).
In contrast, the two Baseline families show no such tilt: adding parameters improves short and long samples roughly equally (ratio 0.98), and adding experts even favors short samples slightly (ratio 0.88).

\paragraph{\archname{} benefits more from in-context examples.}
Figure~\ref{fig:length-icl}(b) averages accuracy over 16 few-shot tasks, the 15 DCLM tasks that take demonstrations plus MMLU, as a function of the number of in-context examples $k$.
This sweep is a separate evaluation from the aggregate metrics: it re-runs each task at $k \in \{0,1,2,4,8\}$ (and QA Wikidata, ARC Easy, CS Algorithms, Dyck Languages, and Operators additionally at $k = 16$ and $32$) rather than at the fixed shot counts of Appendix~\ref{app:dclm-categories}, which is why the accuracies here do not match the DCLM Core numbers elsewhere.
The gap between the two architectures is 0.9~pp at $k=0$ and widens to 1.9~pp once demonstrations are provided ($k=1$); it then holds through $k=8$.
We highlight Dyck Languages (Figure~\ref{fig:length-icl}(c)) because it is a demonstration-sensitive exact-match task: both architectures score near 0\% without demonstrations, and the only way to improve is to read the example answers.
On this task, \archname{} reaches 29.8\% at $k=32$ vs.\ the Baseline's 26.4\%.
On both the average and this demonstration-sensitive task, \archname{} gains more when in-context examples are available.
Per-task accuracy curves for the 14 of these 16 tasks where \archname{}'s max-shot gain is significant ($p < 0.05$, paired permutation test) are provided in Appendix~\ref{app:icl-tasks}.

\FloatBarrier
\section{Inside the Second Pass}
\label{sec:model-analysis}

Sections~\ref{sec:scaling-law}--\ref{sec:data-analysis} quantify the advantage of the second visit.
To build intuition for what happens inside the model, we run all matched pairs on a 1M-token held-out validation sample and probe expert routing, residual-stream updates, and attention patterns, then ground the observations in a Dyck-language case study.

\subsection{Do the two visits activate the same experts?}
\label{sec:expert-routing}

Each MoE layer selects a subset of experts for each token.
When the same token passes through the same physical layer on the second visit, does the router pick the same experts, or does the updated residual-stream input lead it to a different subset?

\begin{figure}[!ht]
\centering
\includegraphics[width=\linewidth]{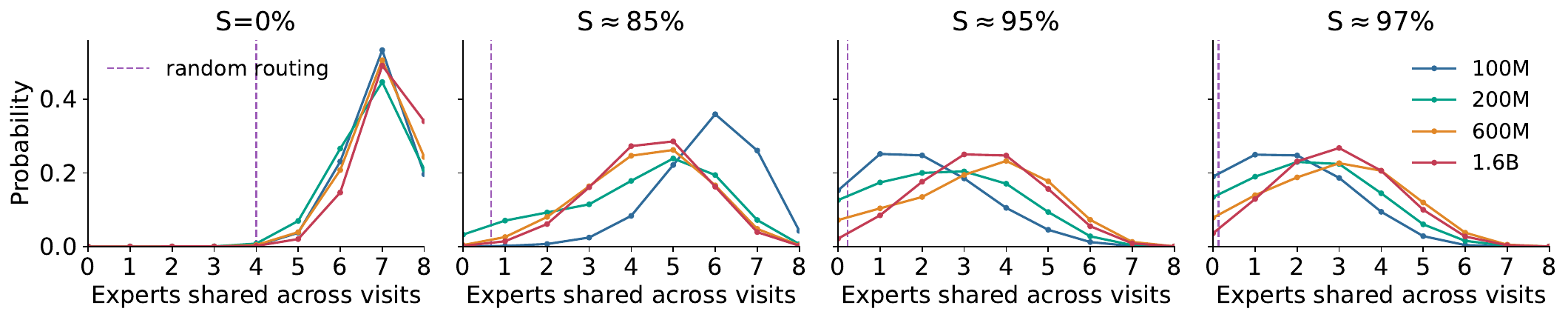}
\caption{Distribution of experts shared between the two visits for the same
  token and physical gate.  Four panels show the $S = 0\%$ dense-reference
  control and $S \approx \{85\%,95\%,97\%\}$.
  Dashed purple line: expected overlap under independent random routing ($8^2/n$ for top-8 routing over a candidate pool of $n$ experts).
  At $S = 0\%$, the same token reuses
  nearly all 8 selected experts; at $S \approx 97\%$, it reuses only 2--3, still far above chance.}
\label{fig:expert-reuse}
\end{figure}

\paragraph{Measurement.}
For each token and each repeated physical MoE layer, we record the top-8 expert sets selected by the gating network on visit~1 and visit~2, then count how many experts appear in both sets.
This overlap count ranges from 0 (completely different routing) to 8 (identical routing).
Figure~\ref{fig:expert-reuse} plots the distribution of this overlap count across all tokens at four sparsity levels: the $S{=}0\%$ dense-reference control and $S \approx \{85\%,95\%,97\%\}$.

\paragraph{The router reuses a core subset and diversifies the rest.}
At $S{=}0\%$, where each layer has the smallest candidate pool in the grid, the router reuses nearly all~8.
As sparsity increases and the pool grows by more than an order of magnitude, the overlap drops to 2--3.
Even there it stays far above what independent random routing over a pool that large would give (dashed line in Figure~\ref{fig:expert-reuse}), so the router deliberately reuses a core subset.
The second visit reuses part of the first visit's expert set and diversifies the rest.

\subsection{Does the second visit write more or less into the residual stream?}
\label{sec:residual}

Each Transformer layer updates the residual stream as $x_{\ell+1} = x_\ell + \Delta x_\ell$.
When the same physical layer executes twice, is the second write $\Delta x_\ell^{(2)}$ larger or smaller than the first $\Delta x_\ell^{(1)}$?
And do the two writes point in the same direction, or does the second visit write something orthogonal?

\paragraph{Setup.}
In a pre-norm Transformer, each layer decomposes into an attention write and an MoE write (we write $\mathrm{LN}$ for the pre-norm; our implementation uses RMSNorm~\cite{zhang2019root}).
Within the looped span, every sublayer's residual update is scaled by $1/r$ where $r$ is the loop count (Section~\ref{sec:protocol}):
\begin{align}
\Delta x_\ell^{\mathrm{attn}} &= \tfrac{1}{r}\,\mathrm{Attn}\bigl(\mathrm{LN}(x_\ell)\bigr), \label{eq:attn-write} \\
\Delta x_\ell^{\mathrm{moe}} &= \tfrac{1}{r}\,\mathrm{MoE}\bigl(\mathrm{LN}(x_\ell + \Delta x_\ell^{\mathrm{attn}})\bigr), \label{eq:moe-write}
\end{align}
so the full layer update is $\Delta x_\ell = \Delta x_\ell^{\mathrm{attn}} + \Delta x_\ell^{\mathrm{moe}}$ and the MoE sub-layer sees the already-scaled attention write in its input.
Outside the looped span $r = 1$, recovering the standard pre-norm form.
For each of these three quantities, we measure the $\ell_2$ norm on visit~1 and visit~2 and report the ratio $\lVert\Delta x_\ell^{(2)}\rVert / \lVert\Delta x_\ell^{(1)}\rVert$; a ratio above~1 means the second visit writes a larger update.

\begin{figure}[!ht]
\centering
\includegraphics[width=\linewidth]{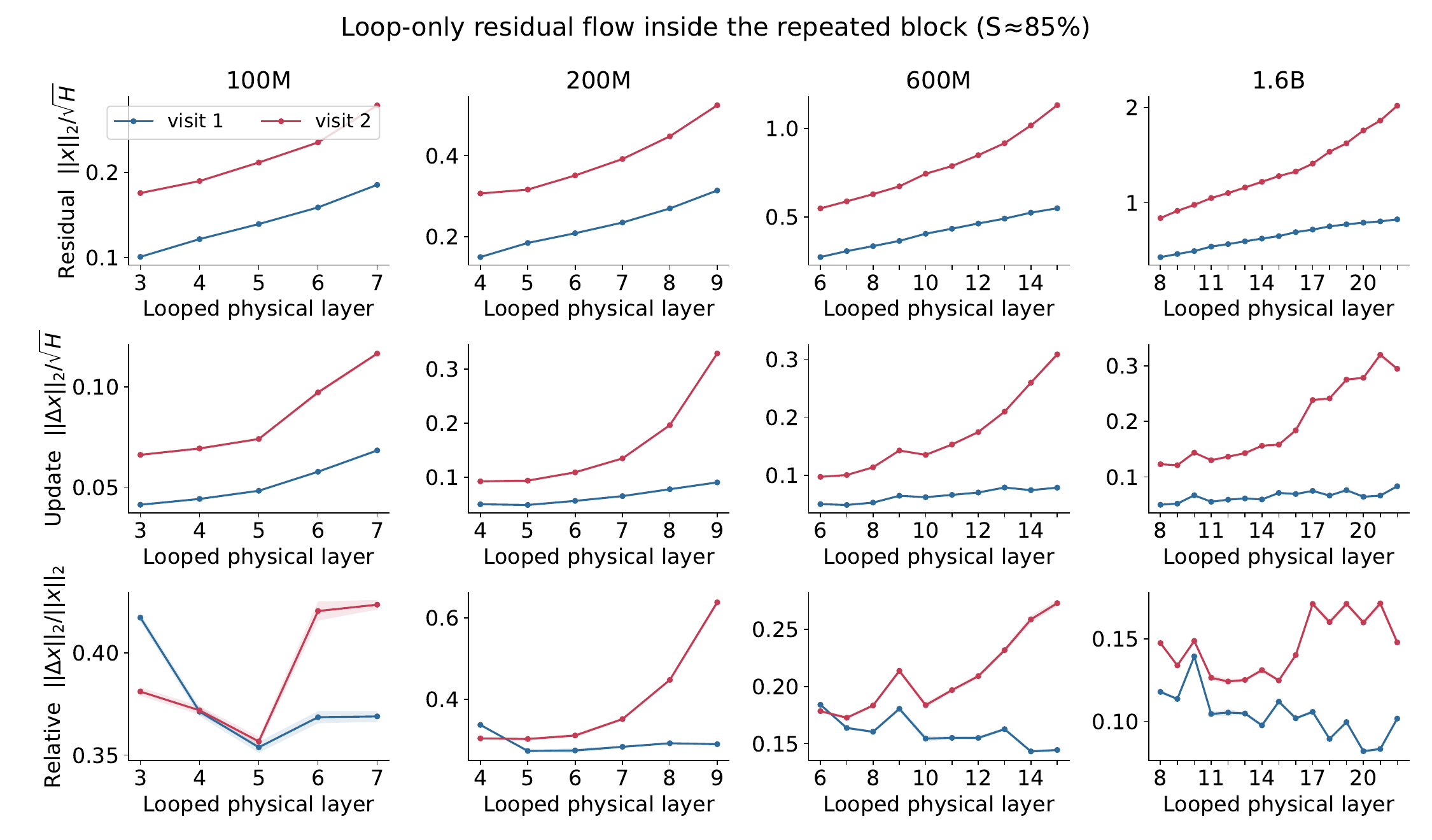}
\caption{Residual flow inside the repeated block (matched $S \approx 85\%$,
  four scales).  Row~1: residual RMS $\lVert x_\ell\rVert_2/\!\sqrt{H}$; row~2: update
  RMS $\lVert\Delta x_\ell\rVert_2/\!\sqrt{H}$; row~3: relative update
  $\lVert\Delta x_\ell\rVert_2 / \lVert x_\ell\rVert_2$.  Visit~2 (red) writes
  larger updates than visit~1 (blue) absolutely at every layer (row~2) and
  relative to the residual at the three larger scales (row~3).}
\label{fig:residual-zoom}
\end{figure}

\paragraph{Visit 2 writes more, both absolutely and relatively.}
Figure~\ref{fig:residual-zoom} shows the layer-by-layer residual flow inside the repeated block at $S \approx 85\%$ across four scales.
Row~2 plots the update norm $\lVert\Delta x_\ell\rVert$: at all four scales, visit~2 (red) writes a larger update than visit~1 (blue) at every layer.
Row~3 normalizes by the residual itself, plotting $\lVert\Delta x_\ell\rVert / \lVert x_\ell\rVert$.
At the three larger scales the relative update is also consistently higher on visit~2.
At 100M, whose repeated block spans only five layers, a few early layers are comparable or reversed, likely because the residual has not yet diverged enough between visits.

\begin{figure}[!ht]
\centering
\includegraphics[width=\linewidth]{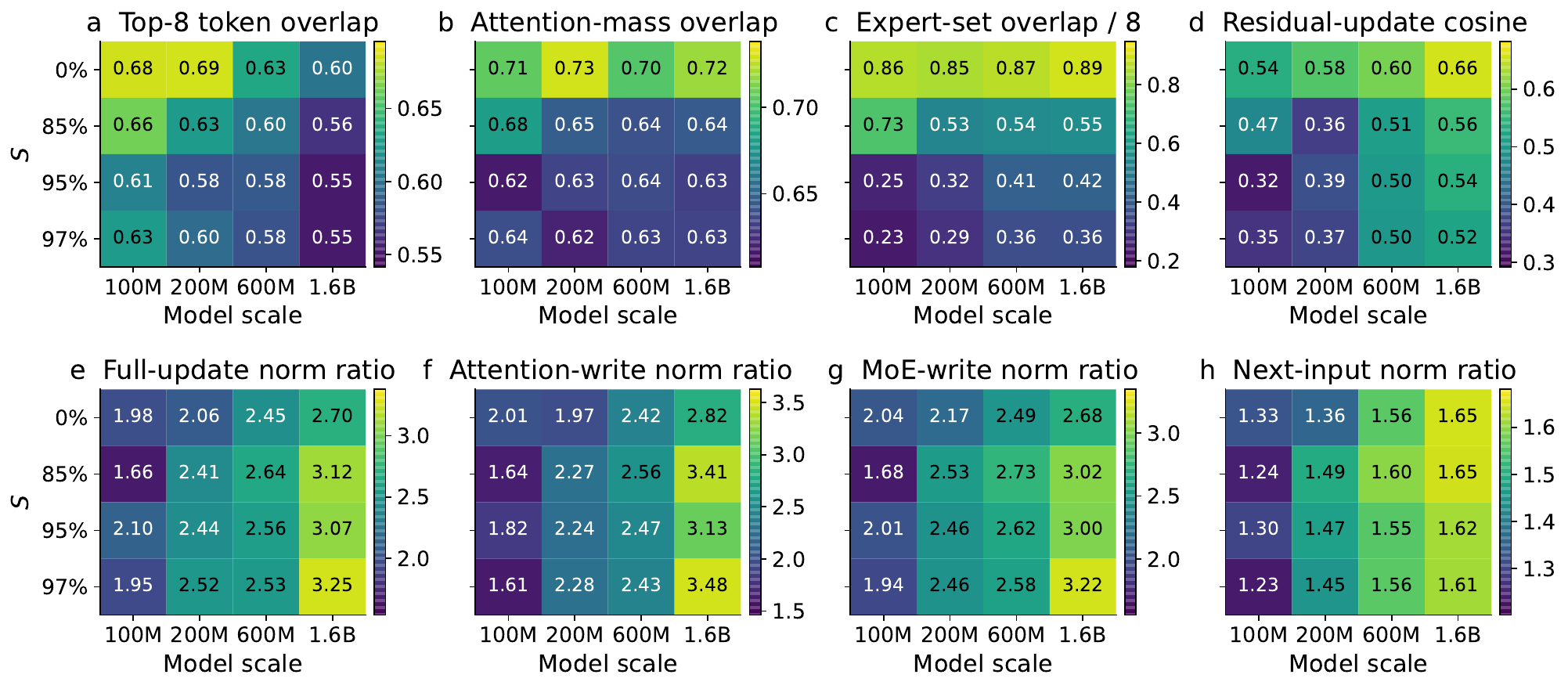}
\caption{Eight cross-visit diagnostics over the $4 \times 4$ grid, with scale on
  the horizontal axis and compute-equivalent sparsity $S$ on the vertical axis.
  (a)--(d)~Similarity between the two visits: top-8 attended-token overlap,
  attention-mass overlap, expert-set overlap divided by 8, and residual-update
  cosine.
  (e)--(h)~Visit-2\,/\,visit-1 norm ratios: the full update $\Delta x_\ell$, the
  attention write $\Delta x_\ell^{\mathrm{attn}}$, the MoE write
  $\Delta x_\ell^{\mathrm{moe}}$, and the RMSNorm output that feeds the next
  sub-layer.
  All four norm ratios exceed~1 in all 16 cells: the second visit writes more in
  every component, at every scale and sparsity.}
\label{fig:internal-grid}
\end{figure}

\paragraph{Both attention and MoE sub-layers contribute.}
Where does the extra write come from, attention or MoE?
Figure~\ref{fig:internal-grid} collects eight cross-visit diagnostics over the $4 \times 4$ grid.
Panels~e--h each give the visit-2\,/\,visit-1 norm ratio of one component: the full update $\Delta x_\ell$ (panel~e), the attention write $\Delta x_\ell^{\mathrm{attn}}$ (panel~f), the MoE write $\Delta x_\ell^{\mathrm{moe}}$ (panel~g), and the RMSNorm output that feeds the next sub-layer (panel~h).
Panels~a--d carry the similarity measures of Sections~\ref{sec:expert-routing}--\ref{sec:attention-internal} across the same grid, so the single-cell results in those sections can be checked at every scale and sparsity.

In all 16 cells, all four norm ratios exceed~1 (range 1.2--3.5$\times$).
Both the attention and MoE sub-layers contribute: panel~h shows that the RMSNorm output fed to each sub-layer already has a larger norm on visit~2.
Since RMSNorm removes input scale, this difference reflects a directional shift in the residual stream: the element-wise gain $g$ amplifies different dimensions unequally, so a change in direction alone changes the output norm.
Both sub-layer writes grow on top of this larger input.

\begin{figure}[!ht]
\centering
\includegraphics[width=\linewidth]{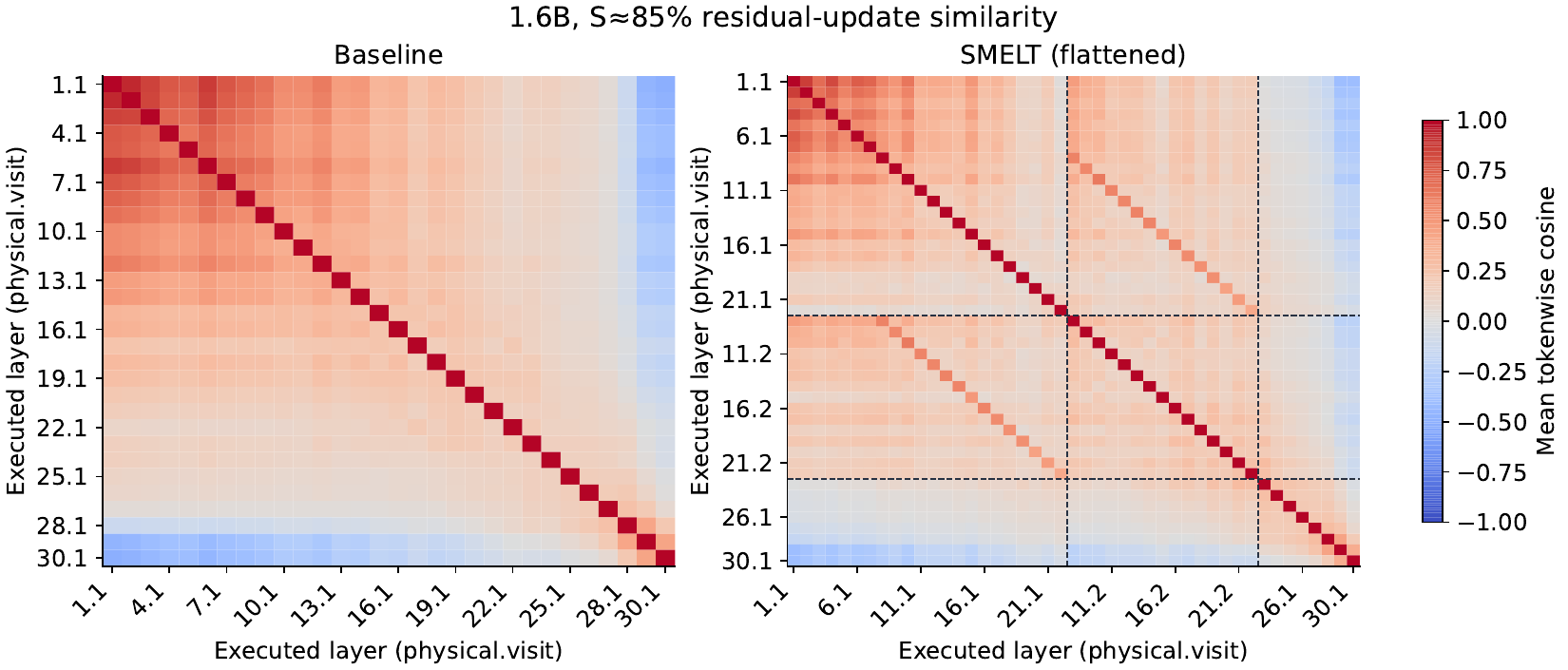}
\caption{Pairwise residual-update cosine for the 1.6B model.
  The Baseline panel (left; 30 layers) provides the unlooped reference for ordinary layer-to-layer cosine structure.
  The \archname{} panel (right) uses flattened execution order; dashed lines mark the repeated block.
  Along the same-physical-layer cross-visit diagonal, cosine ranges from 0.42 to 0.65 (mean 0.56), compared with a mean of 0.16 for nonmatching cross-visit pairs: the two visits write in more closely aligned directions at the same physical layer.}
\label{fig:cosine-matrix}
\end{figure}

\paragraph{The two writes are aligned.}
Knowing that the second visit writes more, we next ask whether it writes in the same direction.
We compute the pairwise cosine similarity between residual updates $\Delta x_\ell$ across all layers.
Figure~\ref{fig:cosine-matrix} shows the resulting matrix for the 1.6B model: the Baseline on the left (30 unique layers), \archname{} on the right (flattened execution order).

The cross-visit block (same physical layer, two visits) has positive cosine: the second visit writes in a similar direction to the first.
The second visit does not override the first; it amplifies the signal the first visit established.

\subsection{Does the second visit look at the same places or different ones?}
\label{sec:attention-internal}

The second visit recomputes Q, K, and V from the updated residual stream; it does not reuse cached activations from the first visit.
Does this lead to different retrieval patterns (different Q and K, hence attending to different positions), or does the model preserve where it looks and only change what it reads?

\paragraph{Setup.}
Recall that the attention write from Eq.~\ref{eq:attn-write} decomposes as
\begin{equation}\label{eq:attn-stages}
\underbrace{\hat x_\ell}_X = \mathrm{LN}(x_\ell),
\quad
Q = \hat x_\ell\, W_Q,
\quad
K = \hat x_\ell\, W_K,
\quad
V = \hat x_\ell\, W_V,
\quad
P = \mathrm{softmax}\!\Bigl(\frac{QK^\top}{\sqrt{d_k}}\Bigr),
\end{equation}
where $Q$ and $K$ set the retrieval coordinates (which positions to attend to), $V$ carries the content at those positions, and $P$ is the attention weight matrix.
The attended context is $PV$ and the final output written into the residual stream is $O = PV\,W_O$.
For each of these six quantities ($X$, $Q$, $K$, $V$, $PV$, $O$), we compute the cosine similarity between visit~1 and visit~2 to see which stages change across visits and which stay fixed.

\begin{figure}[!ht]
\centering
\includegraphics[width=\linewidth]{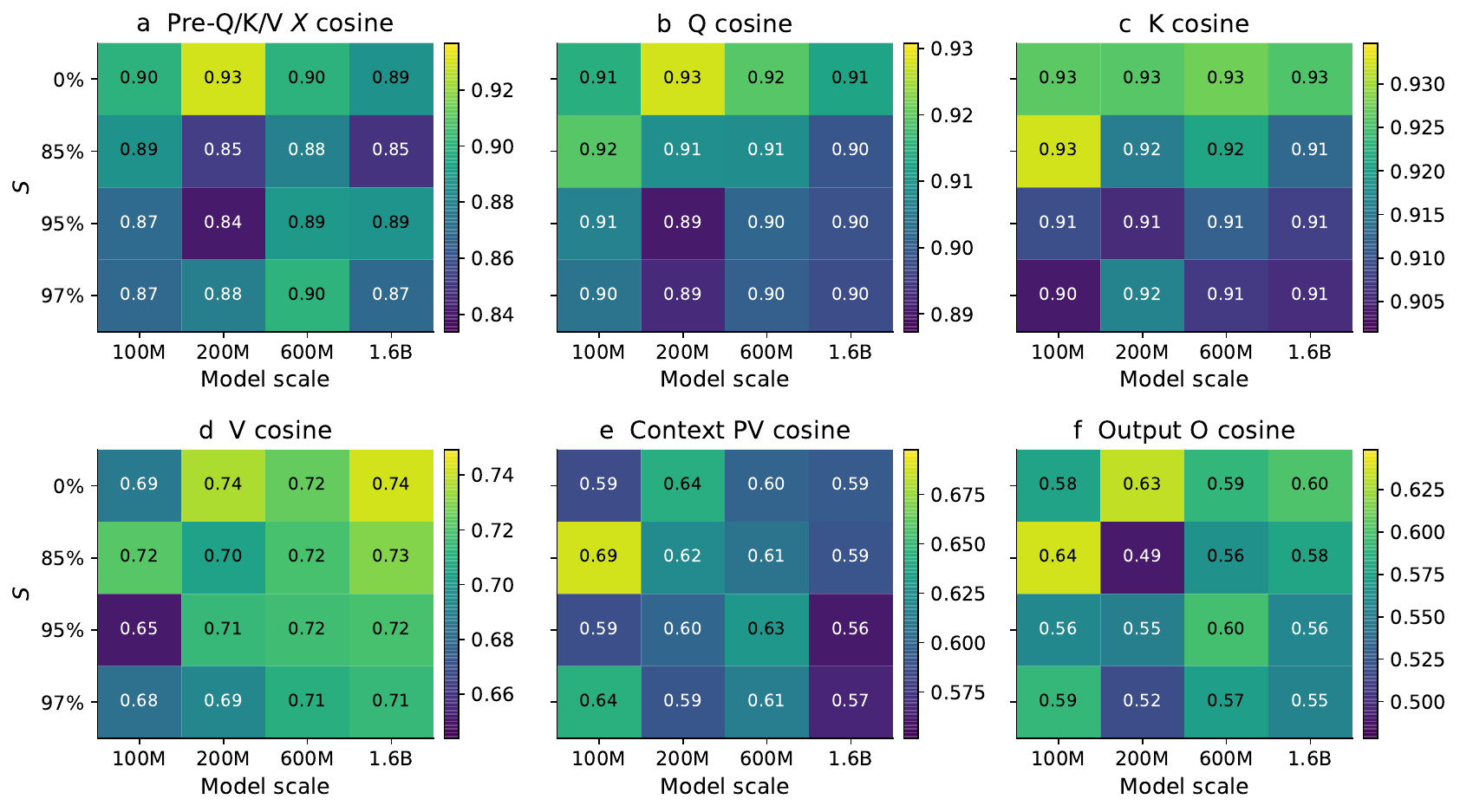}
\caption{Cross-visit cosine similarity at six stages of attention, on the
  full $4 \times 4$ grid: (a)~the shared pre-norm input $X$, (b)~Q, (c)~K,
  (d)~V, (e)~the attended context $PV$, and (f)~the output $O$.
  Q and K stay at 0.89--0.93, at or above the input's own 0.84--0.93, so the
  retrieval coordinates carry over.  V falls below the input to 0.65--0.74, and
  $PV$ and $O$ diverge further.}
\label{fig:qkv-decomp}
\end{figure}

\paragraph{Q and K stay similar; V diverges.}
Figure~\ref{fig:qkv-decomp} shows the result across the full $4 \times 4$ grid.

The shared input $X$ already has cross-visit cosine 0.84--0.93, which sets the reference level for the three projections that read it.
Q and K sit at 0.89--0.93 across all 16 cells, at or above that level: the first visit establishes the retrieval coordinates and the second visit preserves them.
V falls below it to 0.65--0.74, and the downstream stages $PV$ and $O$ diverge further (0.56--0.69 and 0.49--0.64).
The divergence is therefore a property of the value projection rather than of the input it reads: the two visits retain substantially overlapping attention patterns while V changes more than Q/K.

\begin{figure}[!ht]
\centering
\includegraphics[width=\linewidth]{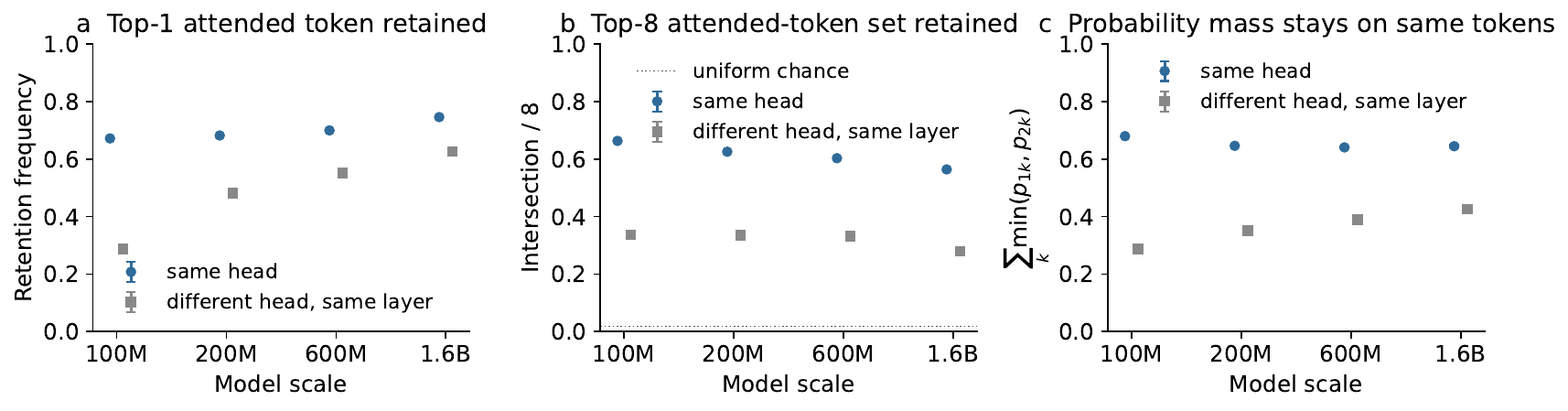}
\caption{Attention token overlap between the two visits of the repeated block ($S \approx 85\%$, four scales): (a)~top-1 retention, (b)~top-8 set intersection, and (c)~overlap of the full attention distributions.
  Blue compares the same head across visits; grey is a control that pairs each head with a different head in the same layer and visit, sharing the positional context but not the learned weights.
  The top-8 sets overlap by 56--66\% across visits against 28--34\% for the control, so the two visits attend to nearly the same tokens and the overlap is head-specific.}
\label{fig:attn-tokens}
\end{figure}

\paragraph{The two visits attend to nearly the same tokens.}
The Q/K similarity predicts that the two visits should attend to similar positions.
We verify this at the discrete level: for each head, we record the top-8 attended tokens on each visit and measure set intersection.
As a control, we pair each head with a different head at the same layer and visit, which shares the same positional context but uses different learned weights; the pairing is a fixed offset of half the head count, so every head is used exactly once.
Figure~\ref{fig:attn-tokens} reports the result across all heads in the repeated block.

The top-8 attended tokens overlap by 56--66\% across visits (Figure~\ref{fig:attn-tokens}b).
Selecting 8 out of up to 4{,}096 positions and recovering more than half the same set is a strong signal that the retrieval pattern is preserved.
The other-head control sits at 28--34\%, confirming that this overlap is head-specific, not an artifact of positional bias.
The gap is narrower on top-1 retention (Figure~\ref{fig:attn-tokens}a), where every head in a segment tends to land on the sink token and at the larger scales the control therefore starts high.
Together with the Q/K--V split above, this section's picture is: the two visits retain substantially overlapping attention patterns while V changes more than Q/K.

\subsection{Case study: Dyck language}
\label{sec:dyck}

Section~\ref{sec:length-icl} shows that \archname{}'s advantage grows with the number of in-context examples.
To understand why, we study Dyck-language bracket matching, a demonstration-sensitive exact-match task where the correct answer is unambiguous: to close a bracket, the model must retrieve the matching opening symbol from a demonstration in the prompt.
This lets us trace, at the individual token level, whether the second visit improves retrieval of the right demonstration.

\paragraph{Setup.}
We construct nested Dyck prompts at $k \in \{1, 2, 4, 8\}$ shots, with three fixed target positions whose correct closing symbol already appears in the $k \ge 1$ prefix.
Each shot presents an input bracket sequence followed by its completion; we call the completion tokens that follow its ``Output:'' marker the \textit{demonstration answer}.
For each scale, we identify the \textit{Dyck head}: we ablate each attention head in the repeated block one at a time and select the head whose removal increases Dyck perplexity the most.
We then visualize this head's full attention pattern to see where mass moves between the two visits.

\begin{figure}[!ht]
\centering
\includegraphics[width=\linewidth]{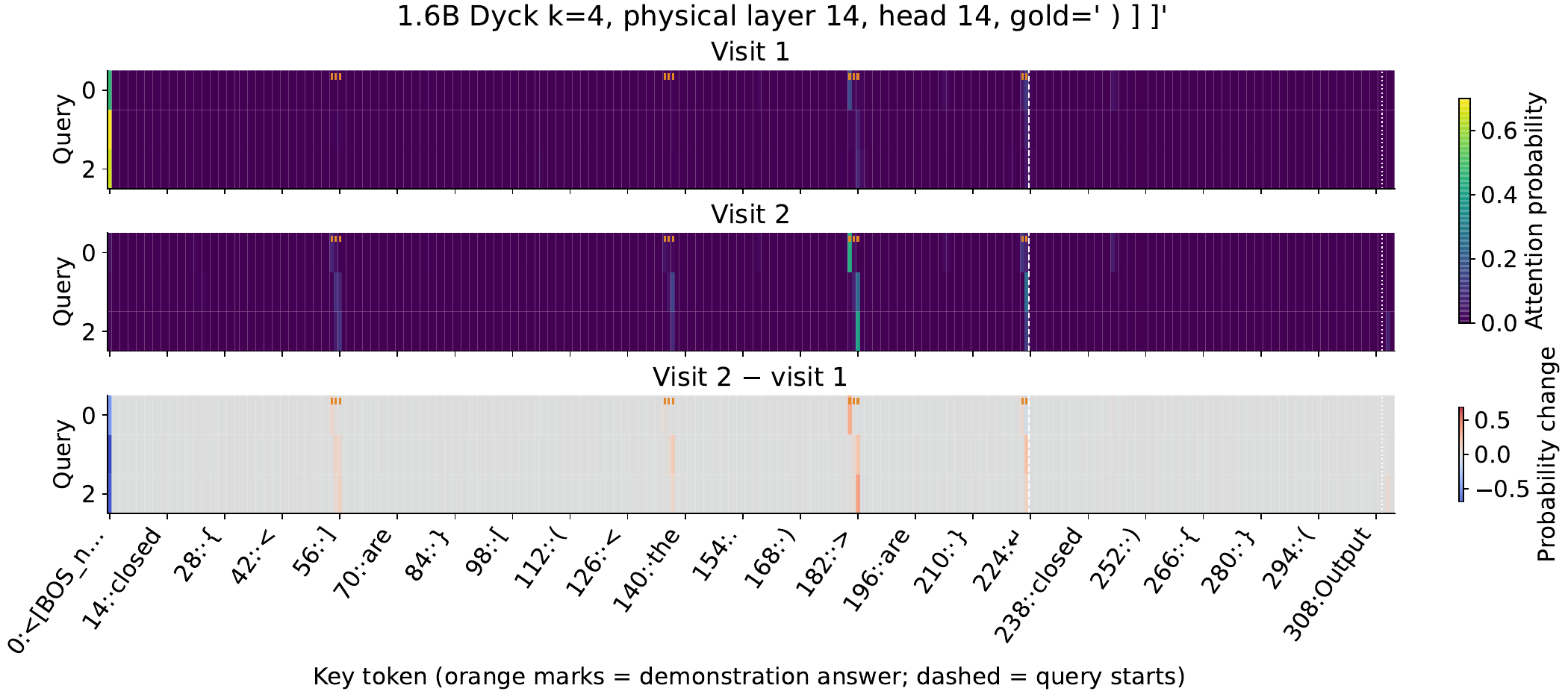}
\caption{Token-level attention map for the \textit{Dyck head} at 1.6B, $S \approx 85\%$, Dyck $k = 4$.
  The Dyck head is the head in the repeated block whose ablation raises Dyck perplexity the most, identified separately at each scale.
  Rows are the three prediction targets; columns are key positions.
  Top: visit~1; middle: visit~2; bottom: difference.
  Orange marks are the 11 demonstration-answer tokens of the four shots, the completions that follow each shot's ``Output:'' marker.
  Averaged over the three targets, the second visit cuts BOS mass from 0.60 to 0.02 and lifts demonstration-answer mass from 0.24 to 0.85, so the sink loses almost exactly what the demonstrations gain.}
\label{fig:dyck-map}
\end{figure}

\paragraph{The attention sink nearly vanishes on the second visit.}
Figure~\ref{fig:dyck-map} shows the Dyck head's attention map at 1.6B for a $k = 4$ example.
On visit~1, the head concentrates mass on the attention sink at the segment-start BOS token and on nearby positions, as if the model does not yet know where to look.
On visit~2, the sink nearly vanishes: mass shifts from BOS onto the demonstration answers in the prefix.
The difference row makes the redistribution explicit.
Averaged over the three targets, BOS mass falls from 0.60 to 0.02 while the demonstration answers rise from 0.24 to 0.85.

We suspect this pattern is related to the ICL advantage reported in Section~\ref{sec:length-icl}: more demonstrations give the second visit more answer spans to attend to, and the sink reduction frees mass to reach them.

\begin{figure}[!ht]
\centering
\begin{subfigure}[t]{0.605\linewidth}
\centering
\includegraphics[width=\linewidth]{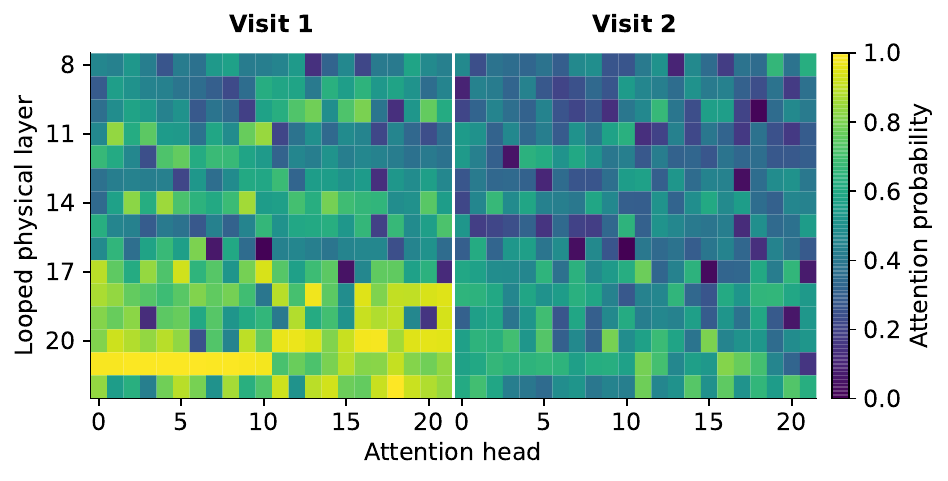}
\caption{Per-head segment-start attention mass at 1.6B.
  Visit~1 (left) concentrates mass on the segment start; visit~2 (right) reduces it across nearly every head.}
\label{fig:sink-heatmap}
\end{subfigure}\hfill
\begin{subfigure}[t]{0.375\linewidth}
\centering
\includegraphics[width=\linewidth]{img/attention_sink_physical_depth_1p6b/attention_sink_physical_depth_1p6b.pdf}
\caption{Sink mass at matched physical depth.
  The bottom axis indexes Baseline physical layers; the top axis indexes \archname{} execution layers.
  The Baseline grows with depth; \archname{} visit~2 stays below visit~1.}
\label{fig:sink-depth}
\end{subfigure}
\caption{Segment-start attention mass (the attention sink) on the 1M-token held-out sample ($S \approx 85\%$).
  The second visit reduces the sink across nearly all heads and layers, running counter to the Baseline's depth trend at matched physical depth.}
\label{fig:attention-sink-general}
\end{figure}

\paragraph{The sink reduction generalizes beyond Dyck.}
Dyck is a controlled task; does the same sink reduction occur on general pretraining data?
We repeat the measurement on the full 1M-token held-out sample spanning all five source categories.
The per-head heatmap (Figure~\ref{fig:sink-heatmap}) shows that the bright, high-mass cells in visit~1 cool broadly on visit~2.
We further find that this reduction goes against the ordinary depth trend.
Figure~\ref{fig:sink-depth} compares \archname{} and the unlooped Baseline at the same physical depth: in the Baseline, sink mass grows toward later layers, consistent with the sink strengthening with depth in standard Transformers~\cite{barbero2025firsttoken,ruscio2025sinking}, whereas \archname{}'s visit~2 lies below visit~1 throughout the repeated block.
The second visit thus reverses the depth trend instead of inheriting the larger sink that its later position in the execution trace would predict; Appendix~\ref{app:attention-sink-scales} shows all four scales.

\section{Conclusion and Future Work}
\label{sec:conclusion}

We demonstrate that depth reuse is a net win for MoE Transformers under budget matching.
By searching the loop configuration space under budget matching, we identify the \archname{} recipe, which loops the middle half of layers twice at a larger effective depth-to-width ratio and consistently beats the Baseline across four scales and four sparsity levels, saving 6.8--18.0\% of training compute on the frontier.
The advantage extends beyond validation loss: on downstream benchmarks the gain exceeds what validation loss predicts, concentrates on structured data and long samples, and grows with in-context examples.
Our mechanistic probes find a consistent internal signature: the second visit reuses retrieval coordinates, amplifies residual writes, and frees attention mass from the sink.
This suggests that looping acts as a refinement step rather than simply adding capacity.

Several directions remain open.
Due to compute constraints, we conduct the design ablations at the 200M scale (up to 3.9B non-embedding parameters); the optimal loop span or count may differ at larger scales.
We also study only the simplest form of looping: repeating a contiguous block with fully shared weights.
Richer variants relax this in different directions, including per-visit low-rank adapters~\cite{bae2025relaxed}, token-level adaptive recursion depth~\cite{bae2025mixture}, learned halting~\cite{dehghani2019universal,elbayad2020depth}, block-selective sharing~\cite{kapl2026growing}, and cross-token state reuse~\cite{cai2026t2mlr}.
Whether these designs preserve or amplify the CE Gain under budget matching is left to future work.
Separately, our budget matching equates arithmetic FLOPs, parameters, and KV-cache size, not wall-clock cost; serial block re-execution and sparse routing might introduce hardware-efficiency gaps that require systems-level optimization to close.
Finally, understanding exactly which mechanisms inside the second visit are responsible for the gain remains an open question; our probes in Section~\ref{sec:model-analysis} offer a descriptive starting point, and we expect future work to build on it toward a causal account.

Taken together, our results suggest that weight sharing across depth is a viable axis for improving language models at matched compute, complementary to the established axes of scaling width, depth, and expert count.
We look forward to seeing how far this axis can be pushed as compute budgets and model scales continue to grow.
 \section{Contributions}
\textbf{Core Contributors}

Shaowen Wang, Ge Zhang

\textbf{Contributors}

Kairong Luo, Yuhao Wu, Shaofan Liu, Jiaheng Liu, Wenhao Huang, Shen Yan, Jian Li

\textbf{Corresponding Authors}

Shaowen Wang (\href{mailto:wangsw23@mails.tsinghua.edu.cn}{\texttt{wangsw23@mails.tsinghua.edu.cn}})\\
Ge Zhang (\href{mailto:gezhang@umich.edu}{\texttt{gezhang@umich.edu}})\\
Shen Yan (\href{mailto:sheny@bytedance.com}{\texttt{sheny@bytedance.com}})\\
Jian Li (\href{mailto:lijian83@mail.tsinghua.edu.cn}{\texttt{lijian83@mail.tsinghua.edu.cn}})
 \section*{Acknowledgments}
We thank Kaifeng Lyu, Mianqiu Huang, Jinhan Li, Tansheng Zhu, Huaqing Zhang, Yilong Chen, Guancheng Du, Kexian Tang, Yihao Xiao, Zhiyuan Zeng, Jiameng Huang, Bingrui Li, Binghui Li, Li Cao, Zitian Gao, and Rui-Jie Zhu for their valuable input and helpful discussions.

The authors are supported in part  by the National Natural Science Foundation of China Grant 04130200126.

\bibliographystyle{plainnat}

\beginappendix
\seedinputappendix

\end{document}